\documentclass[]{interact}

\usepackage{subcaption}
\usepackage{array}
\usepackage{algorithmic}
\usepackage{amsmath}
\usepackage{amsfonts}
\usepackage{graphicx}
\usepackage{cite}

\usepackage{soul}
\usepackage{multirow}
\usepackage{bm}
\usepackage{arydshln}
\usepackage{mathrsfs}
\usepackage{amsfonts}   
\usepackage{amssymb}    
\usepackage{enumitem}
\usepackage[natbibapa,nodoi]{apacite}
\usepackage{tabularx}

\usepackage{epstopdf}

\usepackage[longnamesfirst,sort]{natbib}
\bibpunct[, ]{(}{)}{;}{a}{,}{,}
\renewcommand\bibfont{\fontsize{10}{12}\selectfont}

\theoremstyle{plain}

\theoremstyle{definition}

\theoremstyle{remark}

\begin{document}


\title{Modeling and Control of AWOISV: A Filtered Tube-Based MPC Approach for Simultaneous Tracking of Lateral Position and Heading Angle}

\author{
\name{Xu Yang, Jun Ni, Hengyang Feng, Feiyu Wang and Tiezhen Wang}}

\thanks{This work is supported by the National Natural Science Foundation of China under Grant 52272431.}
\thanks{The authors are with the School of Mechanical Engineering, Beijing Institute of Technology, Beijing 100081, China. (e-mail:Rvlab@163.com).}

\maketitle

\begin{abstract}
  An all-wheel omni-directional independent steering vehicle (AWOISV) is a specialized all-wheel independent steering vehicle with each wheel capable of steering up to ±90°, enabling unique maneuvers like yaw and diagonal movement. This paper introduces  a theoretical steering radius angle and sideslip angle (\( \theta_R \)-\(\beta_R \)) representation, based on the position of the instantaneous center of rotation relative to the wheel rotation center, defining the motion modes and switching criteria for AWOISVs. A generalized \( v\)-\(\beta\)-\(r \) dynamic model is developed with forward velocity \(v\), sideslip angle \(\beta\), and yaw rate \(r\) as states, and \(\theta_R\) and \(\beta_R\) as control inputs. This model decouples longitudinal and lateral motions into forward and rotational motions, allowing seamless transitions across all motion modes under specific conditions. A filtered tube-based linear time-varying MPC (FT-LTVMPC) strategy is proposed, achieving simultaneous tracking of lateral position and arbitrary heading angles, with robustness to model inaccuracies and parameter uncertainties. Co-simulation and hardware-in-loop (HIL) experiments confirm that FT-LTVMPC enables high-precision control of both position and heading while ensuring excellent real-time performance.
\end{abstract}

\begin{figure}[!t]
  \centering
  \includegraphics[width=0.6\textwidth]{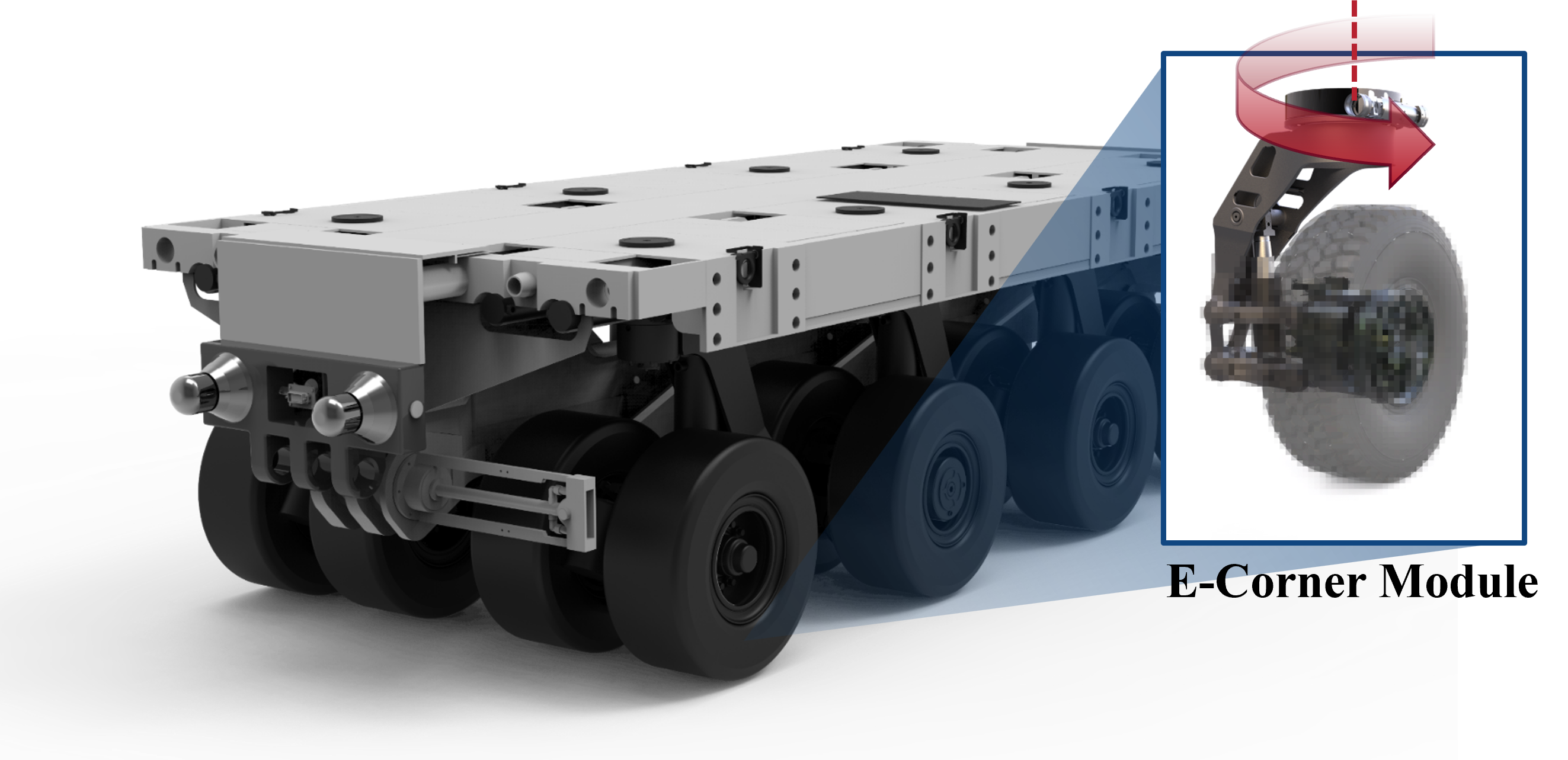}
  \caption{Concept of AWOISV.}
  \label{fig:Cncpt}
\end{figure}

\section{Introduction}

wheel omnidirectional independent steering vehicle (AWOISV) refers to a special type of all-wheel independent steering and driving vehicle (AWISV) in which each wheel can independently steer up to ±90°. The AWOISV is typically equipped with an E-corner module, as illustrated in Fig.\ref{fig:Cncpt}. Compared to traditional vehicles, the AWOISV offer greater freedom in position and pose adjustment. They can perform unique motion modes such as zero radius steering, diagonal steering, and yaw steering, enabling precise positioning in narrow spaces. Typical applications of AWOISV include self-propelled modular transporters (SPMT) \citep{mammoetSelfPropelledModular}, heavy-duty flat transporters (HFT) \citep{comettoSelfpropelledElevatingTransporter}, and military unmanned platforms \citep{niAWIDAWISXByWire2019}. As all operational modes of AWISVs are covered by AWOISVs, researches in this field holds broad applicability. AWOISV greatly enhances maneuverability, yet introduces more significant challenges for dynamic modeling and path tracking control. 

AWISV and AWOISV have clear advantages over traditional vehicles, especially in maneuverability and handling stability. Numerous prototype AWISDVs have been developed. Innovative methods for describing and controlling the spatial motion behavior of AWISV have been proposed. Huang et al. developed a four-wheel independent steering and driving (4WISD) passenger car to enhance handling stability and active safety \citep{wuRearSteeringBasedDecentralized2020}. The configuration of this AWISV closely resembles that of conventional vehicles. Key research areas include fault-tolerant control under failure conditions, integrated lateral and longitudinal control, and stability control based on active yaw rate and sideslip angle regulation \citep{liuFaulttolerantControlApproach2021, liModelIndependentAdaptiveFaultTolerant2013a, hangHandlingStabilityAdvancement2021, zhangIntegratedMotionControl2019}. To obtain higher maneuverability and flexibility, Zhang et al. designed a 4WISD passenger car featuring a novel steering system that integrates an improved two-front-wheel steering mechanism with an omnidirectional independent steering (OIS) mechanism \citep{zhangNovelSteeringSystem2017}. The OIS achieves an angle range of $-35^\circ$ to $+90^\circ$ to enable zero-radius turning and lateral parking maneuvers. Xu et al. developed a six-wheel independent drive and four-wheel independent steering (6WID/4WIS) unmanned ground vehicle with a variable wheelbase to meet the operational requirements for crossing barriers and ditches. The vehicle's passing ability, obstacle crossing capability, differential steering performance, and pitch stability are analyzed in detail \citep{chenPitchStabilityControl2023, jiangHeadingTracking6WID2021}.Qimeng et al. designed an omnidirectional steering agricultural vehicle utilizing electric push rods as actuators and proposed a fuzzy-based wheel deflection control method \citep{xuWheelDeflectionControl2021}. Lam et al. presented an omnidirectional steer-by-wire system that consists of an extended steering interface and a behavior-based steering controller for the AWOISV \citep{tinlunlamOmnidirectionalSteeringInterface2010}. Researches on the kinematic modeling of AWOISVs has been actively pursued. Guowang et al. analyze kinematic behaviors during special modes like zero-radius steering and lateral driving \citep{zhangTireRoadFriction2024}. Christoph et al. introduce a control method for pseudo-omnidirectional vehicles based on virtual guide wheels \citep{stogerVirtualWheelConcept2021}. Xu et al. propose a model with a steering center positioned arbitrarily, using steering radius and vehicle axial direction as primary parameters \citep{xuPathTrackingAgricultural2023}.

However, there are still some insufficiencies in the dynamic modeling description of AWOISVs under various special motion modes. (1) The definitions of various motion modes and the switching criteria are unclear. For instance, switching from a normal steering mode to a diagonal steering mode necessitates that the vehicle come to a complete stop, which results in severe time consumption and restricts driving mobility. (2) Most literatures still characterize the lateral dynamics of AWOISVs using only the steering angle as a single input, leading to underactuation and limiting the full potential of mobility. Although some researchers have proposed control methods like mode-switching buttons \citep{zhangNovelSteeringSystem2017} and expanded steering interfaces \citep{tinlunlamOmnidirectionalSteeringInterface2010}, there is still no systematic study on the mapping relationships between AWOISV's control inputs and states. (3) AWOISV dynamic modeling is still predominantly based on the traditional Ackermann steering framework, assuming small wheel steering angles and keeping the vehicle's sideslip angle within a limited range. Such assumptions fail to accurately capture the dynamics in special motion modes with large wheel steering angles, such as lateral steering and yaw steering.

Path tracking is essential for unmanned vehicle, and various control schemes have been developed, including backstepping \citep{zhangAdaptiveBacksteppingSliding2020}, sliding mode control (SMC) \citep{wangFixedTimeApproachAutomated2024, liangModelFreeOutputFeedback2024}, model predictive control (MPC) \citep{zhaiMPCBasedIntegratedControl2022, wischnewskiTubeModelPredictive2022, wischnewskiTubeMPCApproachAutonomous2023}, and robust control \citep{liRobustSwitchedVelocityDependent2023, huL2GainBasedPathFollowing2024, shiRobustLearningBasedGainScheduled2024}. Independent steering offers greater flexibility and control diversity than traditional front-steering but also presents new challenges in ensuring precise and stable path tracking. Most researchers leverage the over-actuation characteristics of AWISV to enhance the robustness, stability, and fault tolerance during path tracking. An adaptive fast terminal sliding mode fault-tolerant control strategy for 4WISV is proposed by Guo, utilizing a composite observer and adaptive parameter adjustments\citep{guoRobustAdaptiveFaultTolerant2020}. Guo et al. propose a robust adaptive fuzzy backstepping controller to address non-linearities and redundant actuators of 4WISV \citep{guoAdaptiveNonlinearTrajectory2018}. Lin et al. and Hang et al. both develop MPC strategies that integrate steering and torque control to enhance vehicle handling at driving limits. Lin et al. focus on a path-tracking controller combining torque vectoring and four-wheel steering \citep{linPathtrackingControlLimits2024}, while Hang et al. present a tube-based MPC that integrates four-wheel steering with direct yaw-moment control \citep{penghangActiveSafetyControl2021}. Li et al. propose a MIMO-NMPC-based control architecture to track predefined trajectory and velocity under variable deceleration conditions, optimizing tire force usage and improving stability even on low-adhesion roads \citep{liTrajectoryTrackingFourwheel2024}. 

In addition to utilizing all wheel steering to enhance path tracking stability, maximizing the maneuverability of AWOISVs in special motion modes has become a recent research focus. Setiawan et al. establish a single-track kinematic model for 4WISV, considering parallel steering and zero-sideslip maneuvers, and employ the backstepping method for path tracking \citep{setiawanPathTrackingController2016}. Chu et al. propose a tracking control method for the rotation center of a 4WIS vehicle under large-curvature turning conditions. This method transforms traditional wheel angle control into rotation center control, effectively achieving both yaw and lateral movement \citep{chuInstantaneousCenterRotation2024}. Xu et al. implemented a path-tracking method for a 4WIS agricultural vehicle using a fuzzy look-ahead distance-based pure pursuit approach, applicable to both fixed and random steering center positions \citep{xuPathTracking4WIS4WID2022, xuPathTrackingAgricultural2023}. 


In summary, current research on path tracking for AWOISVs largely follows traditional vehicle control frameworks, which restricts motion to specific modes, such as front-wheel, rear-wheel, or all-wheel steering. Lacking the application of special motion modes like yaw or lateral steering and inter-mode transition flexibility limits the simultaneous tracking of lateral position and arbitrary desired heading angles. The fundamental reason is that conventional vehicles have a fixed steering center, determined by the X-coordinate projection in the body coordinate system, which restricts dynamic sideslip angle adjustments. In standard path tracking, vehicle orientation is typically aligned with the reference path tangent \citep{chuanhuShouldDesiredHeading2015}, limiting the ability to track a specified heading angle while maintaining zero lateral deviation. Even with independent steering and driving for sideslip angle correction, heading alignment remains feasible only within a narrow range \citep{tarhiniDualLevelControlArchitectures2024}.

To this end, a novel dynamic model of AWOISV is established in this paper, and a filtered tube-based MPC strategy is proposed and validated to achieve simultaneous tracking of both lateral position and heading angle. The main contributions and innovations of this paper are as follows:
\begin{enumerate}[topsep=0pt, partopsep=0pt]
  \item The motion modes and switching criteria of AWOISVs are rigorously defined based on the relative position between the instantaneous center of rotation and the center of wheel rotation. A theoretical steering radius angle - theoretical sideslip angle (\( \theta_R\)-\(\beta_R \)) representation is proposed to describe the motion state of AWOISVs in different modes.
  \item A generalized dynamic model for AWOISVs with an arbitrary number of axles is constructed, using the forward velocity \( v \), sideslip angle \( \beta \), and yaw rate \( r \) as states, and \( \theta_R \) and \( \beta_R \) as inputs. This model decouples the longitudinal and lateral motions into forward and rotational motions, which covers all motion modes of AWOISVs and enables stepless switching between them under specific conditions. In addition, the dynamic characteristics under different motion modes are analyzed.
  \item A filtered tube-based MPC strategy with a predictive model in Frenet coordinate system is proposed for path tracking of AWOISVs. This method enables simultaneous tracking of both the lateral position and any reference heading angle along a desired path, while providing strong robustness against model uncertainties and feedback state fluctuations. The hardware-in-loop experiment demonstrates good real-time performance.
\end{enumerate}

The remainder of this paper is organized as follows: Section \ref{sec:MtnCharAndRprstaion} introduces the motion characteristics of AWOISVs in different motion modes, and presents the \( \theta_R \)-\(\beta_R \) representation method. In Section \ref{sec:mdlAndAnlysis}, a generalized \(v\)-\( \beta \)-\(r\) dynamic model for the AWOISV is established. Section \ref{sec:ctrlMthd} proposes a filtered tube-based MPC strategy for simultaneously tracking both the lateral position and any reference heading angle. Simulation and experimental results are presented in Section \ref{sec:results}. Section \ref{sec:conclusion} concludes the paper.

\section{Motion Characteristics and Representations of AWOISV}\label{sec:MtnCharAndRprstaion}

\subsection{Motion Characteristics of AWOISV}
\begin{figure}[!t]
  \centering
  \includegraphics[width=0.6\textwidth]{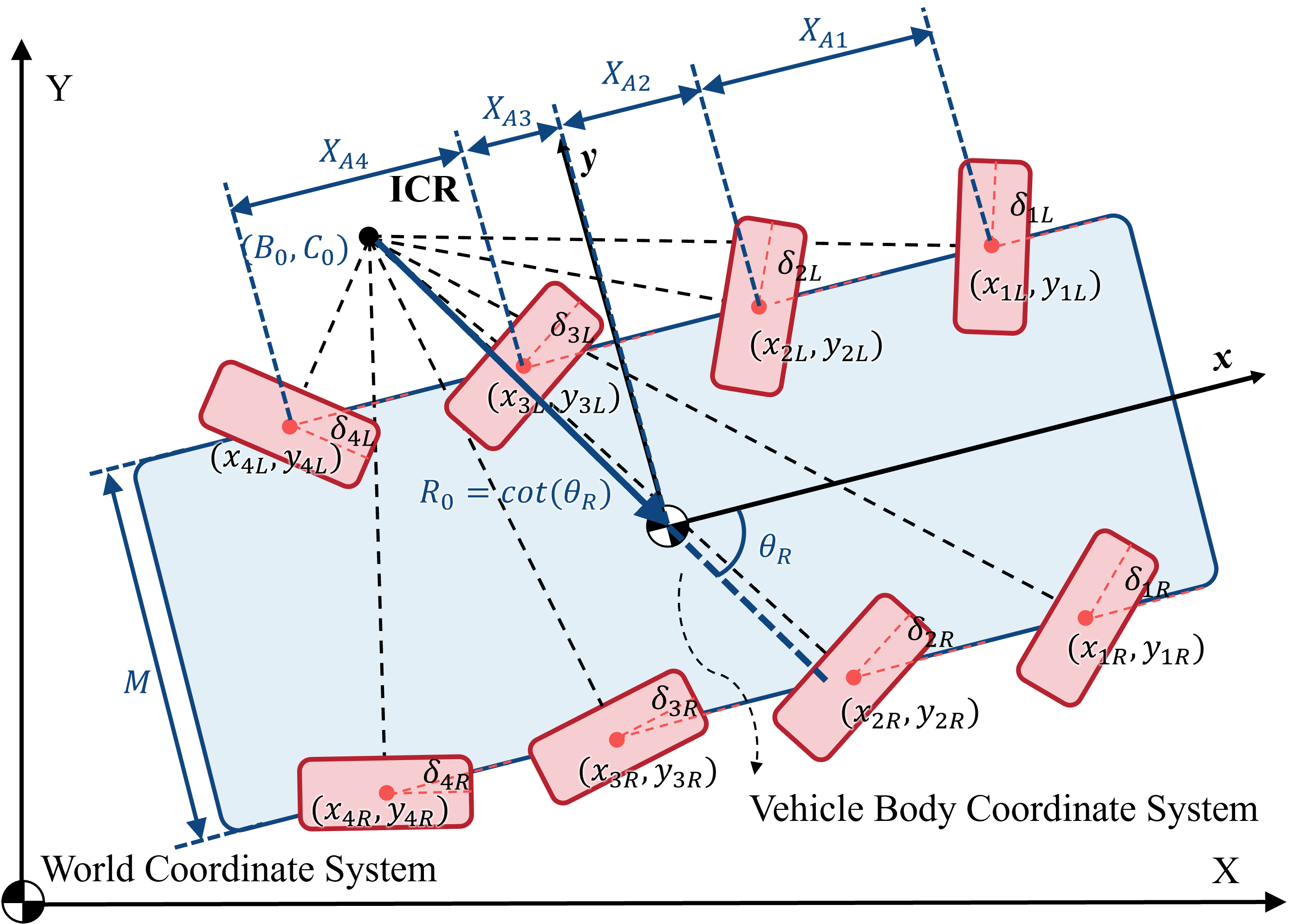}
  \caption{Vehicle coordinate system and relative position between ICR and WCRs.}
  \label{fig:icrPosition}
\end{figure}

Each wheel of the AWOISV can steer independently, with a maximum steering range of $(-\frac{\pi}{2}, \frac{\pi}{2})$. As a result, the instantaneous center of rotation (ICR) can theoretically be located anywhere. When the ICR is on the extension of the rear axle, the vehicle operates in front axle steering mode (FASM). When the steering center is at infinity, the vehicle is in straight or diagonal steering mode (DSM). If the steering center is at the vehicle's centroid, it operates in zero radius steering mode (ZRSM). Therefore, the position of the steering center is crucial for describing the motion characteristics of AWOISV.

Based on the position of the ICR relative to  the center of gravity (CG) of the vehicle, the steering angles of each wheel of the AWOISV can be calculated. In this paper, the CG is defined as the origin of the vehicle coordinate system, with the forward direction as the X-axis and the left direction as the Y-axis. The coordinates of the left and right rotation centers of wheel (WCR) of the i-th axle are denoted as $(x_{iL}, y_{iL})$ and $(x_{iR}, y_{iR})$, respectively. The vehicle coordinate system and the relative position between ICR and WCR is illustrated in Fig.\ref{fig:icrPosition}.

The coordinates of ICR relative to WCR, $(B_{iL}, C_{iL})$ and $(B_{iR}, C_{iR})$,  are defined as follows:
\begin{subequations}
\begin{align}
B_{iL} &= x_{iL} - B_0 \\
B_{iR} &= x_{iR} - B_0 \\
C_{iL} &= C_0 - y_{iL} \\
C_{iR} &= C_0 - y_{iR}.
\end{align}
\end{subequations}

Based on the geometric constraints shown in Fig.\ref{fig:icrPosition}, the steering angle $\delta_{i*}$ of each wheel can be expressed as follows:
\begin{equation}\label{eq:CalDelta}
\delta_{i*} = \arctan \left( \frac {B_{i*}} {C_{i*}}   \right) = \arctan \left( \frac {x_{i*} - B_0} {C_0 - y_{i*}}   \right)
\end{equation}
where * can be L or R, and i can go up to N.

According to the domain of the inverse trigonometric functions, the domain of $(B_0, C_0)$ is defined as follows:

\begin{equation}
\{ (B_0, C_0) |C_0 \not = y_{i*}, \  B_0 \in \mathbb{R}\}.
\end{equation}

When \( C_0 = y_{i*} \), the inverse trigonometric functions are undefined, allowing the steering angle of the wheel to theoretically take on any value. 

When the ICR is located on the extension of the wheel's Y-coordinate, the limit of the wheel's steering angle can be expressed as:
\begin{subequations}
\begin{align}
\lim_{C_0 \to y_{i*}-} \delta_{i*} &= \frac{\pi \,\mathrm{signIm}\left(B_0 \,\imath-x_{i*} \,\imath\right)}{2} \\
\lim_{C_0 \to y_{i*}+} \delta_{i*} &= -\frac{\pi \,\mathrm{signIm}\left(B_0 \,\imath-x_{i*} \,\imath\right)}{2}.
\end{align}
\end{subequations}

Therefore, from the perspective of completeness, this paper presents the following definitions for the steering angle of the wheel when \( C_0 = y_{i*} \).

When \( C_0 \) changes from \( <y_{i*} \) to \( y_{i*} \), the value of \( \delta_{i*} \) is:
\begin{equation}
\delta_{i*} = \left\{
\begin{array}{ll}
  -\pi/2 &  \text{if } \  B_0 > x_{i*} \\
  0 & \text{if }\  B_0 = x_{i*}\\
  \pi/2 & \text{if }\  B_0 < x_{i*}
\end{array}
\right. .
\end{equation}

When \( C_0 \) changes from \( >y_{i*} \) to \( y_{i*} \), the value of \( \delta_{i*} \) is:
\begin{equation}
\delta_{i*} = \left\{
\begin{array}{ll}
  \pi/2 &  \text{if } \  B_0 > x_{i*} \\
  0 & \text{if }\  B_0 = x_{i*}\\
  -\pi/2 & \text{if }\  B_0 < x_{i*}
\end{array}
\right. .
\end{equation}

It's important to note that the situations where the ICR is located near the WCR should be avoided in engineering applications.

\begin{figure}[bt]
  \centering
  \subcaptionbox{}[0.2\textwidth]{
      \includegraphics[width=0.2\textwidth]{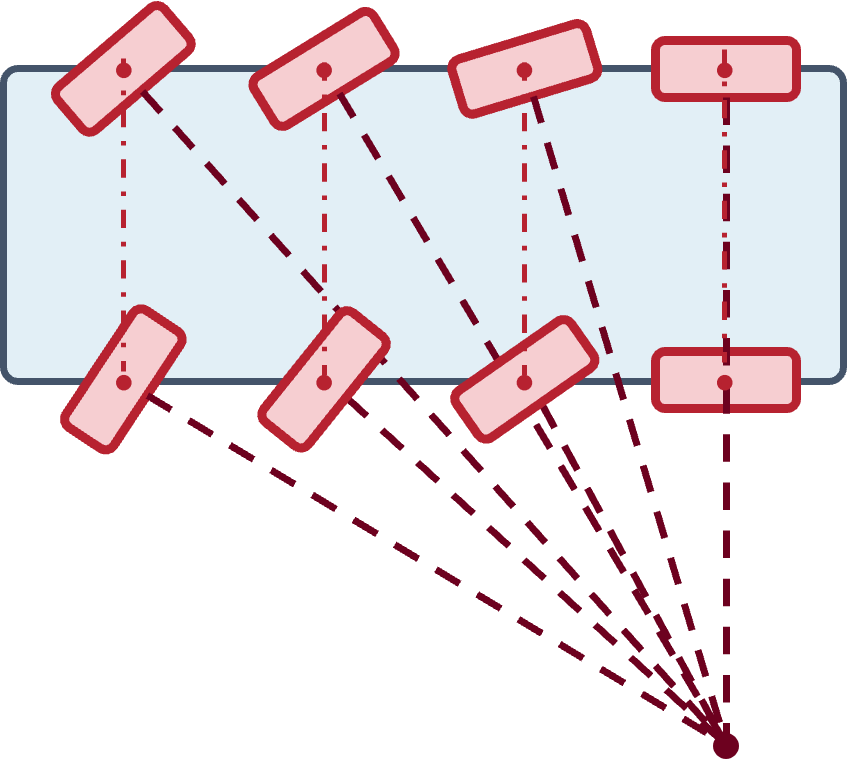} 
  }
  \hspace{2pt}
  \subcaptionbox{}[0.2\textwidth]{
      \includegraphics[width=0.2\textwidth]{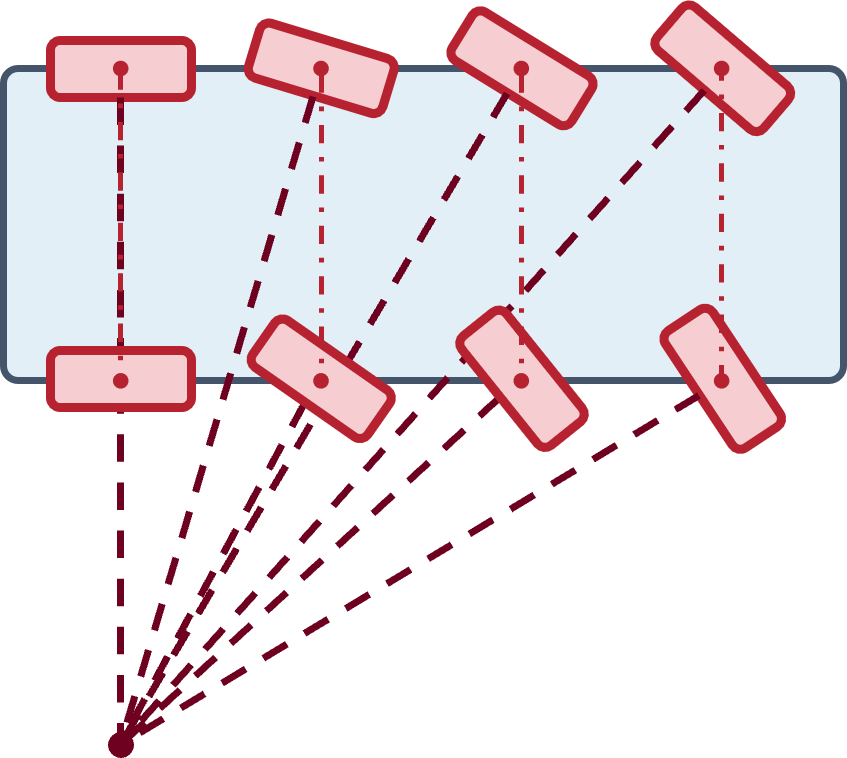}
  }
  \hspace{2pt}
  \subcaptionbox{}[0.2\textwidth]{
      \includegraphics[width=0.2\textwidth]{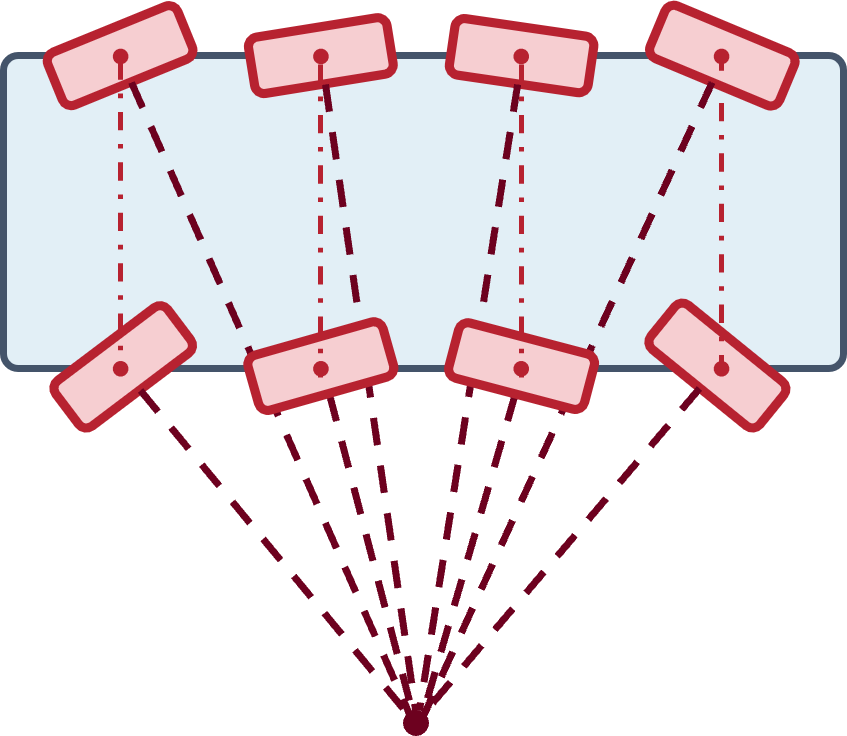}
  }
  
  
  \subcaptionbox{}[0.2\textwidth]{
      \includegraphics[width=0.2\textwidth]{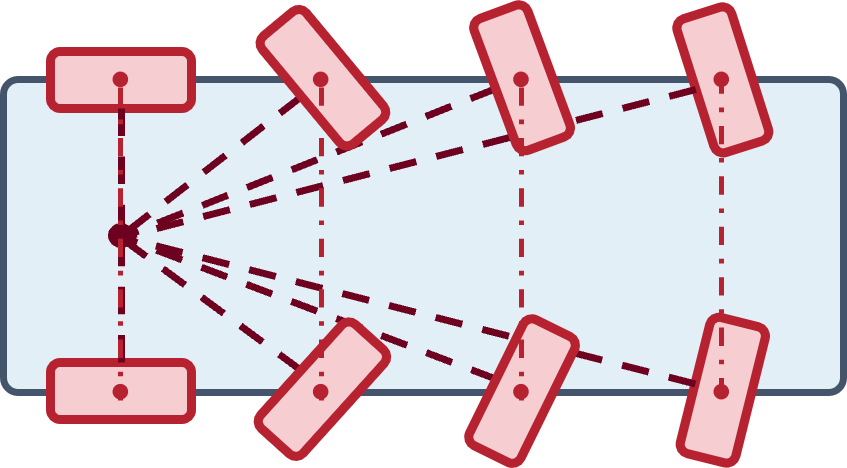}
  }
  \hspace{2pt}
  \subcaptionbox{}[0.2\textwidth]{
      \includegraphics[width=0.2\textwidth]{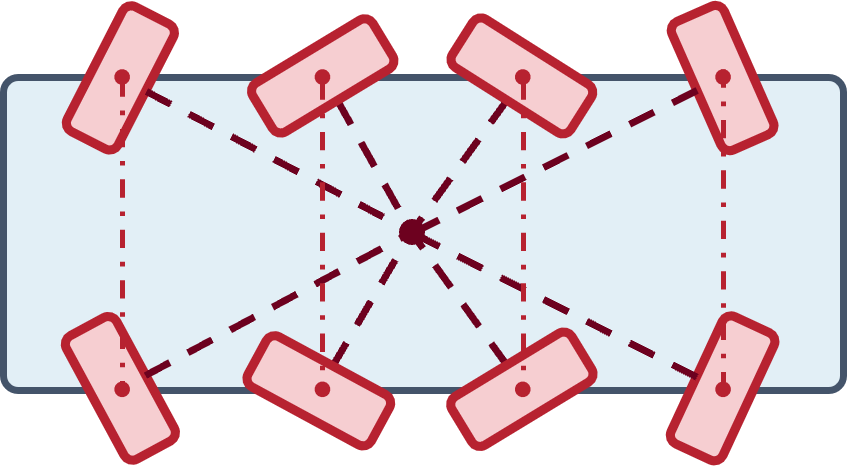}
  }
  \hspace{2pt}
  \subcaptionbox{}[0.2\textwidth]{
      \includegraphics[width=0.2\textwidth]{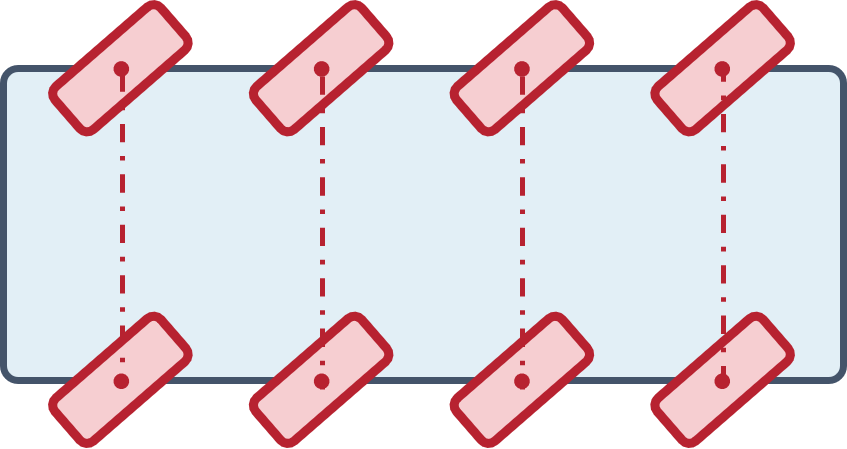}
  }

  \subcaptionbox{\label{fig:RltnBtwnMM}}[0.6\textwidth]{
      \includegraphics[width=0.6\textwidth]{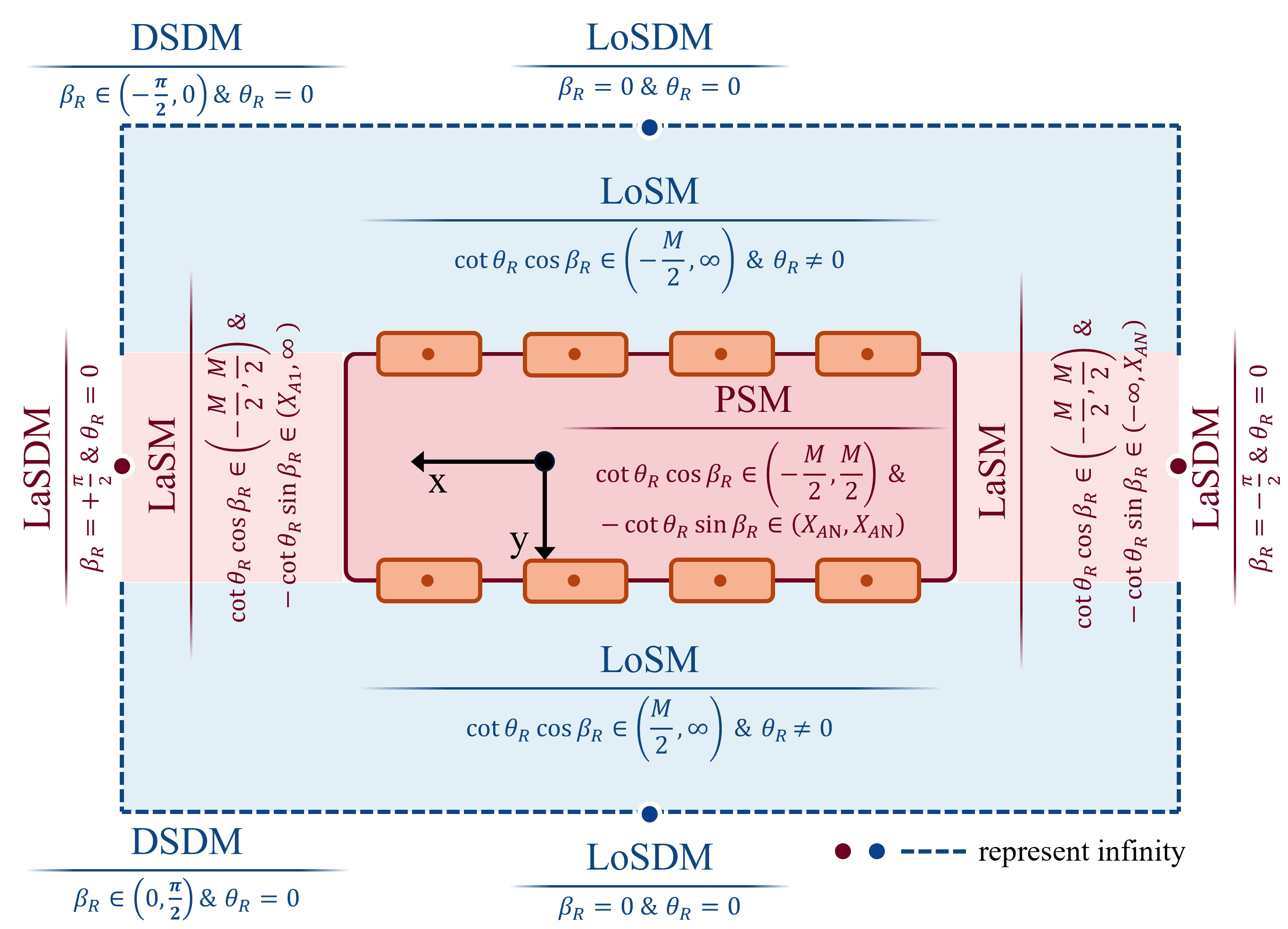}
  }
  
  \caption{Conventional motion modes of AWOISV and definitions proposed in this paper. (a)FASM, (b)RASM, (c)NSM, (d)YSM, (e)ZRSM, (f)DSM, (g)Relationship between motion modes and \( (\theta_R, \beta_R) \).}
  \label{fig:CmmnMtnMd}
\end{figure}


\subsection{Definitions of AWOISV's Motion Modes} \label{Sec:DfntAwoisMM}
The conventional motion modes of AWOISV are listed in Fig.\ref{fig:CmmnMtnMd}. The traditional methods for defining the motion modes of AWOISV lack consistency and are generally categorized in two ways. The former method classifies motion modes based on the relative position of the ICR to the vehicle’s axis, such as normal steering mode (NSM), FASM, and yaw steering mode (YSM). The latter method defines modes based on the steering angles of the wheels, such as DSM and 90-degree Steering Mode.

In traditional motion mode definitions, the position of the ICR is typically fixed at a single point or constrained to a line, rather than moving freely across the entire plane, which significantly limits the maneuverability of AWOISVs. For example, in FASM, the ICR moves along the extended line of the rear axle. In NSM, the ICR is located along the centerline between the vehicle's middle two axles. In YSM, the ICR is fixed at a point outside either the front or rear axle. Due to the inaccurate definition of motion modes for AWOISV, the vehicle can only operate within the selected mode. Switching between different motion modes requires the vehicle to be stationary, which significantly reduces the operational efficiency. To this end, this paper intends to accurately define the motion modes of AWOISV and the switching criteria between them, based on the relationship between the position of the ICR, $(B_0, C_0)$, and each WRC, $(x_{i*}, y_{i*})$.

When either \( B_0 \) or \( C_0 \) approaches infinity, the vehicle operates in a straight driving mode, with the value of \( \arctan \left( \frac{C_{0}}{B_{0}} \right) \) determining the direction of movement. Therefore, based on whether the vehicle's heading angle changes, the motion modes can be categorized into straight driving and steering. The straight driving mode includes longitudinal straight driving mode (LoSDM), lateral straight driving mode (LaSDM), and diagonal straight driving mode (DSDM).

For steering Motion, when the position of the ICR is located along the longitudinal extension of the wheel center, specifically when \( B_0 = y_{i*} \), the corresponding wheel steering angle is 90°. At this point, the wheel has reached its maximum range of motion. From an engineering perspective, the wheel steering angle cannot be increased further. From a mathematical perspective, the wheel angles are discontinuous at \( B_0 \neq y_{i*} \), and any further change in \( B_0 \) will lead to an abrupt transition between +90° and -90°. Therefore, whether \( B_0 \) falls within the range of \((- \frac{M}{2}, \frac{M}{2})\) can be served as one criterion for categorizing motion modes, where \( M \) is the wheel track. Thus, this paper defines the steering mode as lateral steering mode (LaSM) when \( B_0 \) is within \((- \frac{M}{2}, \frac{M}{2})\), and as longitudinal steering mode (LoSM) when \( B_0 \) is outside this range.

According to Eq.\ref{eq:CalDelta}, it can be inferred that in LoSM, the left and right wheel angles must be of the same sign. In contrast, in LaSM, the left and right wheel angles must have opposite signs, meaning one wheel turns right while the other turns left. In LaSM, when the ICR is located within the envelope of the WRC, the vehicle performs a “pivot” motion. Although this pivot motion is not fundamentally different from other lateral steering movements, this paper adopts the traditional definition and revises the lateral steering definition: when \( B_0 \in \left(-\frac{M}{2}, \frac{M}{2}\right) \) and \( C_0 \in \left(X_{AN}, X_{A1}\right) \), where \( X_{Ai} \) represents the X-coordinate of the \( i \)-th axle, the vehicle is considered to be in pivot steering mode (PSM); in all other cases, it is classified as LaSM.

In summary, the motion modes of AWOISV based on \( (B_0, C_0) \)  are defined in the Table.\ref{tab:DfntMtnMd}.

\begin{table}[t]
  \centering
  \caption{Definitions of AWOISV's motion modes}
  \scriptsize
    \begin{tabular}{lllll}
        \toprule
        & $\bm {B_0}$ & $\bm {C_0}$ & $\bm {\theta_R}$ & $\bm {\beta_R}$ \\
        \midrule
        LoSDM & 
        $B_0$ is finite & 
        $C_0 \to \infty \ \text{or} \ C_0 \to -\infty$ & 
        $\theta_R = 0$ & 
        $\beta_R = 0$ \\

        DSDM & 
        $B_0 = \alpha C_0 \ \text{and} \ C_0 \to \infty$ & 
        & 
        $\theta_R = 0$ & 
        $\beta_R \in \left(-\frac{\pi}{2}, \frac{\pi}{2}\right)$ \\

        LaSDM & 
        $B_0 = \infty \ \text{or} \ B_0 = -\infty$ & 
        $C_0$ is finite & 
        $\theta_R = 0$ & 
        $\beta_R = -\frac{\pi}{2} \ \text{or} \ \beta_R = \frac{\pi}{2}$ \\

        LoSM & 
        $B_0 \in \mathbb{R}$ & 
        $C_0 \in \left(-\infty, -\frac{M}{2}\right) \cup \left(\frac{M}{2}, \infty\right)$ & 
        \multicolumn{2}{l}{$\begin{aligned} 
            &\cot \theta_R \cos \beta_R \in \left(-\infty, \frac{M}{2} \right) \cup  \left(\frac{M}{2}, \infty \right) \\
            &\text{and} \ \theta_R \neq 0 
        \end{aligned}$} \\

        LaSM & 
        $B_0 \in \left(-\infty, X_{AN}\right) \cup \left(X_{A1}, \infty\right)$ & 
        $C_0 \in \left(-\frac{M}{2}, \frac{M}{2}\right)$ & 
        \multicolumn{2}{l}{$\begin{aligned} 
            &\cot \theta_R \cos \beta_R \in \left(-\frac{M}{2}, \frac{M}{2}\right) \\
            &\text{and} \ -\cot \theta_R \sin \beta_R \in \left(-\infty, X_{AN}\right) \cup \left(X_{A1}, \infty\right) 
        \end{aligned}$} \\

        PSM & 
        $B_0 \in \left(X_{AN}, X_{A1}\right)$ & 
        $C_0 \in \left(-\frac{M}{2}, \frac{M}{2}\right)$ & 
        \multicolumn{2}{l}{$\begin{aligned}
            &\cot \theta_R \cos \beta_R \in \left(-\frac{M}{2}, \frac{M}{2} \right) \\
            &\text{and} \ -\cot \theta_R \sin \beta_R \in \left(X_{AN}, X_{A1}\right)
            \end{aligned}$} \\
        \bottomrule
    \end{tabular}
    \label{tab:DfntMtnMd}
\end{table}


\subsection{$\theta_R$-$\beta_R $ Methods for AWOISV's Motion Representations}\label{sec:vbrMethod}
Although the \( (B_0, C_0) \) intuitively represents the AWOISV's motion state and provides clear conditions for switching between modes. However, using \( (B_0, C_0) \) directly as control inputs for the AWOISV presents certain challenges. 

Firstly, the domain of \( (B_0, C_0) \) is an infinite set. In contrast, traditional vehicle control methods, such as steering wheels or joysticks, operate within finite sets. The individual wheel angle \( \delta_{i*} \), which directly influence the vehicle’s steering state, are also finite. Using an infinite set as control inputs is overly abstract. Additionally, representing straight driving with \( (B_0, C_0) \) in DSDM requires both \( B_0 \) and \( C_0 \) to be infinite while maintaining a specific proportional relationship, complicating practical implementation. Moreover, using \( (B_0, C_0) \) as control inputs makes switching between left and right turns difficult in both LaSM and LoSM. For instance, in LoSM, transitioning from a left to a right turn necessitates \( B_0 \) moving from \( +\infty \) to \( -\infty \), posing engineering challenges.


To address this, this paper proposes a novel \( \theta_R \)-\(\beta_R \) representation method, replacing \( (B_0, C_0) \) with \( (\theta_R, \beta_R) \). Here, \( \theta_R \) denotes the theoretical steering radius angle, and \( \beta_R \) denotes the theoretical slide slip angle, which is the angle between the direction of the vehicle body and the direction of the vehicle speed. The core idea of this method is to map the vehicle's ICR using two finite angles defined within the range \((- \frac{\pi}{2}, \frac{\pi}{2})\).

First, transform \( (B_0, C_0) \) into \( (R_0, \beta_0) \), where:
\begin{subequations}
\begin{align}
R_0 &= \sqrt{{C_0^2 + B_0^2}} \\
\beta_0 &= \arctan \frac{C_0}{B_0}.
\end{align}
\end{subequations}

Let $\text{sgn} (\theta_R) R_0 = \cot \theta_R$. The mapping relationship between \( (B_0, C_0) \) and \( (\theta_R, \beta_R) \) can be obtained as follows:
\begin{subequations}\label{eq:thetaRDefination}
\begin{align}
B_0 & = -\cot \theta_R \sin \beta_R \\
C_0 & = \cot \theta_R \cos \beta_R \\
\{ (\theta_R, \beta_R) &|\cot \theta_R \cos \beta_R \not = \pm \frac{M}{2} \}.
\end{align}
\end{subequations}

The motion mode definition of AWOISV based on \( (\theta_R, \beta_R) \)  is also shown in the Table.\ref{tab:DfntMtnMd}. According to the definitions, LoSDM, DSDM, and LoSM can transit between each other continuously, as can LaSDM, LaSM, and PSM. For all other modes, transitions can only occur when the vehicle comes to a complete stop. Fig.\ref{fig:RltnBtwnMM} illustrates the relationship between motion modes and \( (\theta_R, \beta_R) \), which corresponds to the ICR. In the diagram, motion modes that can switch between each other are marked with the same color. The area of the regions where the ICR is located under different modes reflects the AWOISV’s ability to adjust its position and orientation in each specific mode. It is evident that LoSM alone can achieve most of the motion states. The FASM, RASM, NSM and partial YSM all belong to LoSM.

\section{Dynamic Modeling and Analysis for AWOISV}\label{sec:mdlAndAnlysis}
\subsection{\(v_x\)-\( v_y \)-\(r\) Dynamic Modeling}
The single-track vehicle model is a classic vehicle dynamics model widely used in conventional vehicle control. It uses the front and rear wheel steering angles, which satisfy the Ackermann steering geometry, as control inputs, and the longitudinal velocity \( v_x \), lateral velocity \( v_y \), and yaw rate \( r \) in the vehicle's coordinate system as state variables. The model is based on two main assumptions: (1) the wheel steering angles are small, implying that the turning radius is sufficiently large; and (2) \( v_y \) is much smaller than \( v_x \), meaning the vehicle's sidesilp angle \( \beta \) is sufficiently small. Based on these assumptions, the following approximations can be made:
\begin{subequations}
\begin{align}
\alpha_{i*} &=\delta_{i*} - \arctan \left( \frac{v_y + x_{i*} r}{v_x - y_{i*} r} \right)\\ 
& \approx \delta_{i*} - \arctan \left( \frac{\beta + x_{i*} r/v_x }{1 - y_{i*} r/v_x} \right)\\
& \approx \delta_{i*} - \beta - \frac{x_{i*} r}{v_x}.
\end{align} 
\end{subequations}

However, these assumptions are only valid under limited conditions within the LoSM of AWOISV, specifically when \( \beta_0 \approx 0 \). For example, in LaSM, the wheel steering angles can exceed 60°, and \( v_y \) can be much larger than \( v_x \), making the single-track vehicle model inadequate for calculating the lateral tire forces. To accurately describe the dynamics of AWOISV under large steering angles and across various modes such as LoSM, LaSM, PSM, and others, a generalized dynamic model for AWOISV based on the \( \theta_R\)-\(\beta_R \) method is proposed in this paper.

\begin{figure}[!t]
  \centering
  \includegraphics[width=0.6\textwidth]{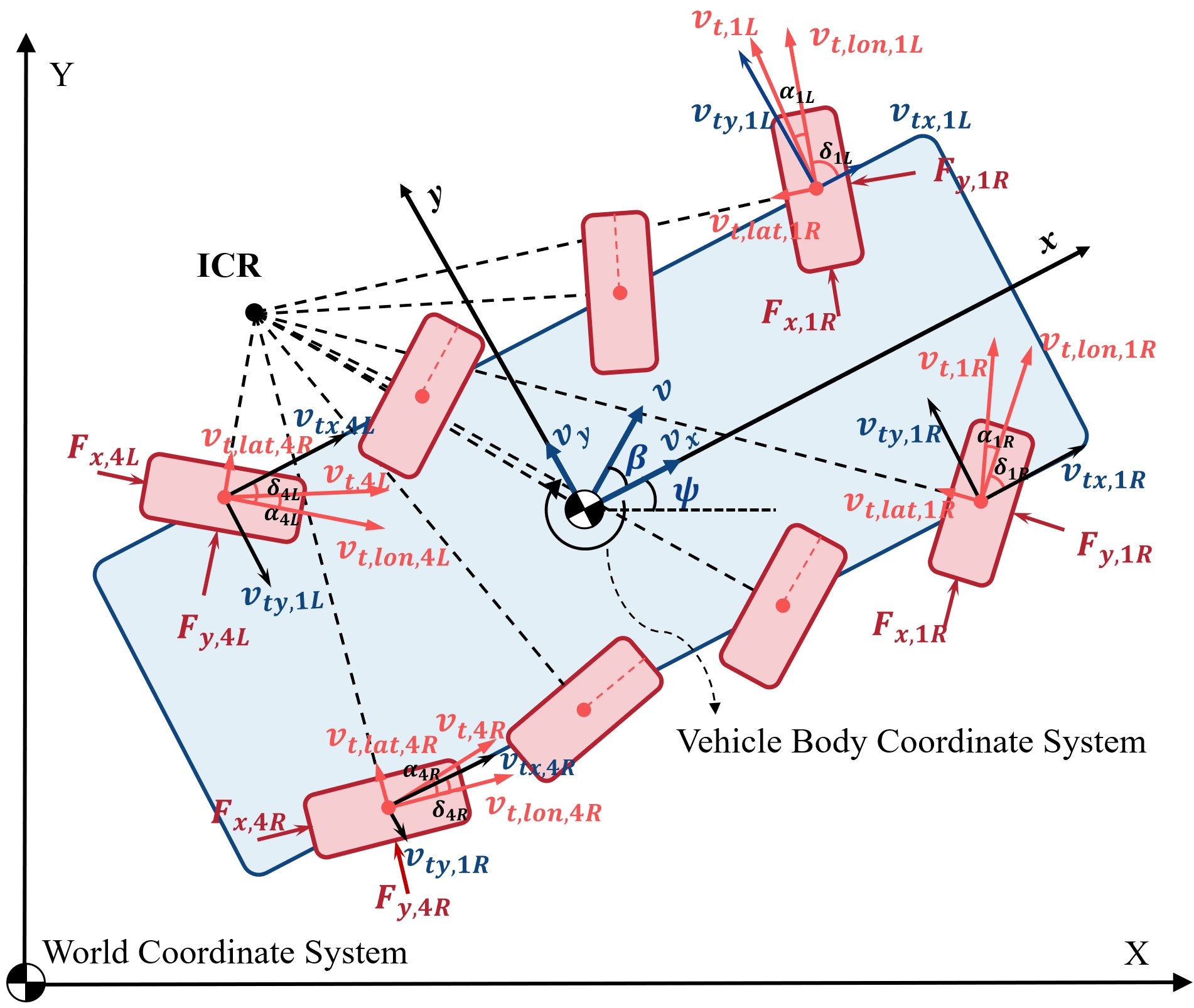}
  \caption{Coordinate system and dynamic model of AWOISV.}
  \label{fig:CrdntSystmDynMdl}
\end{figure}

Without loss of generality, it is assumed that the AWOISV is equipped with \( n \) axles, where \( n \geq 2 \). Based on the tire force directions shown in Fig.\ref{fig:CrdntSystmDynMdl}, the AWOISV's force equations in the vehicle body coordinate system are as follows:
\begin{subequations}
\begin{align}
m \dot{v}_{x}-m r v_{y} &= \sum_{* = L}^{R} \sum_{i =1}^ n \left( F_{x,i*} \cos \delta_{i*} -F_{y,i*} \sin \delta_{i*}\right) \\
m \dot{v}_{y}+m r v_{x} &= \sum_{* = L}^{R} \sum_{i =1}^ n \left( F_{x,i*} \sin \delta_{i*} + F_{y,i*} \cos \delta_{i*}\right) \\
I_{z} \dot{r} &= \sum_{* = L}^{R} \sum_{i=1}^n \left( x_{i*} \left( F_{x,i*} \sin \delta_{i*} + F_{y,i*} \cos \delta_{i*}\right)
\right. \notag \\
& \left. - y_{i*} \left( F_{x,i*} \cos \delta_{i*} -F_{y,i*} \sin \delta_{i*}\right) \right) \\
I_{w,i*} \dot \omega_{i*} &= T_{w, i*} - R_{eff} F_{x, i*} 
\end{align}
\end{subequations}
where \( m \) represents the total vehicle mass, while \( F_{x,i*} \) and \( F_{y,i*} \) denote the longitudinal and lateral tire forces in the tire coordinate system, respectively.

To accurately capture the nonlinear relationship between tire slip angles and lateral forces as tires approach saturation, a nonlinear tire model is required. Several well-established nonlinear tire models have been studied in the literature \citep{wangTireForceDistribution2018}, including the Magic Formula, Dugoff, and UniTire models. As proposed in \citep{liuDynamicModelingControl2018}, the improved Fiala brush model is adopted in this paper to model the lateral tire forces \( F_y \). The fundamental form of this model is expressed as follows:
\begin{equation}\label{eq:lateralFrc}
F_{y}= \overline {C_{\alpha}} \alpha =
\left\{\begin{array}{l}-C_{\alpha} \tan \alpha+\frac{C_{\alpha}^{2}}{3 F_{y \max }}|\tan \alpha| \tan \alpha \\ -\frac{C_{\alpha}^{3}}{27 F_{\max }^{3}} \tan ^{3} \alpha \quad|\alpha|<\alpha_{s l} \\ -F_{y \max } \operatorname{sgn}(\alpha) \quad|\alpha| \geq \alpha_{s l}\end{array}\right.
\end{equation}
where 
\begin{subequations}
\begin{align}
F_{y, \text { max }} &=\sqrt{\left(\mu F_{z}\right)^{2}-F_{x}^{2}} \\
\alpha_{s l} &=\arctan \left(\frac{3 F_{y \max }}{C_{\alpha}}\right).
\end{align}
\end{subequations}

Here, \( C_\alpha \) represents the tire’s lateral stiffness, \( \mu \) denotes the coefficient of friction between the tire and the road surface, and \( F_z \) stands for the vertical tire force.

According to the relative coordinate relationship between the tire and the vehicle body, the lateral and longitudinal tire velocities in the tire coordinate system, \( v_{t,lat,i*} \) and \( v_{t,lon,i*} \), can be determined as follows:
\begin{subequations}
\begin{align}
v_{t,lat,i*} &= v_{tx,{i*}} \cos \delta_{i*} + v_{ty,{i*}} \sin \delta_{i*} \notag \\
& = \left({v_{x}-y_{i*} r}\right) \cos \delta_{i*} + \left({v_{y}+x_{i*} r}\right) \sin \delta_{i*} \\
v_{t,lon,i*} &= v_{tx,{i*}} \sin \delta_{i*} + v_{ty,{i*}} \cos \delta_{i*} \notag \\
& = -\left({v_{x}-y_{i*} r}\right) \sin \delta_{i*} + \left({v_{y}+x_{i*} r}\right) \cos \delta_{i*}
\end{align}
\end{subequations}
where \( v_{tx,{i*}} \) and \( v_{ty,{i*}} \) represent the longitudinal and lateral velocities of the tire in the vehicle's coordinate system, respectively.

Assuming the tire slip angles for all wheels are small, they can be expressed as follows:
\begin{equation}
\begin{aligned}
\alpha_{i*} &= \frac{v_{t,lat,i*}}{v_{t,lon,i*}} \\
&= \frac{\left({v_{x}-y_{i*} r}\right) \cos \delta_{i*} + \left({v_{y}+x_{i*} r}\right) \sin \delta_{i*}}{-\left({v_{x}-y_{i*} r}\right) \sin \delta_{i*} + \left({v_{y}+x_{i*} r}\right) \cos \delta_{i*}}.
\end{aligned}
\end{equation}

\begin{subequations}\label{eq:VxVyRMdl}
  \begin{align}
  \dot v_x &= \frac 1 {m} \sum_{* = L}^{R} \sum_{i = 1}^{N}  \left( 
  \varXi_{i*} F_{x,i*}
  - \varTheta_{i*} \overline {C_{\alpha, i*}}
  \left(
  \frac{\left(v_{y}+r x_{i*}\right)+\left(v_{x}-r y_{i*} \varTheta_{i*}\right)}
  {\left(v_{x}-r y_{i*}\right)-\left(v_{y}+r x_{i*}\varTheta_{i*}\right)}
  \right)
  \right) + r v_y\\
  \dot v_y &= \frac 1 {m} \sum_{* = L}^{R} \sum_{i = 1}^{N}  \left( \varTheta_{i*} 
  F_{x,i*}
  + \varXi_{i*} \overline {C_{\alpha, i*}} 
  \left(
  \frac{\left(v_{y}+r x_{i*}\right)+\left(v_{x}-r y_{i*} \varTheta_{i*}\right)}
  {\left(v_{x}-r y_{i*}\right)-\left(v_{y}+r x_{i*}\varTheta_{i*}\right)}
  \right)
  \right) - r v_x\\
  \dot r &= \frac 1 {I_Z} \sum_{* = L}^{R} \sum_{i = 1}^{N} 
  \left(
  x_{i*}
  \left(
  \varTheta_{i*} F_{x,i*}
  + \varXi_{i*} \overline {C_{\alpha, i*}} 
  \left(
  \frac{\left(v_{y}+r x_{i*}\right)+\left(v_{x}-r y_{i*} \varTheta_{i*}\right)}
  {\left(v_{x}-r y_{i*}\right)-\left(v_{y}+r x_{i*}\varTheta_{i*}\right)}
  \right)
  \right) 
  \right. \\ &\quad \left. \notag    
  - y_{i*}
  \left(
  \varXi_{i*} F_{x,i*}
  - \overline {C_{\alpha, i*}} \varTheta_{i*}
  \left(
  \varTheta_{i*} \frac{\left(v_{y}+r x_{i*}\right)+\left(v_{x}-r y_{i*}\right)}
  {\left(v_{x}-r y_{i*}\right)-\left(v_{y}+r x_{i*}\varTheta_{i*}\right)}
  \right)
  \right)
  \right).
  \end{align}
\end{subequations}

By combining Eq. 1, Eq. 2, and Eq. 3, the AWOISV dynamic model is derived, treating the longitudinal velocity \( v_x \), lateral velocity \( v_y \), and yaw rate \( r \) as state variables, and the theoretical steering radius angle \( \theta_R \) and the theoretical slide slip angle \( \beta_R \) as control inputs. The resulting model is shown in Eq.\ref{eq:VxVyRMdl}:

Here, 
\begin{subequations}\label{eq:VxVyRMdlDscrptn}
\begin{align}
\sigma_{i*} &= \frac{x_{i*} + \cot \theta_R \sin \beta_R}{y_{i*} - \cot \theta_R \cos \beta_R} \\
\varTheta_{i*} &= \frac {\sigma_{i*}} {\sqrt{1 + \sigma_{i*}^2}} \\
\varXi_{i*} &= \frac {1} {\sqrt{1 + \sigma_{i*}^2}}.
\end{align}
\end{subequations}

\subsection{\(v\)-\( \beta \)-\(r\) Dynamic Modeling}
The terms \( \varTheta_{i*} \) and \( \varXi_{i*} \) in Eq.\ref{eq:VxVyRMdlDscrptn} essentially represent the sine and cosine of the steering angles for each wheel, respectively, with values ranging between \([-1, 1]\). Substituting these into Eq.\ref{eq:VxVyRMdl} reveals that both lateral and longitudinal tire forces equally affect the vehicle’s lateral and longitudinal motions in the body coordinate system. This shows that it is invalid to assume a constant longitudinal velocity in the AWOISV dynamic model, and incorrect to attribute changes in lateral velocity solely to lateral tire forces. The DSDM mode clearly demonstrates this. For example, in a 45°-DSDM's acceleration condition, the longitudinal/lateral velocities and accelerations are equal under ideal conditions, and the lateral tire forces are zero. As a result, both \( \dot{v}_x \) and \( \dot{v}_y \) are entirely due to \( F_{x,i*} \).

Although Eq.\ref{eq:VxVyRMdl} can describe the motion behavior of the AWOISV, it is not intuitive for control purposes. In vehicle dynamic control, especially for path tracking, more attention is given to the vehicle's forward speed, the angle between the vehicle’s heading and its motion direction, and the position changes perpendicular to the direction of motion. Therefore, in this paper, a novel \(v\)-\( \beta \)-\(r\) dynamic model is developed by replacing \(v_x\) and \(v_y\) in Eq. 1 with the forward velocity \(v\) and the sideslip angle \( \beta \).

The relationships between the longitudinal velocity \(v_x\), lateral velocity \(v_y\), forward velocity \(v\), and the sideslip angle \( \beta \) can be expressed as follows:
\begin{subequations}\label{eq:RltnspVxVyVBeta}
\begin{align}
v_x &= v \cos \beta \\
v_y &= v \sin \beta .
\end{align}
\end{subequations}

By differentiating Eq.\ref{eq:RltnspVxVyVBeta} and substituting the yaw rate \(r\), the following expression is obtained:
\begin{subequations}\label{eq:RltnspVxVyVBeta2}
\begin{align}
\dot v_x - r v_y &= \dot v \cos \beta - \dot \beta v \sin \beta - r v \sin \beta\\
\dot v_y + r v_x &= \dot v \sin \beta + \dot \beta v \cos \beta + r v \cos \beta .
\end{align}
\end{subequations}

By multiplying Eq.\ref{eq:RltnspVxVyVBeta2} by \( \cos\beta \) and \( \sin\beta \), and then adding and subtracting the resulting equations, the state equations for the forward velocity \(v\) and the sideslip angle \( \beta \) are derived as follows:
\begin{subequations}\label{eq:StEqVBeta}
\begin{align}
\dot v &= \cos \beta (\dot v_x - r v_y) + \sin \beta (\dot v_y + r v_x) \\
\dot \beta &= \frac 1 v \left( -\sin \beta (\dot v_x - r v_y) + \cos \beta (\dot v_y + r v_x) \right) - r .
\end{align}
\end{subequations}

By combining Eq.\ref{eq:RltnspVxVyVBeta} and Eq.\ref{eq:StEqVBeta}, and then merging the terms for the longitudinal and lateral tire forces, the generalized \(v\)-\( \beta \)-\(r\) dynamic model of the AWOISV is derived, as shown in Eq.\ref{eq:VBRMdl}:

{\scriptsize
\begin{subequations}\label{eq:VBRMdl}
\begin{align}
\dot v &= \frac 1 m \sum_{* = L}^{R} \sum_{i = 1}^{N}
\left(
\left(-\varTheta_{i*} \cos \beta + \varXi_{i*} \sin \beta  \right) 
\right. \\ &\quad \left. \notag
\overline {C_{\alpha, i*}} 
\left(
\frac{\left(v \sin \beta+r x_{i*}\right)+\left(v \cos \beta-r y_{i*}\right) \varTheta_{i*}}
{\left(v \cos \beta-r y_{i*}\right)-\left(v \sin \beta+r x_{i*}\right) \varTheta_{i*}}
\right)
+
\left( \varXi_{i*} \cos \beta + \varTheta_{i*} \sin \beta \right)F_{x,i*}
\right)\\
\dot \beta &= \frac 1 {m v} \sum_{* = L}^{R} \sum_{i = 1}^{N}
\left(
\left( \varXi_{i*} \cos \beta + \varTheta_{i*} \sin \beta  \right) 
\right. \\ &\quad \left. \notag
\overline {C_{\alpha, i*}} 
\left(
\frac{\left(v \sin \beta+r x_{i*}\right)+\left(v \cos \beta-r y_{i*}\right) \varTheta_{i*}}
{\left(v \cos \beta-r y_{i*}\right)-\left(v \sin \beta+r x_{i*}\right) \varTheta_{i*}}
\right)  
+
\left( \varTheta_{i*} \cos \beta - \varXi_{i*} \sin \beta \right)F_{x,i*}
\right) - r\\ 
\dot r &= \frac 1 {I_Z} \sum_{* = L}^{R} \sum_{i = 1}^{N}
\left(
\left( \varXi_{i*} x_{i*} + \varTheta_{i*} y_{i*}  \right) 
\overline {C_{\alpha, i*}} 
\left(
\frac{\left(v \sin \beta+r x_{i*}\right)+\left(v \cos \beta-r y_{i*}\right) \varTheta_{i*}}
{\left(v \cos \beta-r y_{i*}\right)-\left(v \sin \beta+r x_{i*}\right) \varTheta_{i*}}
\right)  
+
\left( \varTheta_{i*} x_{i*} - \varXi_{i*} y_{i*} \right)F_{x,i*}
\right).
\end{align}
\end{subequations}
}

\subsection{Dynamic Characteristic Analysis}\label{sec:dynCharAnalysis}

\begin{figure}[bt]
\centering
  \subcaptionbox{}[0.3\textwidth]{
      \includegraphics[width=0.3\textwidth]{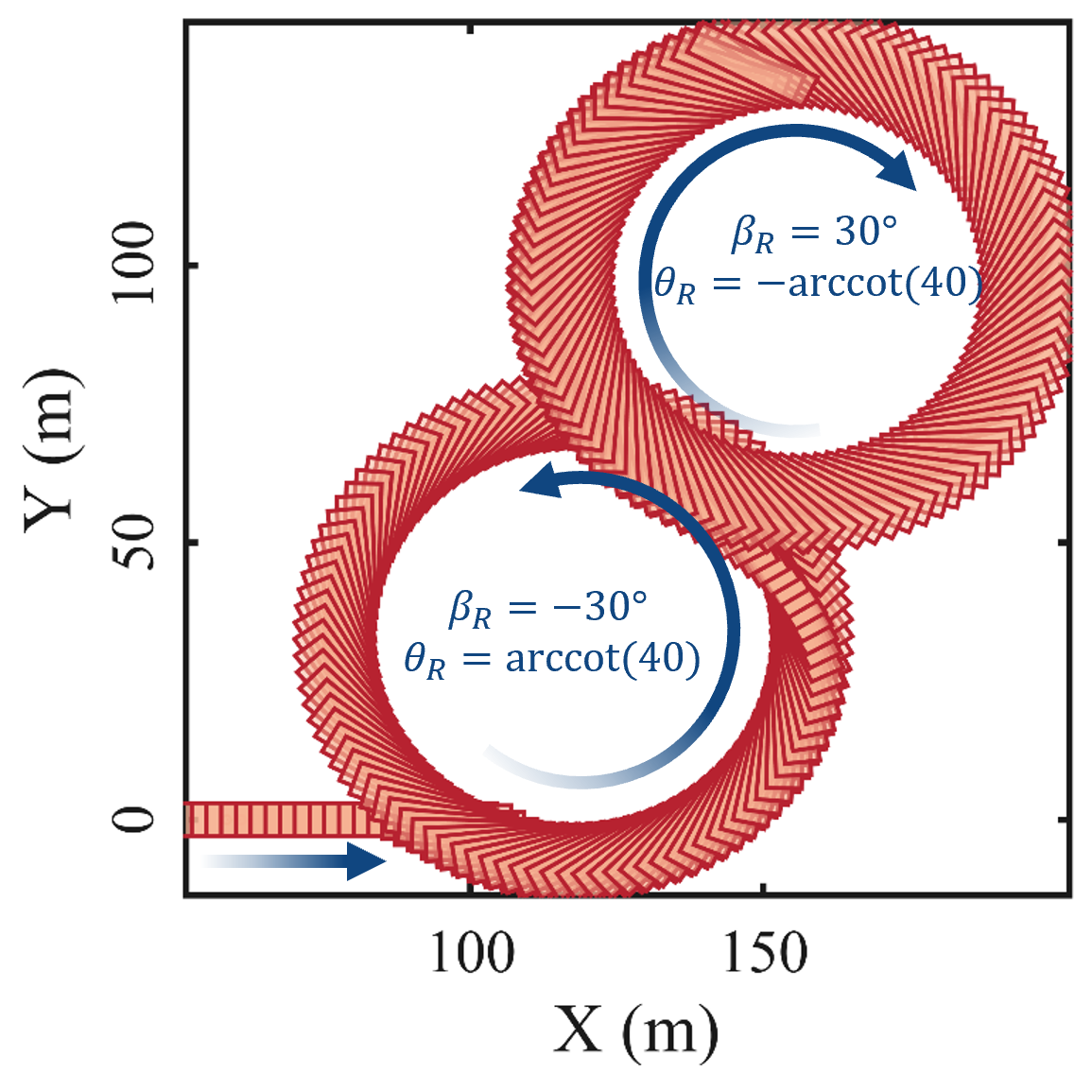} 
  }
  \hspace{2pt}
  \subcaptionbox{}[0.3\textwidth]{
      \includegraphics[width=0.3\textwidth]{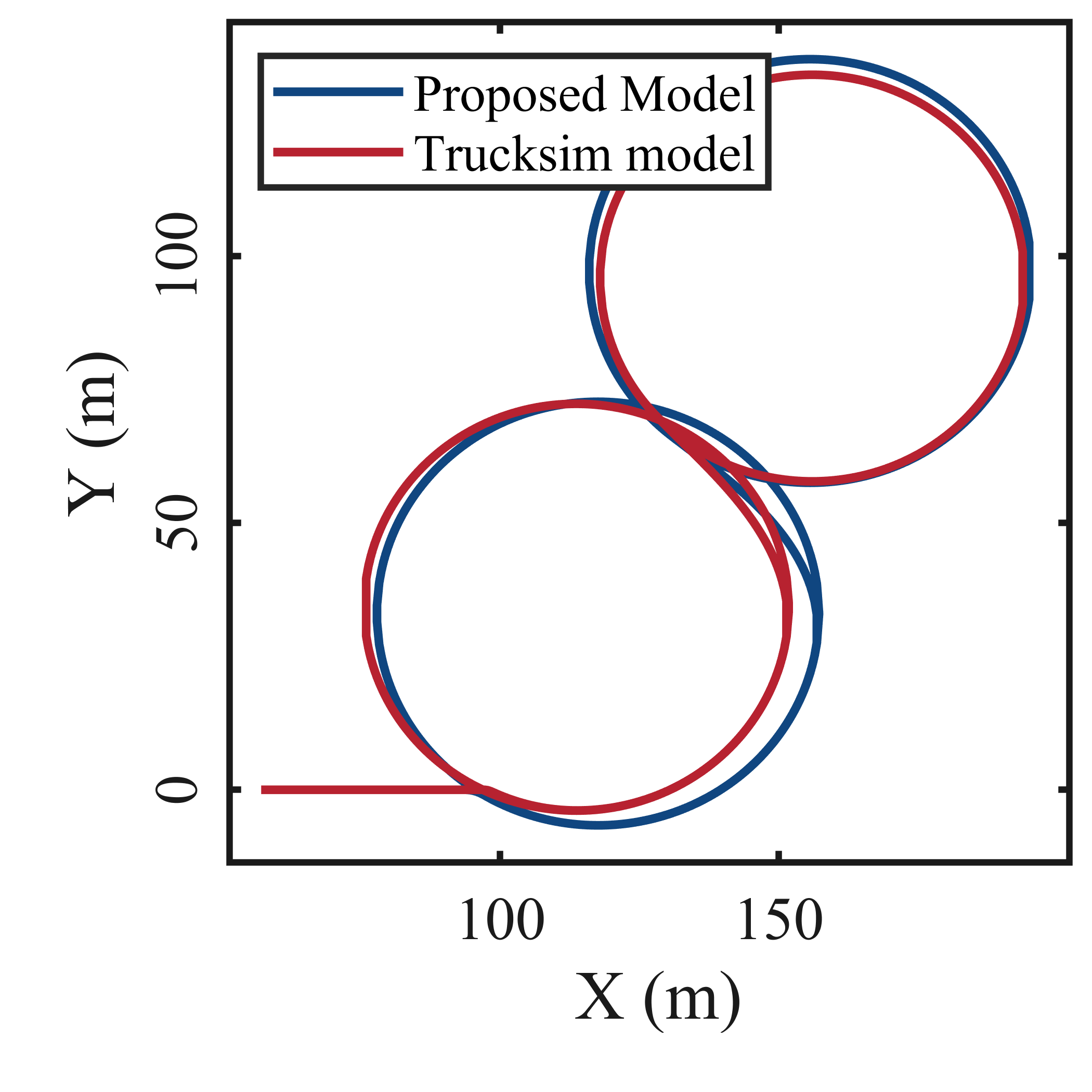}
  }

\subcaptionbox{}[0.6\textwidth]{
    \includegraphics[width=0.6\textwidth]{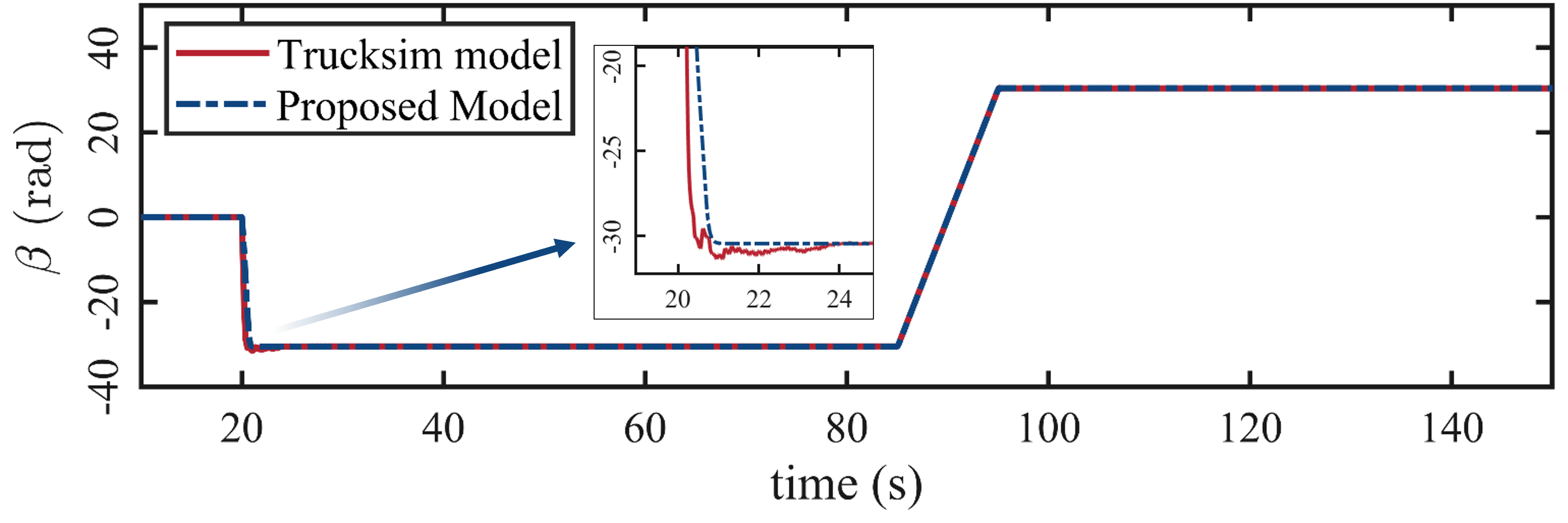} 
}

\subcaptionbox{}[0.6\textwidth]{
    \includegraphics[width=0.6\textwidth]{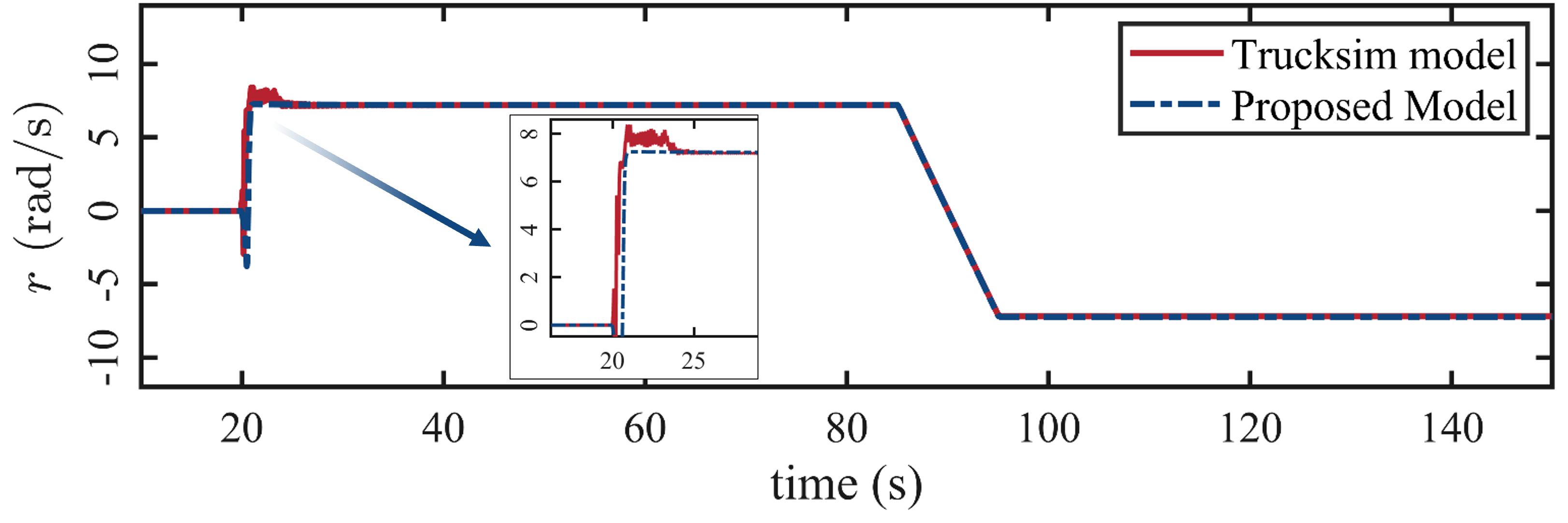} 
}
\caption{Comparison between the \( v\)-\( \beta \)-\( r \) dynamic model and Trucksim. (a)Target trajectory, (b)Trajectory comparison, (c)Sidesilp angle comparison, (d)Yaw velocity comparison.}
\label{fig:ValidationTraj}
\end{figure}

To validate the accuracy of the proposed \(v\)-\( \beta \)-\(r\) dynamic model and analyze the dynamic characteristics, a 4-axle AWOISV is selected as the analysis subject. The parameter settings are shown in Table.\ref{tab:4AxleAwoisvParm}. A comparison is conducted between the established model and the Trucksim model. The comparison scenario involves a figure-eight maneuver, as shown in the Fig.\ref{fig:ValidationTraj}. The forward velocity \( v \) is set to 15 m/s, with a theoretical steering radius of 40 meters and a theoretical sidesilp angle of 30°. The theoretical sidesilp angle $\beta$ is chosen to fully reflect the high maneuverability of the AWOISV. 

Fig.\ref{fig:ValidationTraj} shows that the trajectory and yaw rate of the proposed dynamic model match well with the Trucksim model at low and medium speeds. AWOISVs, exemplified by the SMPT, primarily operate in closed environments such as test tracks or industrial sites. Due to their large mass and the need for significant posture adjustments, their maximum speed is usually below 15 km/h. Therefore, it can be concluded that the proposed \( v \)-\( \beta \)-\( r \) dynamic model can effectively guide the design of path tracking control strategies.

\begin{table}[t]
  \centering
  \caption{AWOISV Parameters}
  \label{tab:4AxleAwoisvParm}
  \small
  \begin{tabular}{l c c c}
  \toprule
  \textbf{Parameter} & \textbf{Symbol} & \textbf{Unit} & \textbf{Value} \\
  \midrule
  Vehicle Mass & $m$ & kg & 10,000 \\
  Vehicle Inertia & $I_Z$ & $\text{kg} \cdot \text{m}^2$ & 34,823 \\
  Wheel Steering Range & $[\delta_{\min}, \delta_{\max}]$ & rad & $[-\pi/2, \pi/2]$ \\
  Tire Cornering Stiffness & $C_\alpha$ & N/rad & 40,000 \\
  Friction Coefficient & $\mu$ & - & 0.85 \\
  Tire Rolling Radius & $R_{eff}$ & m & 0.35 \\
  Axle-to-CG X Distance & $X_{A1} \sim X_{A4}$ & m & \parbox{2cm}{$[3.55, 1.75,$\\ $-1.75, -3.55]$} \\
  Wheel Track & $M$ & m & 2.72 \\
  Height of CG & $H_{CG}$ & m & 1.21 \\
  \bottomrule
  \end{tabular}
\end{table}

\begin{figure}[!t]
\centering
  \subcaptionbox{\label{fig:VehTraj}}[0.6\textwidth]{
    \includegraphics[width=0.6\textwidth]{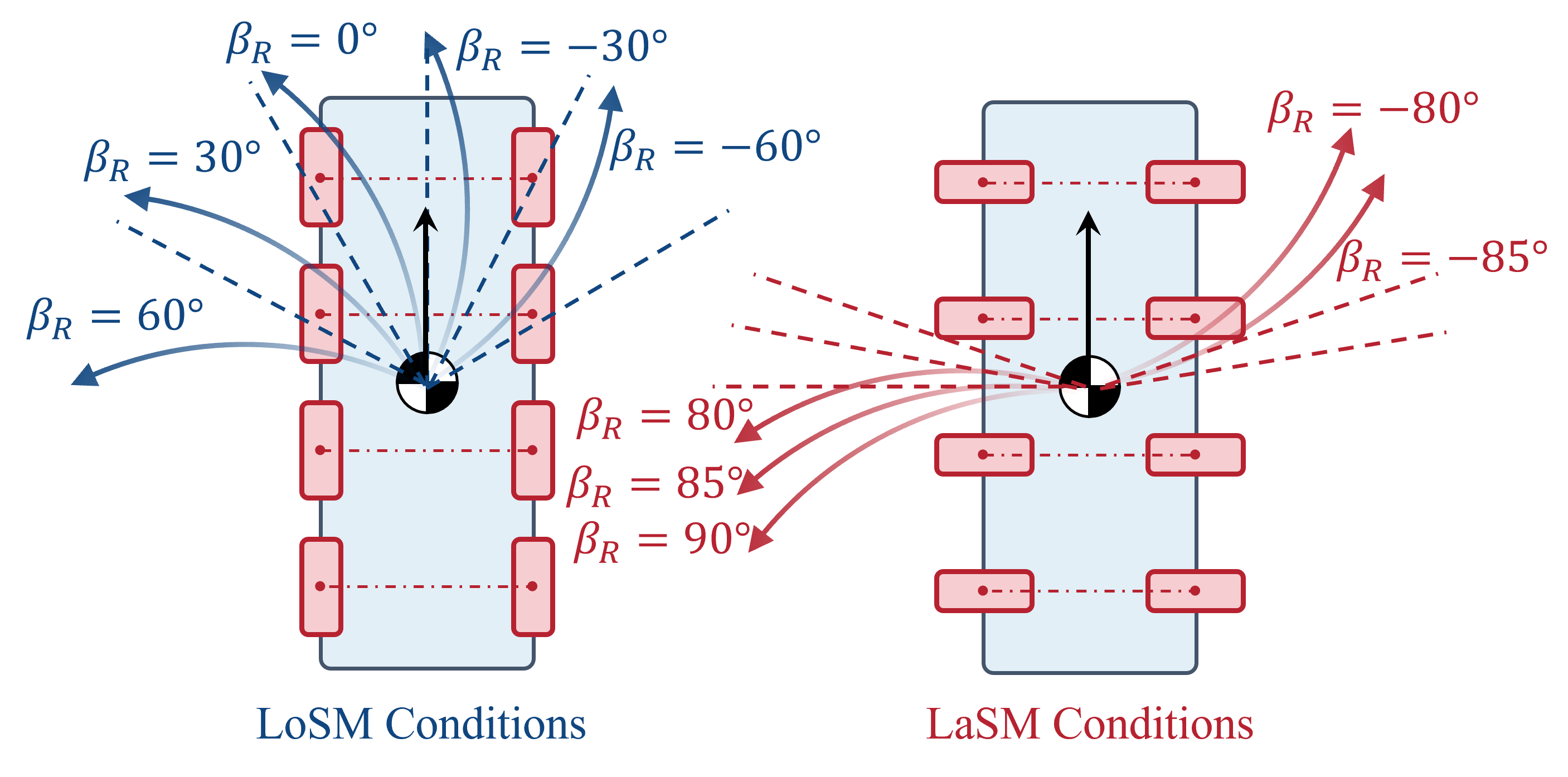}
    
  }

  \subcaptionbox{\label{fig:DynCharLoSM_A}}[0.3\textwidth]{
    \includegraphics[width=0.3\textwidth]{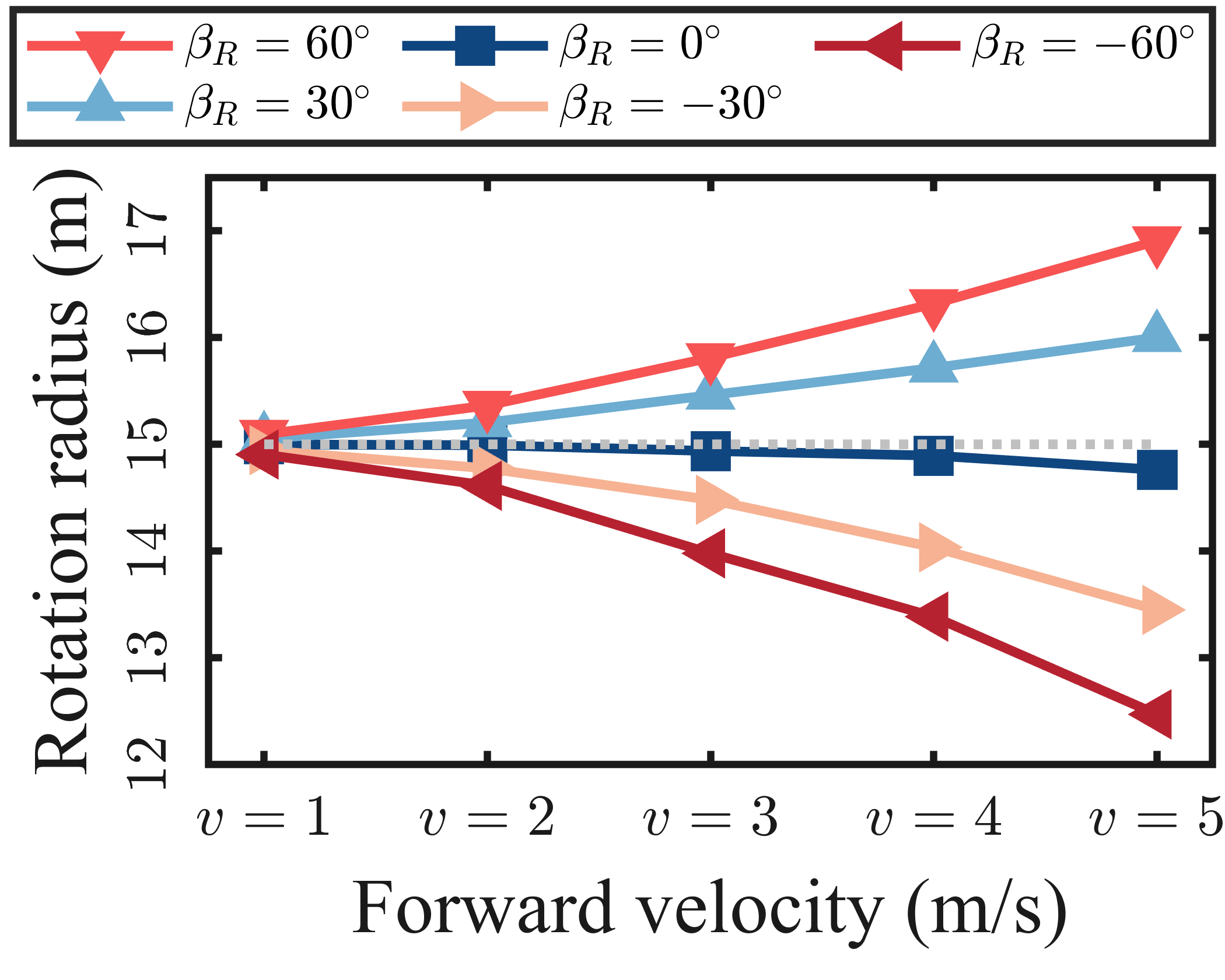}
    
  }
  \hspace{2pt}
  \subcaptionbox{\label{fig:DynCharLoSM_B}}[0.3\textwidth]{
    \includegraphics[width=0.3\textwidth]{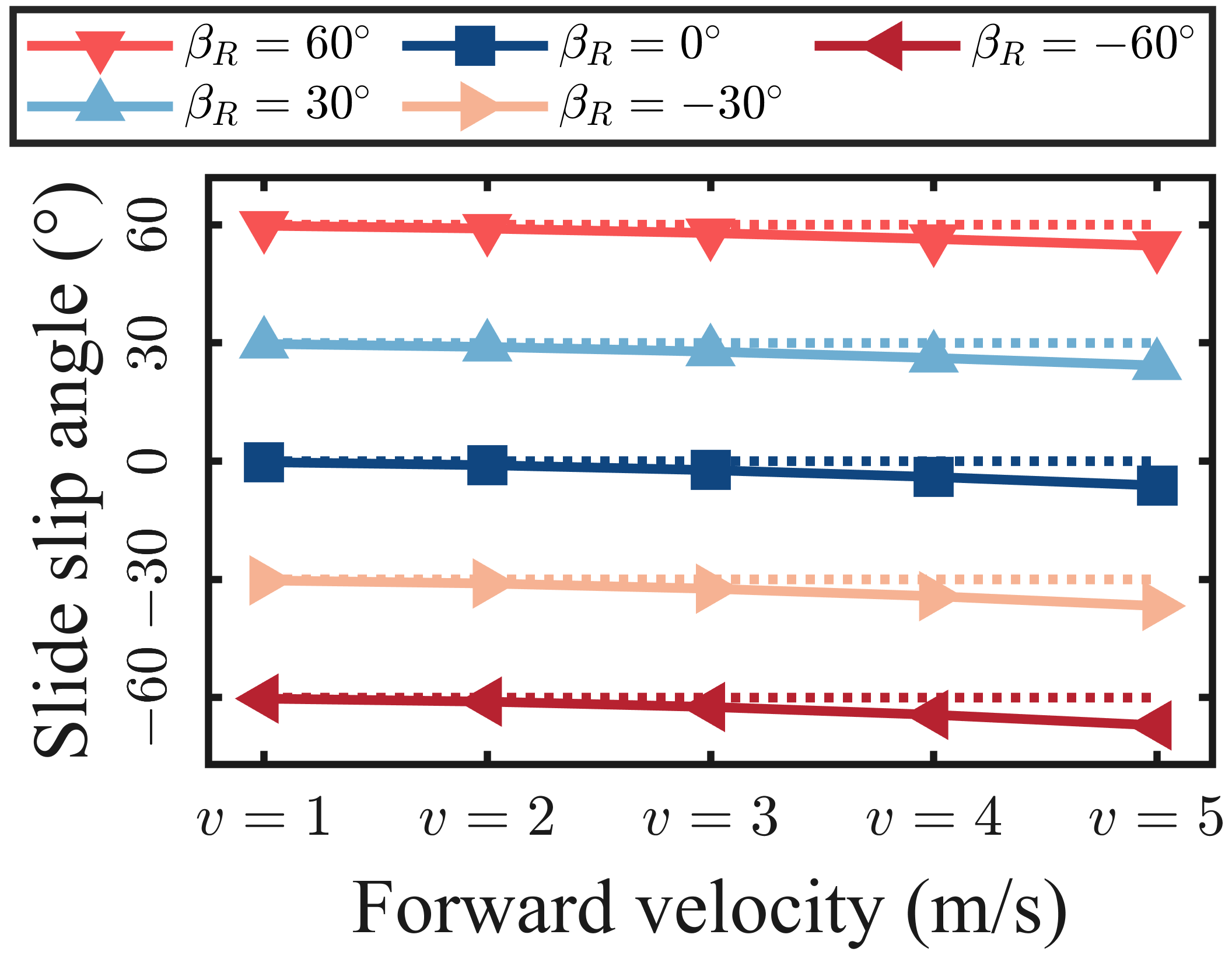}
    
  }

  \subcaptionbox{\label{fig:DynCharLaSM_A}}[0.3\textwidth]{
    \includegraphics[width=0.3\textwidth]{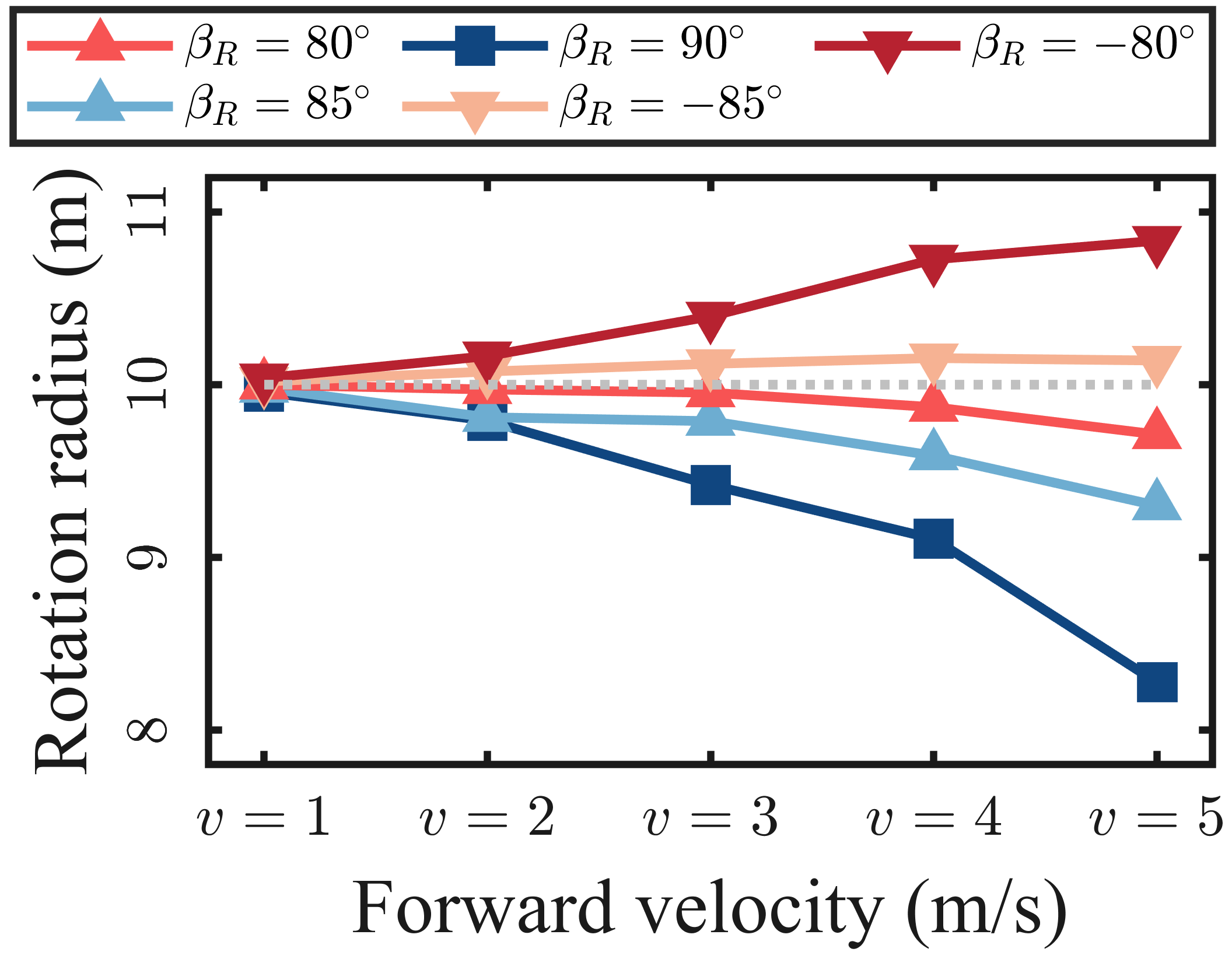}
    
  }
  \hspace{2pt}
  \subcaptionbox{\label{fig:DynCharLaSM_B}}[0.3\textwidth]{
    \includegraphics[width=0.3\textwidth]{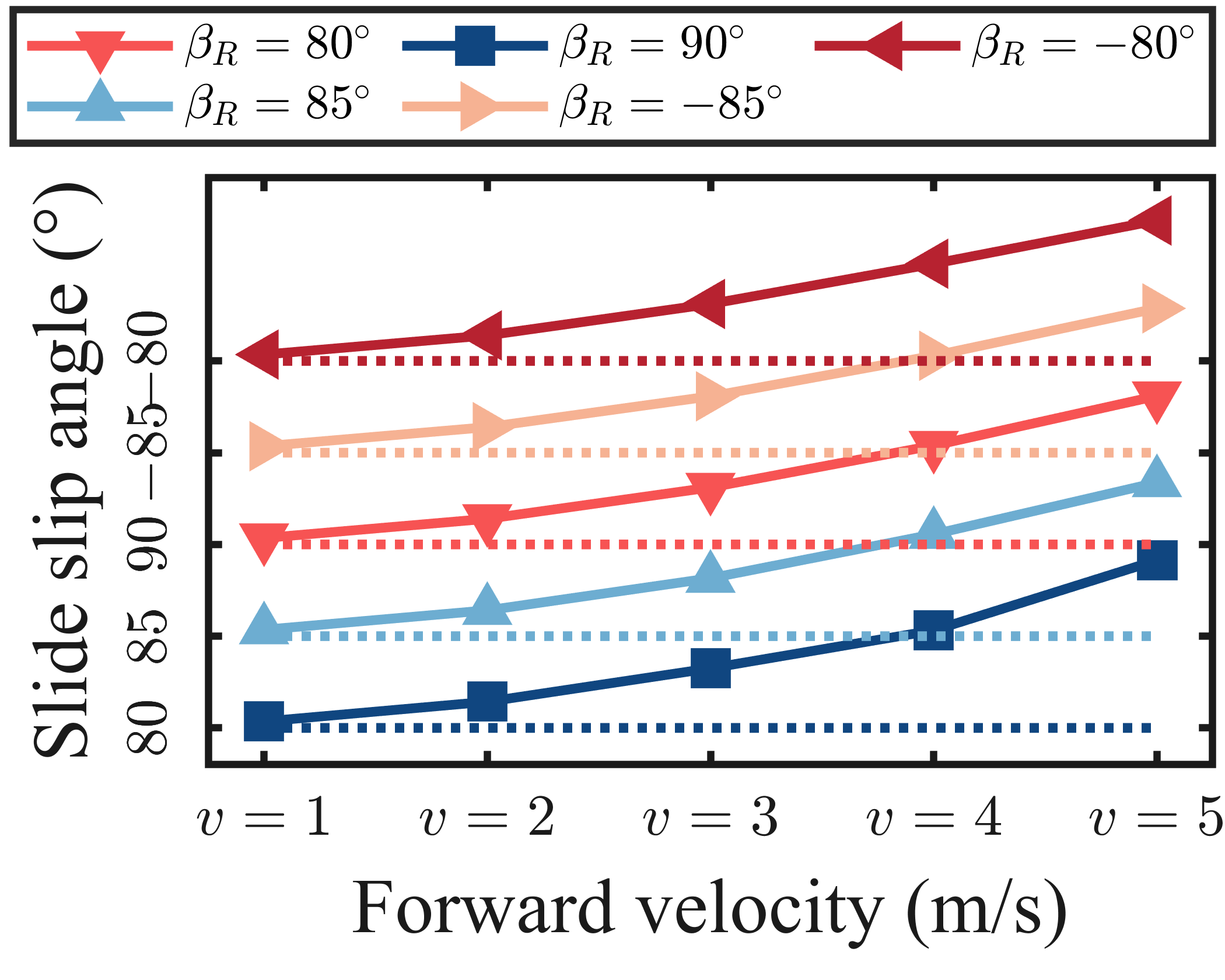}
    
  }
\caption{Dynamic characteristics of LoSM and LaSM. (a)AWOISV trajectory with LoSM and LaSM, (b)$R$ vs $v$ in LoSM, (c)$\beta$ vs $v$ in LoSM, (d)$R$ vs $v$ in LaSM, (e)$\beta$ vs $v$ in LaSM.}
\end{figure}

According to Section \ref{Sec:DfntAwoisMM}, the AWOISV encompasses 6 motion modes: LoDSM, DSDM, LaDSM, LoSM, LaSM, and PSM. In LoSM, LaSM, and PSM, the vehicle undergoes rotational motion, resulting in a yaw rate and tire slip angles. This leads to discrepancies between the actual and ideal motion trajectories. Since there is no fundamental difference between LaSM and PSM, this paper focuses on the dynamic analysis of LoSM and LaSM. The selected vehicle trajectory for the analysis is illustrated in Fig.\ref{fig:VehTraj}.

The LoSM simulation includes 25 test cases, comprising a theoretical steering angle \( \theta_R \) of \( \text{arccot} (15)^\circ \); theoretical sideslip angles \( \beta_R \) of 0°, -30°, -60°, 30°, and 60°; and forward velocities of 1, 2, 3, 4, and 5 m/s. The overall trajectory involves moving forward and turning left. The actual steady-state steering radius $R$ and sideslip angles $\beta$ for each condition are shown in Fig.\ref{fig:DynCharLoSM_A}-\ref{fig:DynCharLoSM_B}.

From Fig.\ref{fig:DynCharLoSM_A}, when \( \beta_R = 0 \), the steering radius \( R \) remains nearly constant as the forward velocity \( v \) increases, indicating neutral steering. In this scenario, the steering radius is perpendicular to the vehicle body, making the AWOISV dynamics consistent with those of a traditional multi-axle vehicle. The use of the traditional understeer coefficient further confirms this neutral steering behavior. This validates the proposed generalized dynamic model of the AWOISV and demonstrates its compatibility with traditional vehicle dynamics. When \( \beta_R \) is negative, the actual steering radius \( R \) of the AWOISV decreases as forward velocity \( v \) increases, indicating oversteering. Moreover, larger \( \beta_R \) values lead to more pronounced oversteering characteristics. Conversely, when \( \beta_R \) is positive, the AWOISV exhibits understeering. This behavior is similar to that of multi-axle vehicles employing out-of-phase steering, where the steering center is located behind the vehicle body. This design improves cornering stability for heavy vehicles at high speeds.

Fig.\ref{fig:DynCharLoSM_B} shows that whether \( \beta_R \) is positive or negative, the actual sidesilp angle \( \beta \) decreases as forward velocity \( v \) increases. When \( \theta_R \) is positive, the vehicle turns left, as defined in Section \ref{sec:MtnCharAndRprstaion}. This indicates that when the AWOISV is in LoSM, the direction of forward velocity consistently changes in the opposite direction to the steering. Additionally, at higher speeds, the magnitude of this directional change becomes more pronounced.

The LaSM simulation also includes 25 test cases, comprising a theoretical steering angle \( \theta_R \) of \( \text{arccot}(10)^\circ \); theoretical sideslip angles \( \beta_R \) of 80°, 85°, 90°, -85°, and -80°; and target speeds of 1, 2, 3, 4, and 5 m/s. The actual steady-state steering radius and sideslip angles for each condition are shown in Fig.\ref{fig:DynCharLaSM_A}-\ref{fig:DynCharLaSM_B}. Please note that in the figure, -$80^\circ$ is equivalent to $100^\circ$.

Fig.\ref{fig:DynCharLaSM_A} shows that when the theoretical sidesilp angle \( \beta_R \) is negative, the AWOISV exhibits oversteering, while a positive \( \beta_R \) leads to understeering. Fig.\ref{fig:DynCharLaSM_B} shows that in lateral mode, the actual sidesilp angle \( \beta \) increases as the forward speed increases. This result is consistent with the longitudinal steering case. The vehicle’s speed direction always changes in the opposite direction to the steering, with larger changes occurring at higher speeds.

In summary, a qualitative conclusion can be drawn that when the theoretical steering radius angle \( \theta_R \) and the theoretical sidesilp angle \( \beta_R \) have the same sign, the AWOISV exhibits understeering. When they have opposite signs, the AWOISV exhibits oversteering.
    
    

\begin{figure}[htbp]
\centering
\subcaptionbox{\label{fig:overallArchtctr}}[0.9\textwidth]{
    \includegraphics[width=0.9\textwidth]{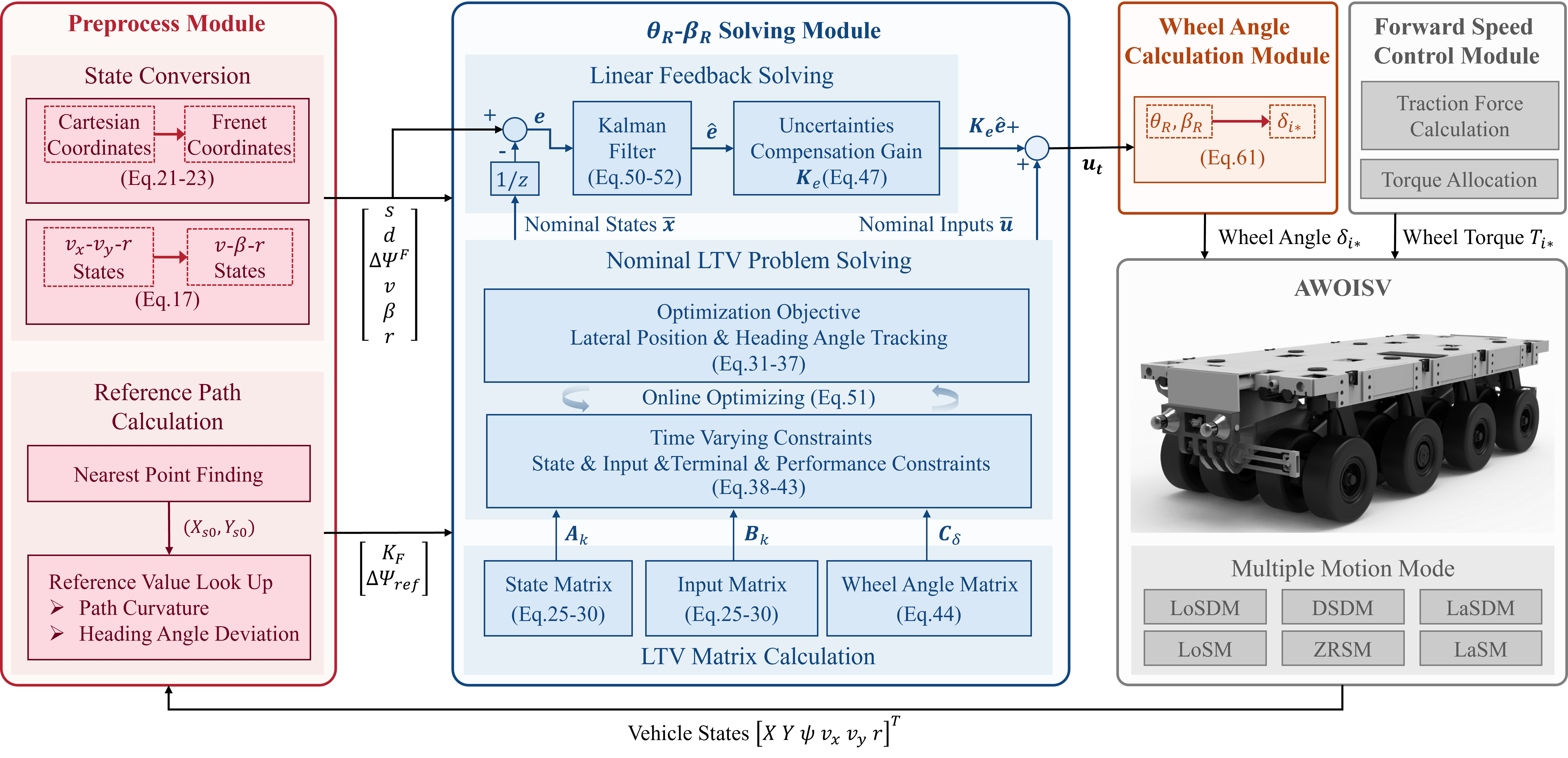} 
}

\subcaptionbox{\label{fig:PrdctvMdlInFrnt}}[0.6\textwidth]{
    \includegraphics[width=0.6\textwidth]{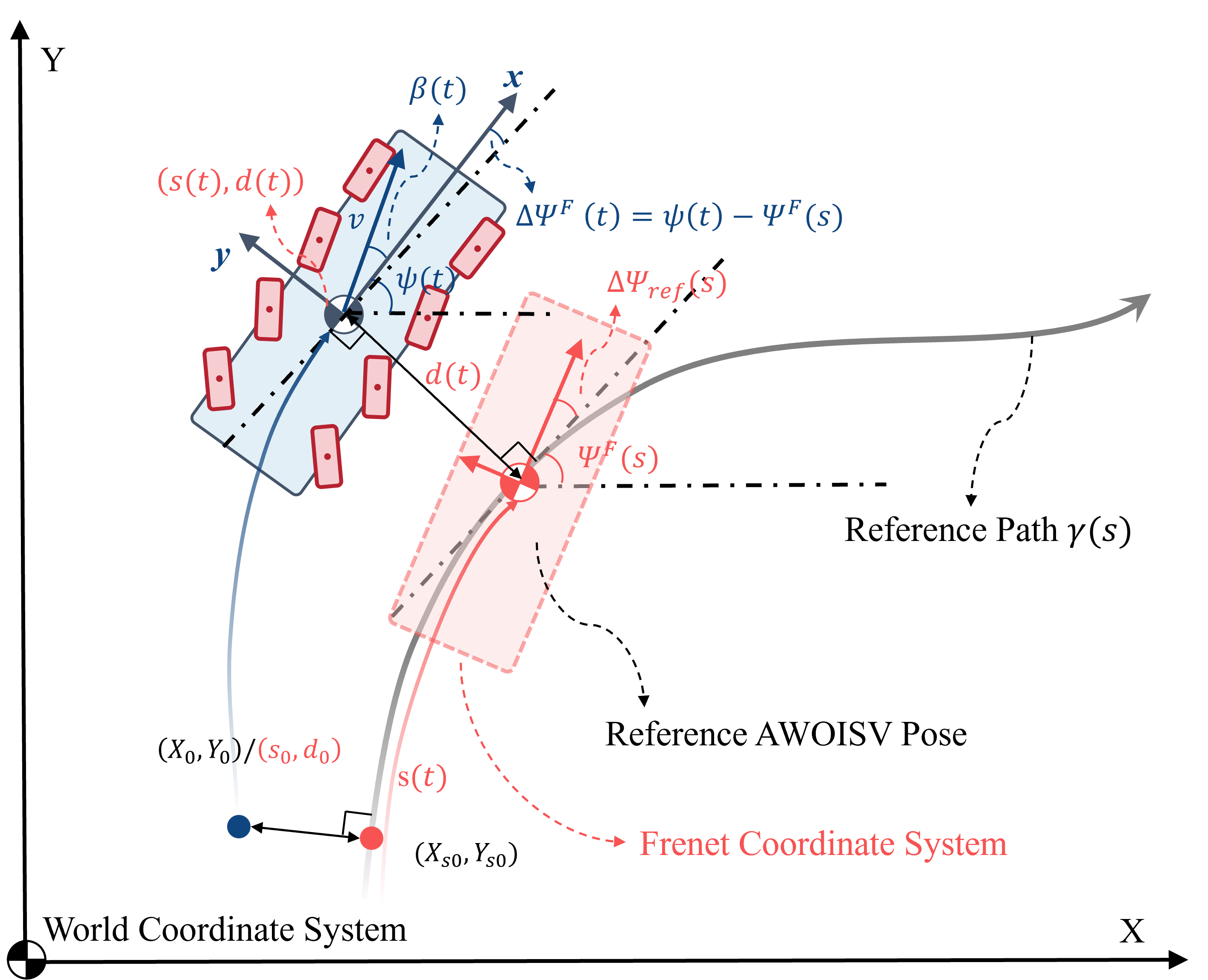} 
}
\caption{Overall architecture of the proposed control strategy and the preidctive model. (a)Overall architecture ,(b)Predictive model of AWOISV in the Frenet coordinate system.}
\end{figure}

\section{Filtered Tube-based Linear Time Varying MPC for Path Tracking}\label{sec:ctrlMthd}
Compared to traditional vehicles, AWOISV offers greater degrees of freedom in path tracking tasks, which allows the AWOISV to not only follow the lateral position but also maintain any desired angle with the reference path based on the actual task requirements. To this end, this paper proposes a novel filtered tube-based linear time varying MPC (FT-LTVMPC) strategy for the path tracking task of the AWOISV. The overall architecture of the path tracking controller is shown in Fig.\ref{fig:overallArchtctr}, which consists of four main modules. The Preprocess Module performs coordinate conversion and reference path calculation. The $\theta_R$-$\beta_R$ Solving Module achieves lateral position and heading tracking through filtered linear feedback for uncertainty compensation and online optimization with time-varying state matrix and constraints. The Wheel Angle Calculation Module converts $\theta_R$ and $\beta_R$ into the wheel angles. On the right, the Forward Velocity Control Module manages traction force and torque allocation, though it is not the focus of this study.


\subsection{Predictive Model in Frenet Coordinate}\label{sec:prdctvMdl}
In the previous sections, a dynamic model for the AWOISV is developed in Eq. \ref{eq:VBRMdl}. However, this model is not ideal for path tracking for several reasons. First, the model's complexity arises from strong coupling among lateral, yaw, and forward dynamics, complicating the solving process. Path tracking and speed tracking are typically treated as decoupled tasks, focusing on lateral and yaw dynamics. To simplify the prediction model, forward velocity is assumed to remain quasi-static and constant within the MPC horizon, while the influence of longitudinal tire forces on sideslip angle and yaw rate is neglected. Second, when the tracking objective involves following the reference path’s \(X_{\text{ref}}\), \(Y_{\text{ref}}\), and heading angle \(\varPsi_{\text{ref}}\) in the global coordinate system, decoupling forward velocity tracking from path tracking is challenging. Minimizing the deviation between the vehicle's coordinates \((X, Y, \varPsi)\) and the reference values inherently includes the vehicle's speed, complicating independent control of forward velocity. Therefore, path tracking should prioritize lateral deviation and heading angle error relative to the reference path.

Given these considerations, this paper will establish a predictive model in the Frenet coordinate system, focusing on the sideslip angle \( \beta \) and yaw rate \( r \), to improve the accuracy and efficiency of path tracking control. The proposed predictive model in Frenet coordinate system is illustrated in Fig.\ref{fig:PrdctvMdlInFrnt}. The reference path \( \gamma \) is parameterized by the arc length \( s \). The position of the vehicle in the Frenet coordinate system is described as \( (s, d, \varPsi) \). 


The initial position of the vehicle \((X_0, Y_0)\) corresponds to the Frenet coordinates \((s_0, d_0)\) at the projection point on the reference path. \( s_0 \) is defined as follows:
\begin{equation}
    s_0 = \arg \  \min_{s} \left( \left[X(s)-X(0)\right]^{2}+\left[Y(s)-Y(0)\right]^{2} \right).
\end{equation}

The initial position of the vehicle \((X_0, Y_0)\) corresponds to the Cartesian coordinates \((X_{s_0}, Y_{s_0})\) at the projection point on the reference path \( d_0 \) is defined as follows:
\begin{equation}
  \begin{aligned}
    d_0 &= \text{sign}\left(\left(Y_0 - Y_{s_0}\right) \cos \varPsi^{F}(s_0)-\left(X_0 - X_{s_{0}}\right) \sin \varPsi^{F}(s_0)\right) \\
    &\quad \cdot \sqrt{\left(X_0-X_{s_{0}}\right)^{2}+\left(Y_0 - Y_{s_0}\right)^{2}}
  \end{aligned}
\end{equation}
where \(\varPsi^{F}(s)\) represents the heading angle of the reference path \(\gamma\) at the position \(s\).

At the initial moment, the angle difference between the direction of the forward velocity $v$ and the heading angle of the reference path at the projection point represents \(\Delta \varPsi(s_0, 0)\), which varies at any time as \(\Delta \varPsi(s, t)\):
\begin{subequations}
  \begin{align}
  \Delta \varPsi(s_0,0) &= \varPsi(0) - \varPsi^{F}(s_0)\\
  \Delta \varPsi(s,t) &= \varPsi(t) - \varPsi^{F}(s).
  \end{align}
\end{subequations}

In summary, the relative position and pose between the AWOISV and reference path in Frenet coordinate system can be represented by \((s, d, \Delta \varPsi^{F})\). By treating \(s\) and \(d\) as functions of time \(t\), and considering the curvature \(\kappa^{F}(s)\) at the reference path as a parameter that varies with \(s\) denoted by \(\kappa^{F}_{s}\), the state equations for \((s, d, \Delta \varPsi^{F})\) can be expressed as follows:
\begin{subequations}\label{eq:stFuncFrnt}
  \begin{align}
    \varPsi(t) &= \psi(t) + \beta(t) \\
    \dot \psi(t) &= r\\
    \dot {\Delta \varPsi^{F}}(s) &= \dot \varPsi(t) - \frac {\text d} {\text d s} {\varPsi^{F}}(s) \dot s = \dot \varPsi(t) - \kappa^F_s \dot s.
  \end{align}
\end{subequations}

By combining Eq.\ref{eq:VBRMdl} and Eq.\ref{eq:stFuncFrnt}, treating the forward velocity \(v(t)\) as a time-varying constant \(v_t\), and neglecting the longitudinal tire forces \(F_{x,i*}\), the predictive model in the Frenet coordinate system can be derived as follows:
\begin{subequations}\label{eq:stFuncFrnt}
  \begin{align}
    \dot {\bm x}(t) &= \bm f_{predict} \left( \bm x(t), \bm u(t), t \right) \\
    \bm x(t_0) &= \left[ s_{0} \  d_{0} \  \Delta \varPsi^{F}_{0}\  \beta_{0} \ r_{0} \right]^T \\
    \bm u(t_0) & = \left[ \theta_{R_0} \  \beta_{R_0} \right]^T
    \in \mathcal{U} \subset \mathbb{R}^{n_u} \\
    \bm x(t) & = \left[ s(t) \  d(t) \   \Delta \varPsi^{F}(t)\  \beta(t) \ r(t) \right]^T
    \in \mathcal{X} \subset \mathbb{R}^{n_x} \\
    \bm u(t) & = \left[ \theta_{R}(t) \  \beta_{R}(t) \right]^T
    \in \mathcal{U} \subset \mathbb{R}^{n_u} \\
    t & \in [t_0, t_f].
  \end{align}
\end{subequations}

Here, $ \bm f_{predict} \left( \bm x(t), \bm u(t), t \right) $ is expressed as Eq.\ref{eq:PrdctvMdl}.
{
  \scriptsize
  \begin{subequations}\label{eq:PrdctvMdl}
  \begin{align}
    \dot s(t) &= \frac{ v_t \cos \left(\psi(t) + {\Delta \varPsi^{F}}(t) \right) }{1 - d(t) \kappa^F_s} \\
    \dot d(t) &= v_t \sin \left(\psi(t) + {\Delta \varPsi^{F}}(t) \right) \\
    \dot {\Delta \varPsi^{F}}(t) &= r(t) - \kappa^F_s \dot s(t) \\
    \dot \beta(t) &= \frac 1 {m v_t} \sum_{* = L}^{R} \sum_{i = 1}^{N} \varXi_{i*}(t)
    \left(
    \left(\cos \beta(t) + \varTheta_{i*}(t) \sin \beta(t)  \right) 
    \right. \\ &\quad \left. \notag  
    \overline {C_{\alpha, i*}} 
    \left(
    \frac{\left(v_t \sin \beta(t) + r(t) x_{i*}\right)+\left(v_t \cos \beta(t) - r(t) y_{i*}\right) \varTheta_{i*}(t)}
    {\left(v_t \cos \beta(t) - r(t) y_{i*}\right)-\left(v_t \sin \beta(t) + r(t) x_{i*}\right) \varTheta_{i*}(t)}
    \right)
    \right) - r(t)\\
    \dot r(t) &= \frac 1 {I_Z} \sum_{* = L}^{R} \sum_{i = 1}^{N} \varXi_{i*}(t)
    \left(
    \left(x_{i*} + \varTheta_{i*}(t) y_{i*}  \right) 
    \overline {C_{\alpha, i*}} 
    \left(
    \frac{\left(v_t \sin \beta(t) + r(t) x_{i*}\right)+\left(v_t \cos \beta(t) - r(t) y_{i*}\right) \varTheta_{i*}(t)}
    {\left(v_t \cos \beta(t) - r(t) y_{i*}\right)-\left(v_t \sin \beta(t) + r(t) x_{i*}\right) \varTheta_{i*}(t)}
    \right)
    \right).
  \end{align}
  \end{subequations}
}

To ensure real-time solvability of the MPC problem, Eq. 26 is further transformed into a linear time-varying (LTV) model, replacing the original nonlinear model. By performing a Taylor expansion of Eq.\ref{eq:PrdctvMdl} around time \(t\) and letting \(\bm x(t) = \bm x_0\) and \(\bm u(t) = \bm u_0\), Eq.\ref{eq:PrdctvMdl} can be approximated as an LTV model:
\begin{equation}\label{eq:ltvMdl}
  \dot {\bm x}(t) = \bm A_t \bm x(t) + \bm B_t \bm u(t) + \bm w(t)
\end{equation}
where \( \bm w(t) \) denotes the disturbance including the bounded additive uncertainty or external disturbance, and the time-varying coefficient matrices are derived as:
\begin{equation}
  \left. \bm A_t = \frac{\partial \bm f_{predict}} {\partial \bm x} \right|_{\bm x_0, \bm u_0},\ 
  \left. \bm B_t = \frac{\partial \bm f_{predict}} {\partial \bm u} \right|_{\bm x_0, \bm u_0}.
\end{equation}

\( \bm A_t \) and \( \bm B_t \) are the functions of state variables \(\left[ s_{0} \  d_{0} \  \Delta \varPsi^{F}_{0}\  \beta_{0} \ r_{0} \right]^T\), inputs \( \left[ \theta_{R_0} \  \beta_{R_0} \right]^T \) and tire slip angles \( \alpha_{i*,0} \) at time \(t\), where \( \alpha_{i*,0} \) is used to approximate proper lateral tire force via Eq.\ref{eq:lateralFrc}.

Furthermore, Eq.\ref{eq:ltvMdl} is expressed as the discretized equation:
\begin{subequations}\label{eq:ltvMdlDscrt}
  \begin{align}
    \bm x(k+1) &= \bm A_k \bm x(k) + \bm B_k \bm u(k) + \bm w(k) \\
    k & \in [0, N]
  \end{align}
\end{subequations}
where
\begin{subequations}
  \begin{align}
    \bm A_k &= \bm I_{n_x} + \bm A_t T_{mpc}, \\
    \bm B_k &= \bm B_t T_{mpc}.
  \end{align}
\end{subequations}

$T_{mpc}$ is the sample time of the MPC.

\subsection{Optimization Objective and Constraints}\label{sec:ObjandCnstrnts}
The first optimization objective $J_1$, which is designed to accurately track the lateral position, is expressed as follows:
\begin{equation}
  J_1 =w_1 \frac {\sum_{k = 0}^{N-1} \left( d(k) - d_{ref} \right)^2}{d_{ {norm}}}.
\end{equation}

Here, \( w_1 \) represents the weight of the first objective. \( d_{{ref}} \) denotes the reference lateral deviation, which is set to 0. \( d_{norm} \) indicates the nominal lateral deviation. It is used for normalization between different optimization objectives and is set to 0.5 m.

The second objective $J_2$ is to track the heading angle. The reference heading angle deviation, denoted as \( \Delta \varPsi_{ref} \), is determined by the information of the reference path. The nominal heading angle deviation, \( \Delta \varPsi_{norm} \), is set to \( \frac{\pi}{6} \):
\begin{equation}
  J_2 = w_2 \frac{ \sum_{k = 0}^{N-1} \left(  \Delta \varPsi^{F}(k) - \Delta \varPsi_{ref} \right)^2}{\Delta \varPsi_ {norm}}.
\end{equation}

The third objective $J_3$ is to prevent excessively large control inputs. The nominal control inputs, denoted as \( [\theta_{R, norm}, \beta_{R, norm}]^T \), are set to \( [\frac{\pi}{2}, \frac{\pi}{2}]^T \).
\begin{equation}
  J_3 =w_3 \left( \frac {\sum_{k = 0}^{N-1} \left(\theta_R(k ) \right)^2} {\theta_{R, norm}} + \frac {\sum_{k = 0}^{N-1} \left(\beta_R(k ) \right)^2} {\beta_{R, norm}} \right).
\end{equation}

The fourth optimization objective  \( J_4 \) is the terminal cost, which ensures the asymptotic stability of the MPC.
\begin{equation}
  J_4 = \bm x(N)^T \bm P_f x(N).
\end{equation}

In summary, the complete optimization objective $J$ can be expressed as follows:
\begin{equation}\label{eq:objecitve}
  \begin{aligned}
    J =& \sum_{k = 0}^{N-1} \left(\bm x(k) - \bm x_{ref} \right)^T \bm Q \left(\bm x(k) - \bm x_{ref} \right)
    + \sum_{k = 0}^{N-1} \bm u(k)^T \bm R u(k) \\
    & + \left(\bm x(N) - \bm x_{ref} \right)^T \bm P_f \left(\bm x(N) - \bm x_{ref} \right)
  \end{aligned}
\end{equation}
where
\begin{subequations}
  \begin{align}
    \bm Q &= \text{diag} \left(0, \frac{w_1}{d_{norm}}, \frac{w_1}{\Delta \varPsi_{norm}}, 0, 0, 0 \right) \\
    \bm R &= \text{diag} \left(\frac{w_3}{\theta_{R, norm}}, \frac{w_3}{\beta_{R, norm}} \right).
  \end{align}
\end{subequations}

\( \bm P_f \) is the solution to the following equation:
\begin{equation}
  \bm P_f-(\bm A_t-\bm B_t \bm K_f)^{T} \bm P_f(\bm A_t-\bm B_t \bm K_f) = \bm Q + \bm K_f^{T} \bm R \bm K_f
\end{equation}
where $\bm {K}_f$ satisfies $|\text{eig}(\bm A_t - \bm B_t \bm K_f)|<1$.

The constraints consist of four parts: state constraints, control input constraints, terminal constraints, and performance constraints.

The state constraints limit the sideslip angle and yaw rate, reflecting the driving stability of the AWOISV. These constraints are expressed as follows:
\begin{subequations}
  \begin{align} 
    \label{eq:beta_constraint}
    \beta(k) &\in \left(\beta_{\min}(v,\theta_R,\beta_R),  \beta_{\max}(v,\theta_R,\beta_R) \right)  \\ 
    \label{eq:r_constraint}
    r(k) &\in \left(r_{\min}(v,\theta_R,\beta_R), r_{\max}(v,\theta_R,\beta_R)  \right)\\ 
    \mathcal{X} &\triangleq \{ \bm{x} \in \mathbb{R}^{n_x} \mid \text{Eq.\ref{eq:beta_constraint}, Eq.\ref{eq:r_constraint}} \}.
  \end{align}\label{eq:state_constraints}
\end{subequations}

Since the core focus of this paper is the path tracking control of the AWOISV rather than its stability, Eq.\ref{eq:state_constraints} is simplified based on empirical considerations as follows:
\begin{subequations}
  \begin{align} 
    \label{eq:beta_constraint_simp}
    \beta(k) &\in \left(\beta_R(k) - \beta_{\max},  \beta_R(k) + \beta_{\max} \right)  \\ 
    \label{eq:r_constraint_simp}
    r(k) &\in \left(-r_{\max}, r_{\max}\right). 
  \end{align}
\end{subequations}

The control inputs are determined by the domains of \( \theta_R \) and \( \beta_R \), as shown below:
\begin{subequations}
  \begin{align} 
    \label{eq:thetaR_constraint}
    \theta_{R}(k) &\in \left(- \frac {\pi} 2, \frac {\pi} 2 \right)\\
    \label{eq:betaR_constraint}
    \beta_{R}(k) &\in \left(- \frac {\pi} 2, \frac {\pi} 2 \right)\\ 
    \mathcal{U} &\triangleq \{ \bm {u} \in \mathbb{R}^{n_u} \mid \text{Eq.\ref{eq:thetaR_constraint}, Eq.\ref{eq:betaR_constraint}} \}.
  \end{align}    
\end{subequations}

The terminal constraint is expressed as follows:
\begin{equation}\label{eq:terminalConstraint}
  \bm x (N) \in \mathcal X_f.
\end{equation}

The performance constraints include limitations on the ICR position and the steering rate of the wheels.

According to Section \ref{sec:vbrMethod}, the AWOIS can achieve most of the motion states using only the LoSM. Therefore, this paper restricts the ICR position within the LoSM range:
\begin{equation}\label{eq:thetaRbetaRCnstrnt}
  \left| \cot \theta_{R}(k) \cos \beta_{R}(k) \right| \in \left( \frac M 2, + \infty \right).
\end{equation}

While the \( \theta_R \)-\( \beta_R \) method effectively characterizes the dynamic behavior of the AWOISV, the rates of change of \( \theta_R \) and \( \beta_R \) do not directly reflect the wheel steering rate. This discrepancy may lead to situations where wheel steering cannot adequately respond to changes in \( \theta_R \) and \( \beta_R \). Therefore, this paper linearizes Eq.\ref{eq:CalDelta} and Eq.\ref{eq:thetaRDefination} to derive the expression for the wheel steering rate constraint as follows:
\begin{equation}\label{deltaCnstrnt}
  \delta_{i*}(k+1) - \delta_{i*}(k) \in (-\omega_{i*,\max}T_{mpc}, \omega_{i*,\max}T_{mpc})
\end{equation}
where
\begin{subequations}
  \begin{align} 
    & \delta_{i*}(k) = \bm C_{\delta} (\bm u(k) - \bm u(0)) + \delta_{i*,0}\\
    & \left. \bm C_{\delta} = \frac {\partial}{\partial \bm u} \left( \arctan \left( \frac {x_{i*} +\cot \theta_R(k) \sin \beta_R(k)} {\cot \theta_R(k) \cos \beta_R(k) - y_{i*}}   \right) \right) \right|_{\bm \delta_{i*,0}, \bm u_0}.
  \end{align}    
\end{subequations}

As derived above, the norminal linear time-varying MPC (LTVMPC) and nonlinear MPC (NMPC) approaches can be obtained.

\subsection{Filtered Tube-based MPC Problem Formation}
The consideration of uncertainties \( \bm{w}(k) \) in the evolution of the system state \( \bm{x}(k) \) requires defining a nominal system as:
\begin{equation}\label{eq:normMdl}
  \bar {\bm x} (k+1) = \bm A_k \bar {\bm x}(k) + \bm B_k \bar {\bm u}(k)
\end{equation}
where \( \bar{\bm x} \) and \( \bar{\bm u} \) represent the state vector and control vector of the nominal model, respectively.

The state error between the nominal model and the actual model is defined as:
\begin{equation}\label{eq:stateErr}
  \bm e(k) = \bm x(k) - \bm {\bar x}(k).
\end{equation}


The control objective of tube-MPC is to compensate for the difference between the real state and the nominal state, \( \bar{\bm x}(k) \), and to steer the nominal system as close as possible to the reference trajectory. To achieve this, the control law is defined as a combination of the nominal optimal control term and a linear feedback control term, expressed as follows:
\begin{equation}\label{eq:inputRealFiltered}
  \bm u(k) = \bar {\bm u}(k) + \bm K_e {\bm e}(k).
\end{equation}

By combining Eq.\ref{eq:normMdl}, Eq.\ref{eq:stateErr}, and Eq.\ref{eq:inputRealFiltered}, the dynamics of the state prediction error can be expressed as follows:
\begin{subequations}\label{eq:errorDynamics}
  \begin{align}
  \bm e(k+1) &= \bm A_e \bm e(k) + \bm w(k)\\
  \bm A_e &= \bm A_k + \bm B_k \bm K_e
  \end{align}
\end{subequations}
were $\bm w(k)$ is assumed to be zero-mean Gaussian with covariance \( \bm Q_k \). If \(\bm K_e\) satisfies \(|\text{eig}(\bm A_k - \bm B_k \bm K_e)| < 1\), then \(\bm A_e\) is Hurwitz and represents a stable system.

To mitigate the effects of measurement noise and reduce control chattering caused by small fluctuations in the error, the state error \( \bm e(k) \) is processed through a Kalman filter and a dynamic hysteresis mechanism to obtain the filtered error \( \hat{\bm e}(k) \). The observation equation for Kalman filter is given by:
\begin{equation}\label{eq:kalmanObsvr}
  \bm y(k) = \bm C \bm e(k) + \bm v(k)
\end{equation}
where \( \bm v(k) \) is the measurement noise with covariance \( \bm R_k \) and $\bm C = \bm I_5$ in this paper.

Then, the Kalman filter can be defined as:
\begin{equation}\label{eq:kalmanFilter}
  \hat{\bm e}(k) = \hat{\bm e}^-(k) + \bm K_k(k) (\bm y(k) - \hat{\bm e}^-(k))
\end{equation}
where
\begin{subequations}
  \begin{align}\label{eq:kalmanUpdate}
  \hat{\bm e}^-(k) &= \bm A_e \hat{\bm e}(k-1)\\\label{eq:kalmanUpdate_a}
  \bm P_k^-(k) &= \bm A_e \bm P_k(k-1) \bm A_e^T + \bm Q_k\\
  \bm K_k(k) &= \bm P_k^-(k) \bm C^T (\bm C \bm P_k^-(k) \bm C^T + \bm R_k)^{-1}\\
  \hat{\bm e}(k) &= \hat{\bm e}^-(k) + \bm K(k) (\bm y(k) - \hat{\bm e}^-(k))\\
  \bm P_k(k) &= (\bm I - \bm K_k(k) \bm C) \bm P_k^-(k).
  \end{align}
\end{subequations}


The dynamic hysteresis mechanism is applied to the filtered error to prevent minor variations from causing frequent control adjustments:
\begin{equation}\label{eq:hysteresis}
  \hat{e}_i(k) =
  \begin{cases}
    \hat{e}_i(k-1), & \text{if } |\hat{e}_i(k) - \hat{e}_i(k-1)| < \epsilon_i \\
    \hat{e}_i(k), & \text{otherwise}
  \end{cases}
\end{equation}
where \( \hat{e}_i(k) \) is the \( i \)-th component of \( \hat{\bm e}(k) \), and \( \epsilon_i \) is the threshold for the \( i \)-th component.





Combining Eq.\ref{eq:kalmanObsvr}, \ref{eq:kalmanFilter} and \ref{eq:kalmanUpdate}, the filtered error can be expressed as:
\begin{equation}
  \hat{\bm e}(k) = \bm A_e \hat{\bm e}(k-1) + \bm K_k(k) \left( \bm e(k) + \bm v(k) - \bm A_e \hat{\bm e}(k-1) \right)
\end{equation}

Assume there exist constants \(\bm M_e\), \(\bm M_{\hat e}\) and \(\bm M_v\) such that the norms of the original error \(\bm e(k)\), filtered error \(\bm {\hat e}(k)\) and measurement noise \(\bm v(k)\) satisfy:
\begin{equation}
    \|\bm e(k)\| \leq \bm M_e, \  \|\bm {\hat e}(k)\| \leq \bm M_{\hat e}, \  \|\bm v(k)\| \leq \bm M_v.
\end{equation}

By substituting these bounds into the previous expression, the following inequality for \(\hat{\bm e}(k)\) can be derived:
\begin{equation}
  \|\hat{\bm e}(k)\| \leq \|\bm A_e - \bm K_k(k) \bm A_e\| \bm M_{\hat e} + \|\bm K_k(k)\| (\bm M_e + \bm M_v).
\end{equation}

To ensure that \(\hat{\bm e}(k)\) is bounded, the condition \(\|\bm A_e - \bm K_k(k) \bm A_e\| < 1\) needs to hold. If this condition is satisfied, it can be concluded that there exists a constant \(C_{\hat{e}}\) such that:
\begin{equation}
  \|\hat{\bm e}(k)\| \leq C_{\hat{e}}
\end{equation}

Thus, It can be proved that there exists a robust positively invariant set (RPI), denoted as:
\begin{equation}
  {\mathcal{E}} = \{ \hat{\bm e}(k) \mid \|\hat{\bm e}(k)\| \leq \bm C_{\hat{e}} \}
\end{equation}


Considering the robust positively invariant set \(\mathcal{E}\), the state constraint set \(\mathcal{X}\), and the input constraint set \(\mathcal{U}\) of the actual model, the state and control input sets of the nominal model can be defined as follows:
\begin{subequations}
  \begin{align}
    \bar {\bm x}(k) & \in \mathcal X \ominus \mathcal E \label{eq:normXSet} \\
    \bar {\bm u}(k) & \in \mathcal U \ominus \bm K_e \mathcal E \label{eq:normUSet}.
  \end{align} 
\end{subequations}

Here, $\ominus$ indicates the Pontryagin difference.

In summary, the optimization of the FT-LTVMPC can be expressed as:
\begin{subequations}
  \begin{align}
    \min_{\bar {\bm x}, \bar {\bm u}} \quad & \text{The objective of Eq.\ref{eq:objecitve}} \\
    \text {s.t.} \quad  & \text{The dynamics of Eq.\ref{eq:normMdl}} \\
    & \text{The nominal state constraints of Eq.\ref{eq:normXSet}} \\
    & \text{The nominal input constraints of Eq.\ref{eq:normUSet}} \\
    & \text{The terminal constraints of Eq.\ref{eq:terminalConstraint}} \\
    & \text{The performance constraints of Eq.\ref{eq:thetaRbetaRCnstrnt} and \ref{deltaCnstrnt}}.
  \end{align} 
\end{subequations}

At any given time \( t \), solving the nominal MPC problem yields the optimal control sequence and optimal state sequence:
\begin{subequations}
  \begin{align}
    \bar{\bm U}^*_t &= \left[ \bar {\bm u}^*(1|t),\bar {\bm u}^*(2|t), \cdots, \bar {\bm u}^*(N-1|t) \right]\\
    \bar{\bm X}^*_t &= \left[ \bar {\bm x}^*(1|t),\bar {\bm x}^*(2|t), \cdots, \bar {\bm x}^*(N|t) \right].
  \end{align}
\end{subequations}

Finally, by combining Eq.\ref{eq:CalDelta} and Eq.\ref{eq:thetaRDefination}, the steering angle control law for each wheel at time \( t \) can be expressed as follows:
\begin{subequations}
  \begin{align}
    \delta_{i*,t} &= \arctan \left( \frac {x_{i*} +\cot \theta_{R,t} \sin \beta_{R,t}} {\cot \theta_{R,t} \cos \beta_{R,t} - y_{i*}}   \right)\\
[\theta_{R,t}\  \beta_{R,t}]^T &= \bm u_t = \bar {\bm u}^*(1|t) + \bm K_e \hat{\bm e}_t.
  \end{align}
\end{subequations}

\section{Simulation and Hardware-in-Loop Experiment}\label{sec:results}
To verify the feasibility of the proposed FT-LTVMPC strategy for simultaneous tracking of lateral position and heading pose in AWOISV, simulations and hardware-in-loop (HIL) experiments are conducted.


\subsection{Experiment Setup}

\begin{table}[!htb]
  \centering
  \caption{Parameters of Control Strategy}
  \label{tab:ControlStrategyParams}
  \small
  \begin{tabular}{l c c}
  \toprule
  \textbf{Parameter} & \textbf{Symbol} & \textbf{Value} \\
  \midrule
  Control period & $T_{\text{control}}$ & 0.02 s \\
  MPC sample time & $T_{\text{mpc}}$ & 0.25 s \\
  Prediction horizon & $N$ & 20 \\
  State cost matrix & $\bm{Q}$ & $\operatorname{diag}[0, 10, 11.7, 0, 0]$ \\
  Input cost matrix & $\bm{R}$ & $\operatorname{diag}[19.1, 19.1]$ \\
  Terminal cost matrix & $\bm{P}_f$ & $\operatorname{diag}[0, 3.3, 3.9, 0, 0]$ \\
  Bound on slide slip angle & $\beta_{\max}$ & $10^\circ$ \\
  Bound on yaw rate & $r_{\max}$ & $0.3 \, \text{rad/s}$ \\
  Wheel steering rate & $\omega_{i*,\max}$ & $90^\circ/\text{s}$ \\
  State noise variance & $\bm M_e$ & $[0, 0.075, 0.022, 0.022, 0.009]$ \\
  Feedback gain & $K_e$ & $\begin{bmatrix} 
  0 & 0.35 & 4.66 & 0.05 & 1.27 \\ 
  0 & 3.09 & 2.12 & 1.66 & 0.02 
  \end{bmatrix}$ \\
  \bottomrule
  \end{tabular}
  
\end{table}

\begin{figure}[!t]
  \centering
  \includegraphics[width=0.6\textwidth]{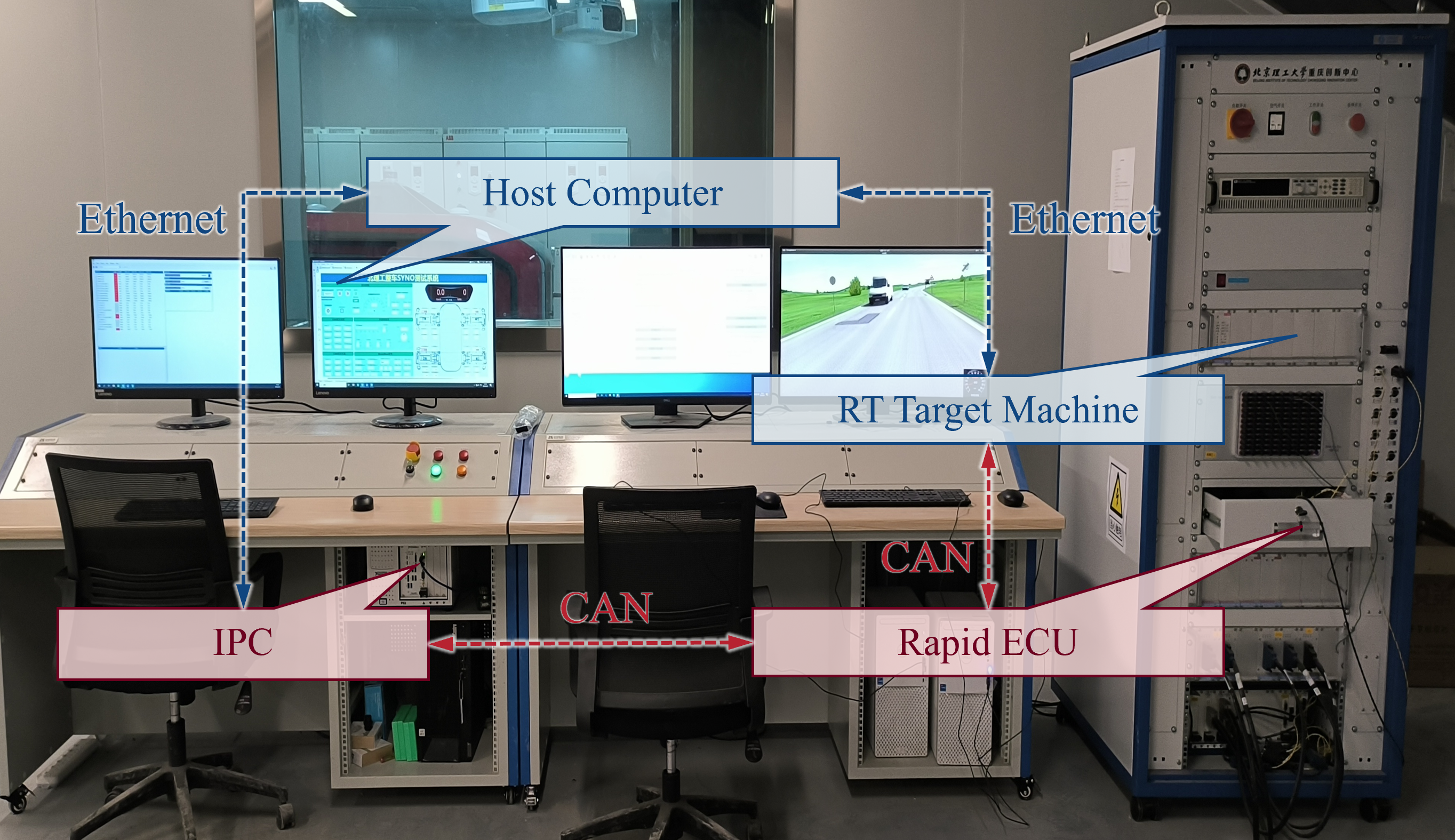}
  \caption{Framework of HIL platform.}
  \label{fig:hilPltfrm}
\end{figure}

The simulation experiment is based on the Trucksim-Matlab/Simulink co-simulation platform, and the HIL experiment is carried out using an industrial personal controller (IPC) and a real-time (RT) target machine. The Trucksim vehicle model is selected as the controlled object, with the steering angle of each wheel independently controlled through a designated control system. The parameter settings of the AWOISV are detailed in Table \ref{tab:4AxleAwoisvParm} in Section \ref{sec:dynCharAnalysis}. The relevant parameters for the proposed control strategy used in both the simulation and HIL experiments are provided in Table \ref{tab:ControlStrategyParams}. To simulate the real-world application of the controller, both the simulation and HIL experiments set the control period to 0.02 seconds (50 Hz).

The platform architecture for the HIL experiments is shown in Fig.\ref{fig:hilPltfrm}. The nonlinear dynamic model of the AWOISV is developed in Trucksim and downloaded to the RT target machine. The proposed control strategies are implemented in the IPC. A rapid ECU processes and distributes execution commands at the lower control level, communicating with both the IPC and the RT target machine via the CAN bus. Experimental data are collected by the host computer, which communicates separately with both the IPC and the RT target machine over Ethernet. 

Three experimental cases are set up in this paper. Cases 1 and 2 are conducted on the co-simulation platform, while Case 3 is performed on the HIL platform. Case 1 aims to verify the feasibility of the proposed FT-LTVMPC and its adaptability to different forward velocities. Case 2 focuses on comparing the control performance of FT-LTVMPC with other control strategies. Finally, Case 3 further analyzes the real-time control effectiveness of FT-LTVMPC and the two other methods using the HIL platform. 

\subsection{Case 1: Sinusoidal Path with Various Forward Velocity}
\begin{figure}[!t]
\centering
  \subcaptionbox{\label{fig:RefPathExprmt}}[0.6\textwidth]{
    \includegraphics[width=0.6\textwidth]{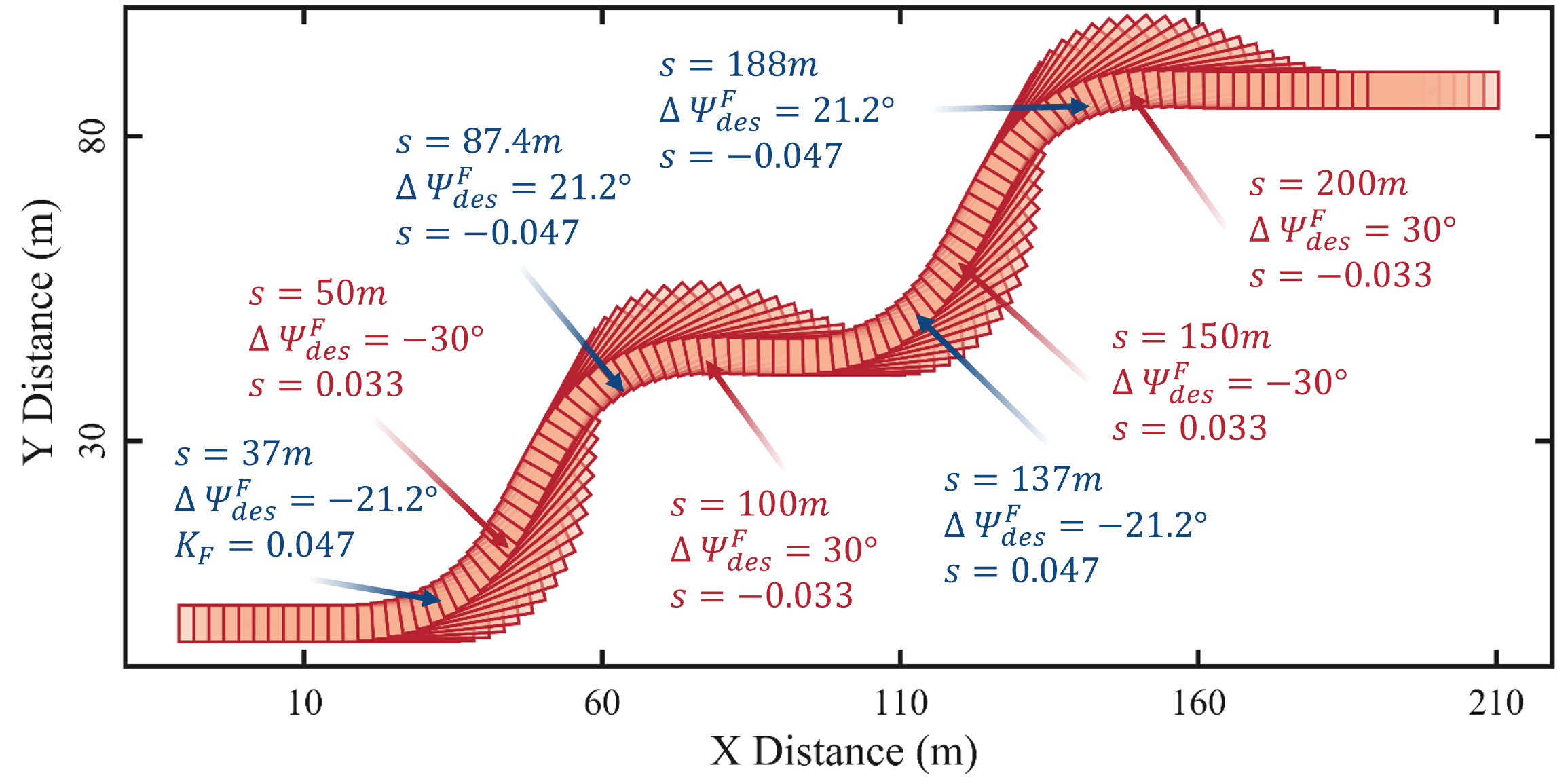}
  }

  \subcaptionbox{\label{fig:TrackingPrfmcofHeadAngle}}[0.6\textwidth]{
    \includegraphics[width=0.6\textwidth]{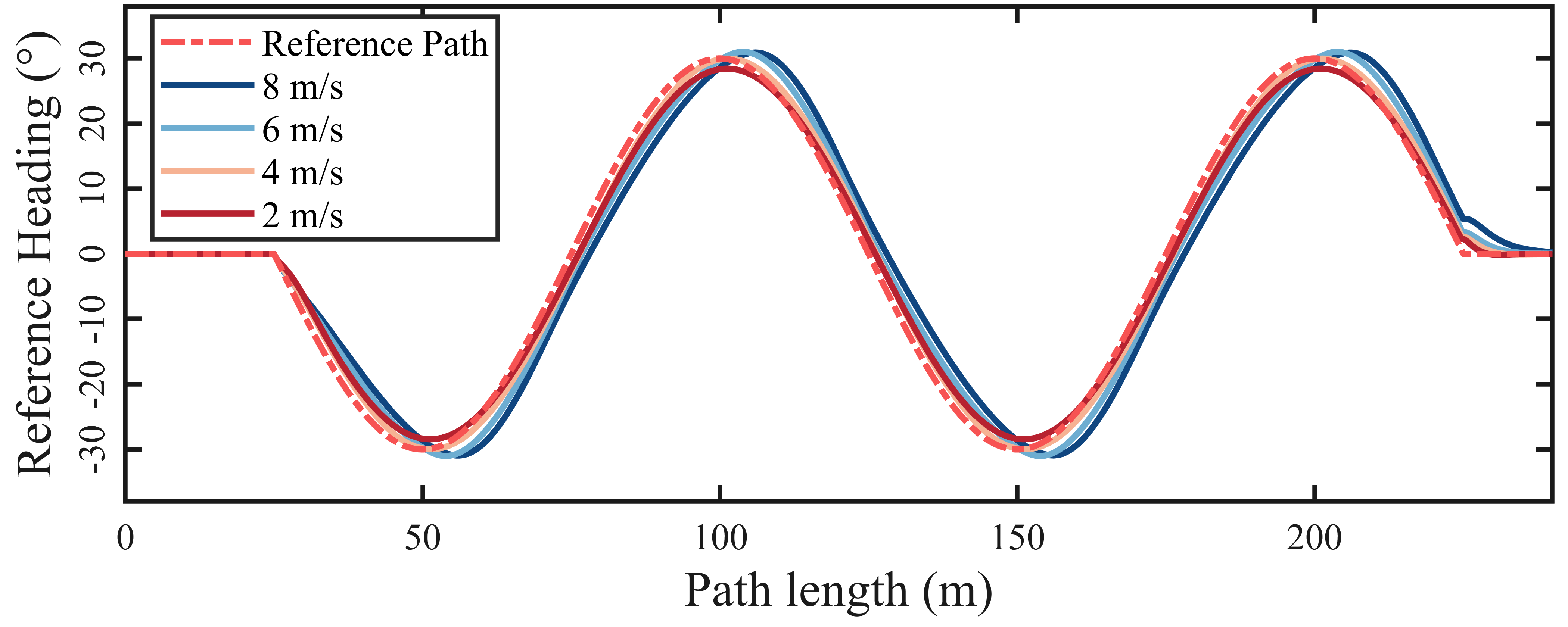}
  }

  \subcaptionbox{\label{fig:TrackingErrinVarVelocity_A}}[0.3\textwidth]{
        \includegraphics[width=0.3\textwidth]{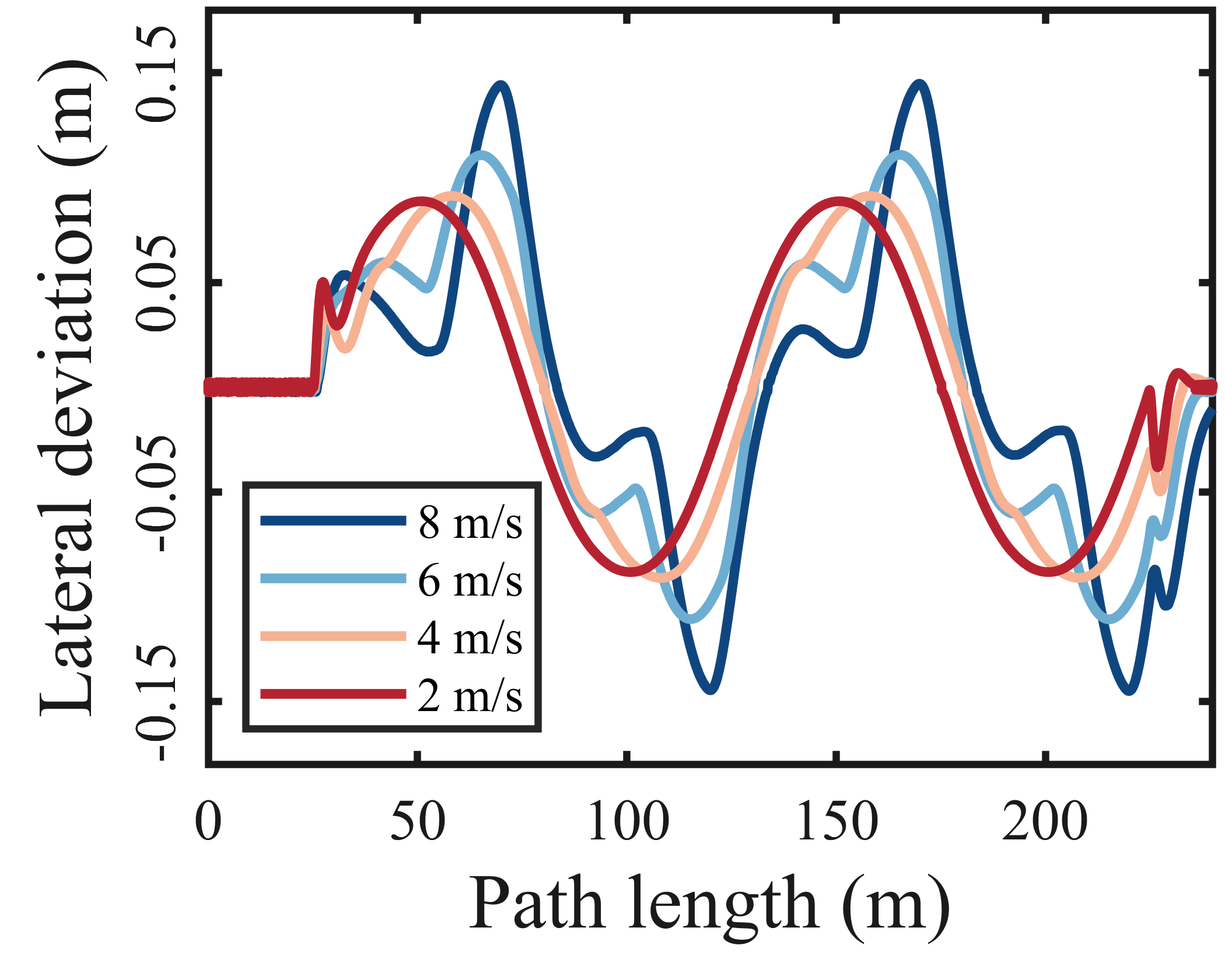} 
    }
    \hspace{2pt}
    \subcaptionbox{\label{fig:TrackingErrinVarVelocity_B}}[0.3\textwidth]{
        \includegraphics[width=0.3\textwidth]{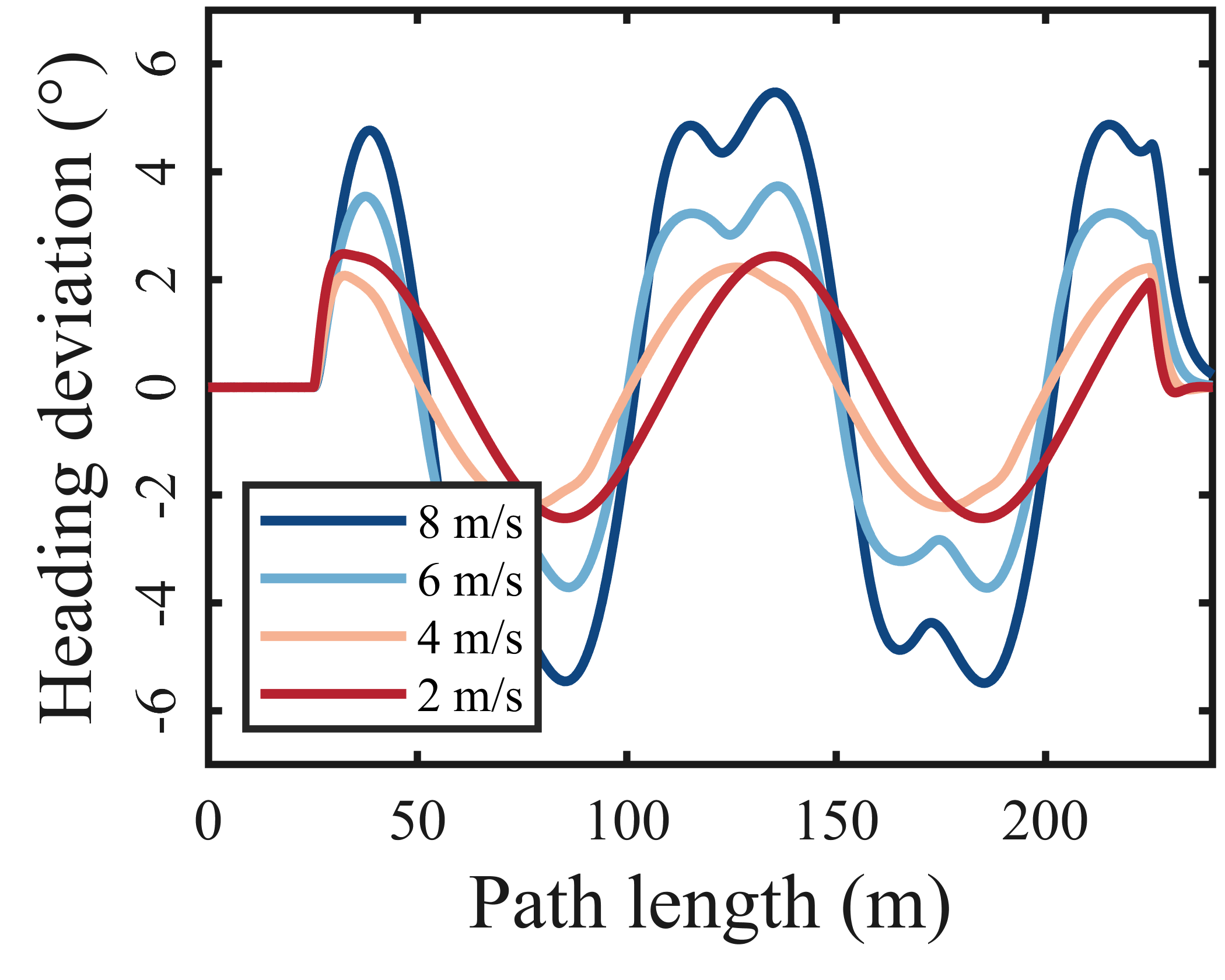}
  }

  \subcaptionbox{\label{fig:betaAndRinVarVelocity_A}}[0.3\textwidth]{
        \includegraphics[width=0.3\textwidth]{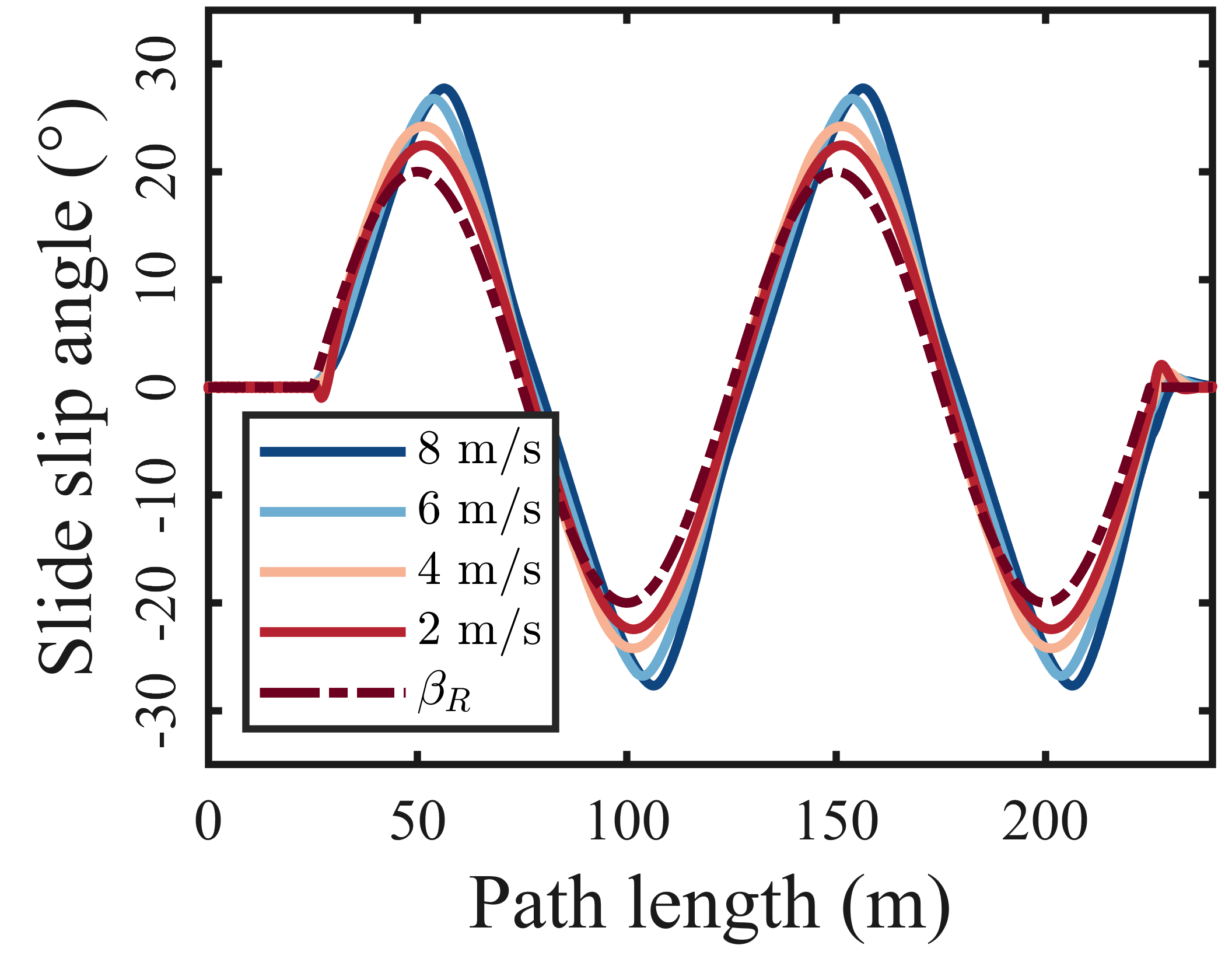} 
    }
    \hspace{2pt}
    \subcaptionbox{\label{fig:betaAndRinVarVelocity_B}}[0.3\textwidth]{
        \includegraphics[width=0.3\textwidth]{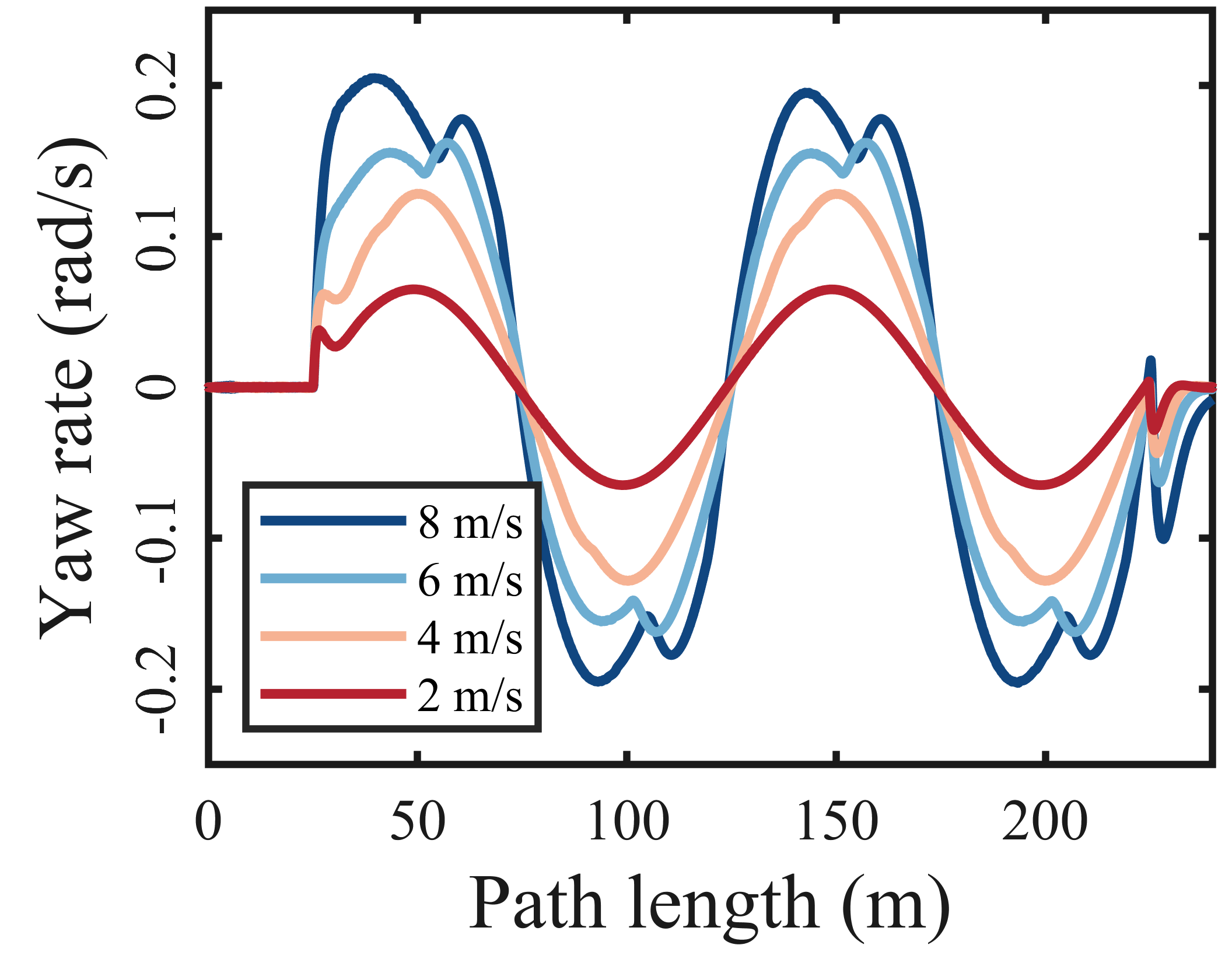}
  }

\caption{Results of case 1. (a)Reference path of simulation experiments, (b)Tracking performance of reference relative heading angle, (c)Lateral position tracking error, (d)Heading angle tracking error, (e)Slide slip angle, (f)Yaw rate.}
\end{figure}

In case 1, the reference path is set as a combination of a straight line and a curve shown in the Fig.\ref{fig:RefPathExprmt}. To fully verify the control performance of the proposed T-LTV-MPC strategy, the curvature and target heading angle of the reference path are set to vary according to a sinusoidal function. The curvature radius ranges from $[-22\,m, 22\,m]$, and the target heading angle varies between $[-30^\circ, 30^\circ]$.




AWOISVs are heavy multi-axle vehicles, which typically operate in special motion modes with arbitrary ICR positions, either remotely controlled or autonomous. The maximum forward velocity of AWOISV is generally below 15 km/h. In case 1, the forward velocities are set to 2 m/s, 4 m/s, 6 m/s, and 8 m/s. These speeds exceed the usual operating velocities of AWOISVs, ensuring the simulation conditions are sufficient to validate the proposed strategy.


Fig.\ref{fig:TrackingPrfmcofHeadAngle} illustrates the tracking performance of the reference relative heading angle. The FT-LTVMPC accurately aligns the reference heading angle across varying forward velocities, following the sinusoidal pattern. As noted in Section \ref{sec:MtnCharAndRprstaion}, this capability highlights a unique motion mode of the AWOISV, which is not available in conventional vehicles. Unlike conventional vehicles with a fixed ICR along the axle line, the AWOISV allows flexible adjustment of the ICR relative to the CG, enabling the forward velocity to align at any angle to the vehicle body. This flexibility supports precise tracking of both lateral position and relative heading angle. Overall, Fig.\ref{fig:TrackingPrfmcofHeadAngle} demonstrates how the proposed FT-LTVMPC leverages the AWOISV’s all-wheel omnidirectional independent steering to maximize maneuverability.

The tracking errors for lateral position and relative heading angle under various forward velocities are illustrated in Fig.\ref{fig:TrackingErrinVarVelocity_A}-\ref{fig:TrackingErrinVarVelocity_B}. Even at a forward velocity of 8 m/s, the FT-LTVMPC effectively maintains the position error and angle error within ranges of 0.15 m and 7°, respectively. Combined with the observations from Fig.\ref{fig:RefPathExprmt}, the maximum position error coincides with the peak of the reference relative heading angle, while the maximum angle error aligns with the point of maximum curvature on the reference path. This phenomenon indicates a certain mutual exclusivity between position tracking and angle tracking. According to the model constructed in Eq.\ref{eq:PrdctvMdl}, an increase in $\Delta \varPsi^{F}$ leads to a corresponding increase in $\dot d$, which subsequently results in a greater lateral position error. Similarly, an increase in $k_F$ yields larger values for $r$ and $\dot {\Delta \varPsi^{F}}$, further contributing to an increase in heading angle error. Therefore, the proposed generalized \(v\)-\( \beta \)-\(r\) dynamic model exhibits a high level of fidelity.

Fig.\ref{fig:betaAndRinVarVelocity_A}-\ref{fig:betaAndRinVarVelocity_B} illustrates the sidesilp angle and the yaw rate under different forward velocities. It can be observed that the sidesilp angle $\beta$ during the path tracking task for the AWOISV differs significantly from that of conventional vehicles. Instead of oscillating around zero, $\beta$ fluctuates around the theoretical sidesilp angle $\beta_R$ derived from Eq.\ref{eq:thetaRDefination}. Furthermore, the deviation from $\beta_R$ increases with higher velocities, which aligns with the analysis presented in Section \ref{sec:dynCharAnalysis}. Meanwhile, both $\beta$ and $r$ values meet the constraints outlined in Eq. 6. These results confirm that the proposed FT-LTVMPC effectively manages the performance constraints of the AWOISV.

\subsection{Case 2: Sinusoidal Path with Various Control Method}

\begin{figure}[htbp]
  \centering
    \subcaptionbox{\label{fig:TrackingPrfmcofHeadAngleMthd}}[0.6\textwidth]{
        \includegraphics[width=0.6\textwidth]{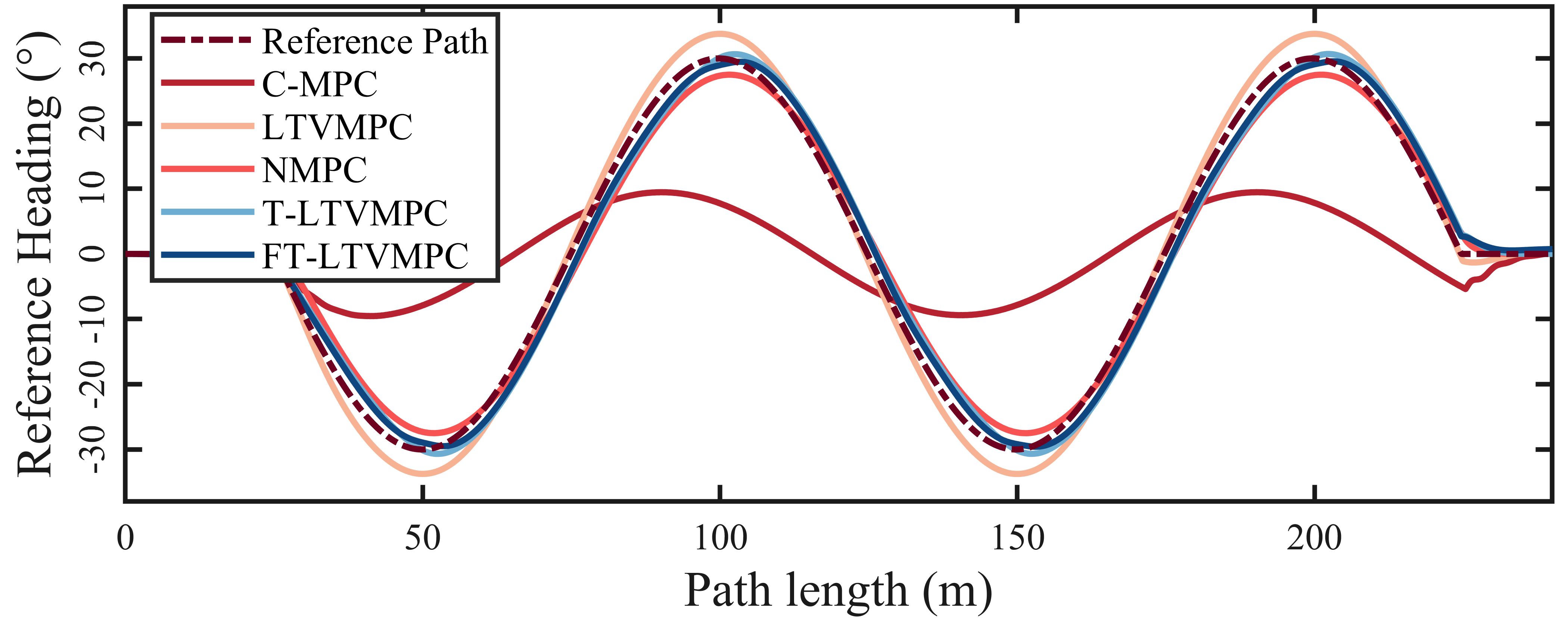} 
    }

    \subcaptionbox{\label{fig:TrackingErrinVarMthd_A}}[0.3\textwidth]{
        \includegraphics[width=0.3\textwidth]{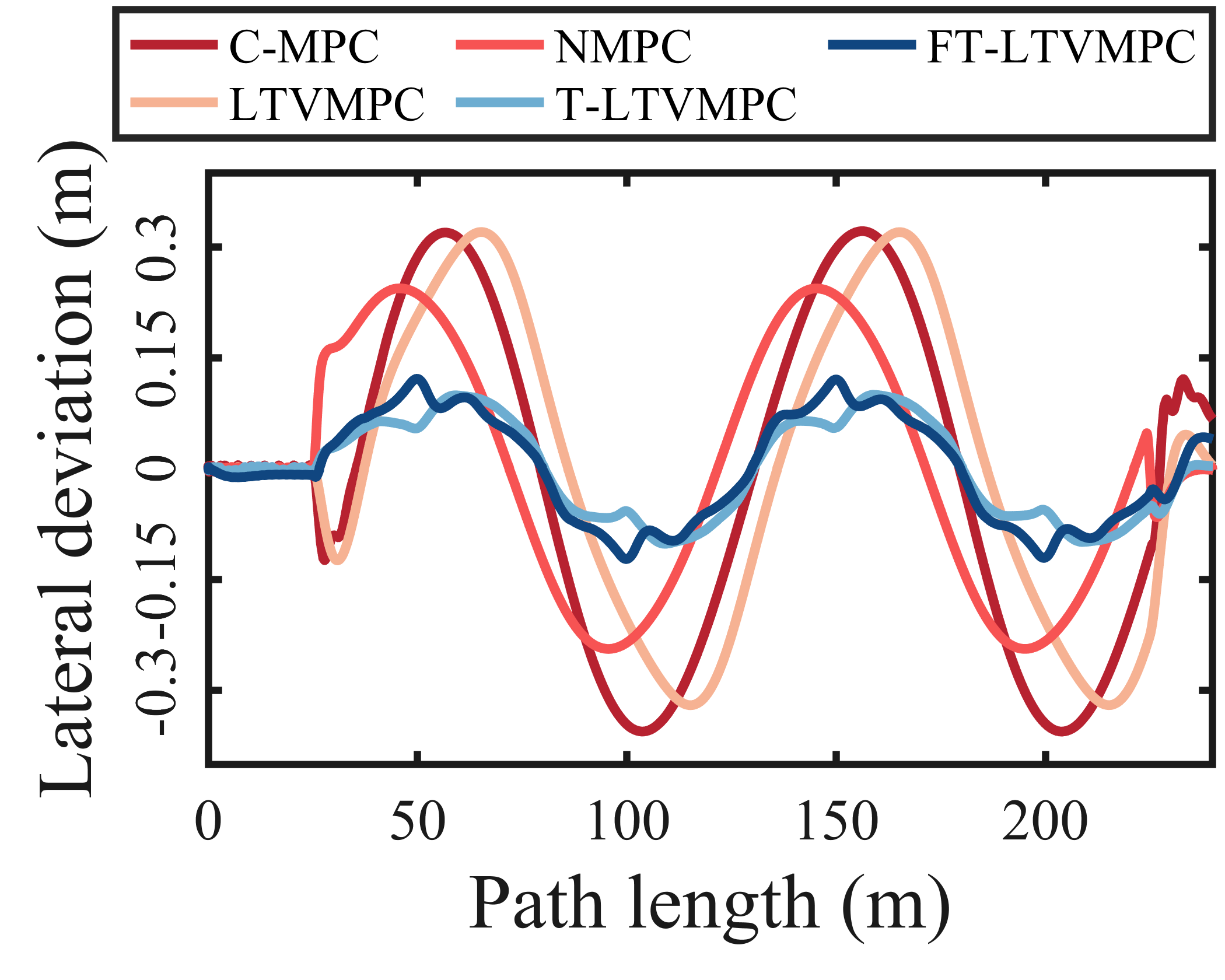} 
    }
    \hspace{2pt}
    \subcaptionbox{\label{fig:TrackingErrinVarMthd_B}}[0.3\textwidth]{
        \includegraphics[width=0.3\textwidth]{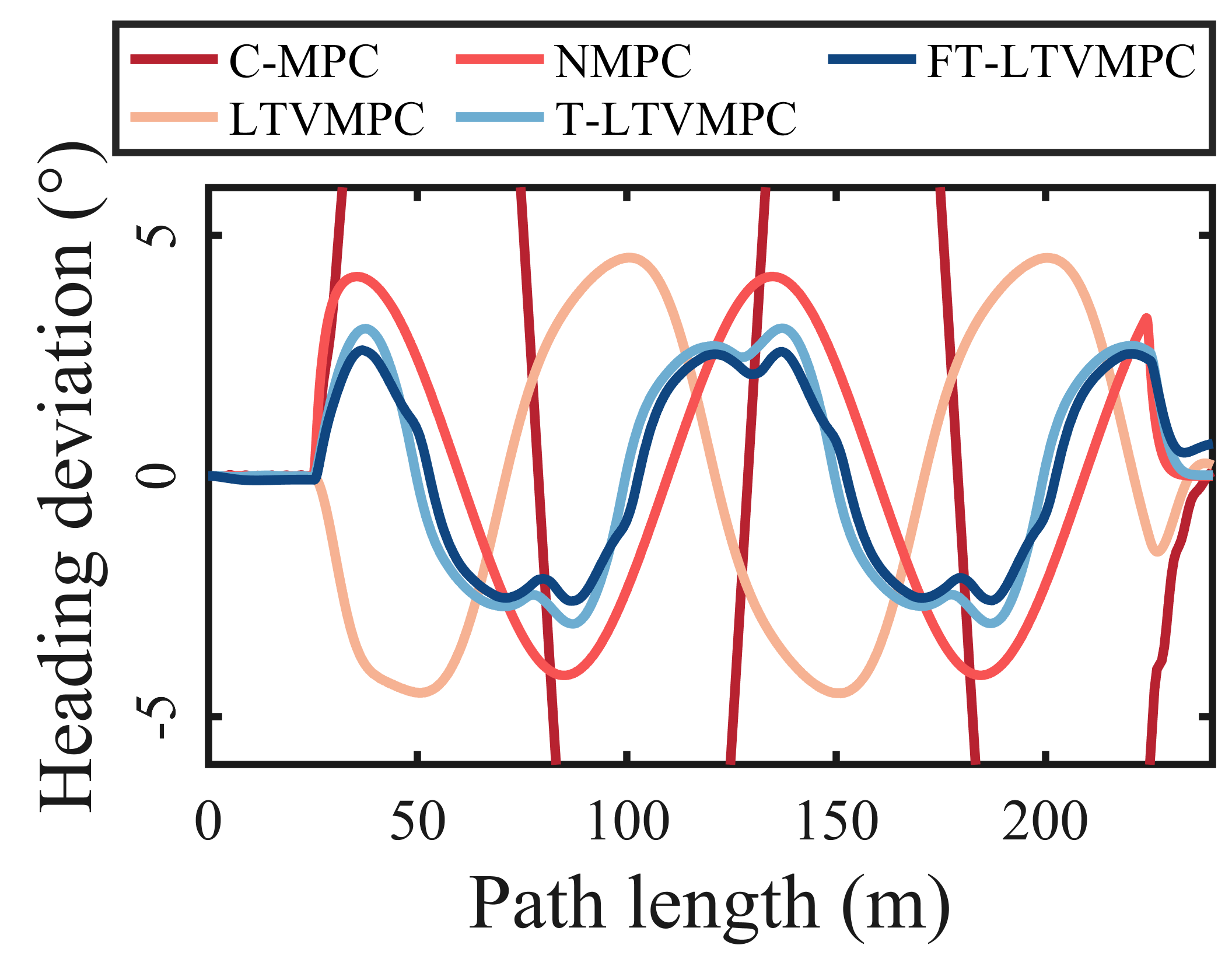}
    }

    \subcaptionbox{\label{fig:SolvingTimeMthd}}[0.6\textwidth]{
        \includegraphics[width=0.6\textwidth]{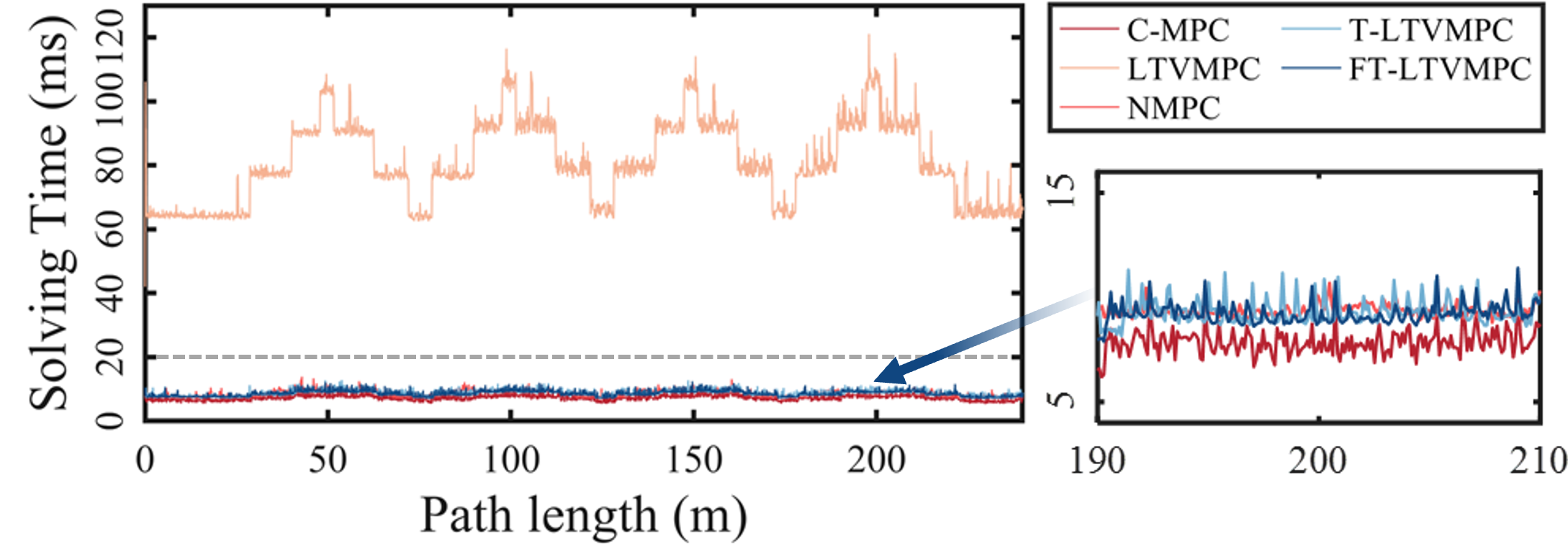} 
    }
  \caption{Results of case 2. (a)Tracking performance of reference relative heading angle, (b)Lateral position tracking error, (c)Heading tracking error, (d)Solving time of various control strategies.}
\end{figure}



In Case 2, the control performance of five strategies is compared: C-MPC, NMPC, LTVMPC, T-LTVMPC, and FT-LTVMPC. C-MPC means conventional MPC, which uses a 2DOF single-track dynamic model for the four-axle vehicle under normal steering mode as its predictive model, while the other strategies employ the proposed generalized \( v \)-\( \beta \)-\( r \) dynamic model in the Frenet coordinate system. NMPC and LTVMPC are shown in Section \ref{sec:ObjandCnstrnts}. T-LTVMPC represents a standard tube-based LTVMPC method, applied without a state error filter. Case 2 maintains the same reference path as Case 1, with a forward velocity of 5 m/s, representative of typical AWOISV operating speed.

The tracking performance for the reference heading angle is shown in Fig.\ref{fig:TrackingPrfmcofHeadAngleMthd}, with lateral and heading angle errors depicted in Fig.\ref{fig:TrackingErrinVarMthd_A}-\ref{fig:TrackingErrinVarMthd_B}. In terms of lateral position tracking, Fig.\ref{fig:TrackingErrinVarMthd_A} reveals that C-MPC has the lowest performance, showing the limitations of its single-track model in adapting to high-curvature paths. NMPC achieves slightly better results than LTVMPC, demonstrating that the proposed dynamic model effectively adapts to high-curvature conditions, while the linear time-varying approach maintains model fidelity. T-LTVMPC and FT-LTVMPC show the highest performance, highlighting the robustness of the tube-based MPC's feedback compensation in addressing model inaccuracies and parameter uncertainties.

In terms of relative heading angle tracking, Figs.\ref{fig:TrackingPrfmcofHeadAngleMthd} indicate that the traditional single-track dynamic model under NSM cannot effectively track arbitrary reference heading angles, as it keeps the steering center aligned with the tangent of the reference path. In Fig.\ref{fig:TrackingErrinVarMthd_B}, the maximum heading angle deviation of C-MPC reaches 25°, exceeding the plot limits. In contrast, the four methods based on the generalized \(v\)-\( \beta \)-\(r\) dynamic model enable simultaneous adjustment of the AWOIS’s theoretical steering radius and sideslip angle through control inputs \(\theta_R\) and \(\beta_R\), allowing simultaneous tracking of lateral position and heading angle. The reduced heading deviation achieved by T-LTVMPC and FT-LTVMPC in Fig.\ref{fig:TrackingErrinVarMthd_B} further demonstrates the robustness of the tube-based approach. Fig.\ref{fig:SolvingTimeMthd} shows the solving times, with all strategies except NMPC remaining under 20 ms, meeting real-time requirements. Considering both tracking performance and real-time capability, LTV-MPC, T-LTVMPC, and FT-LTVMPC are the most suitable for AWOISV's specific applications.

\subsection{Case 3: Composite Path with Time-Varying Forward Velocity}
\begin{figure}[htbp]
  \centering
    \subcaptionbox{\label{fig:hilRfrncPath}}[0.6\textwidth]{
        \includegraphics[width=0.6\textwidth]{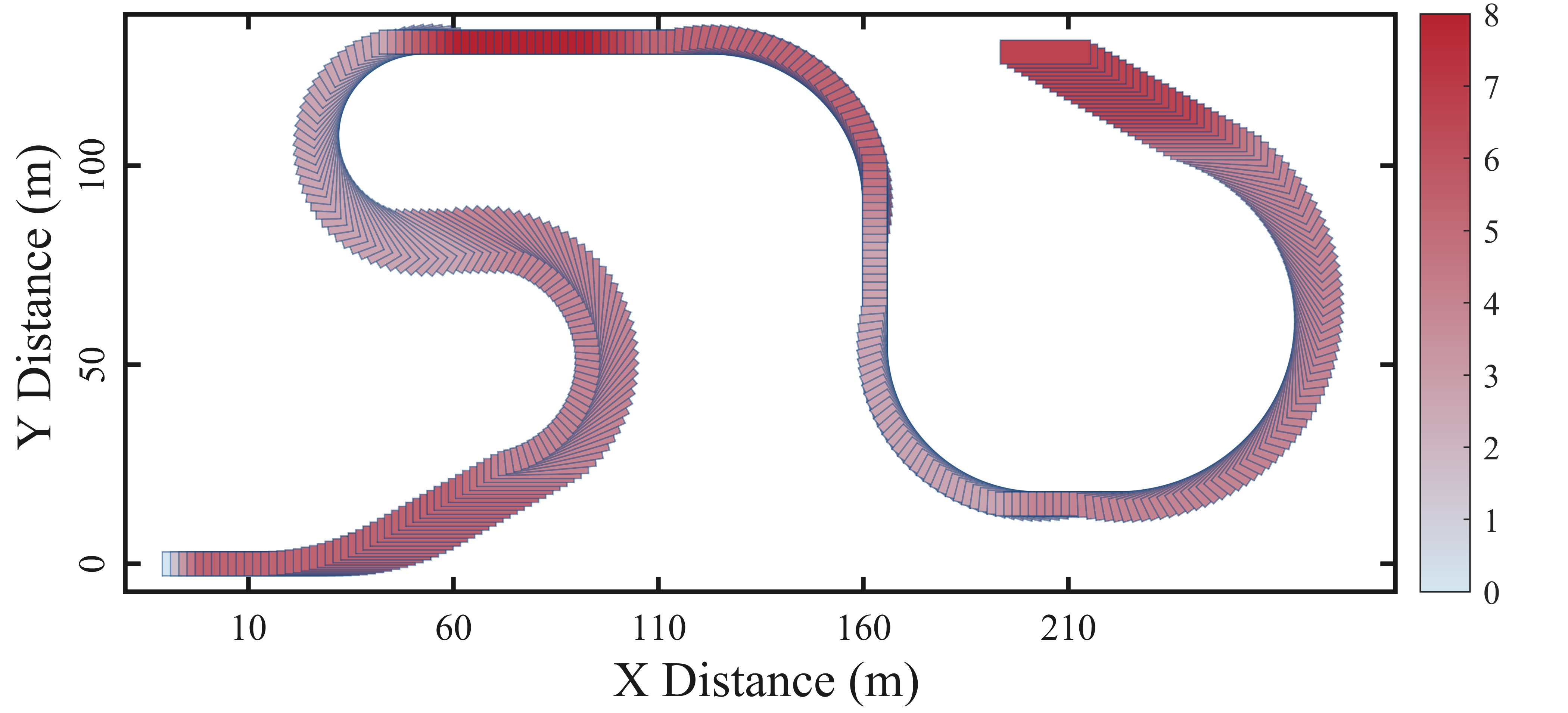} 
    }

    \subcaptionbox{\label{fig:TrackingErrinHIL_a}}[0.3\textwidth]{
        \includegraphics[width=0.3\textwidth]{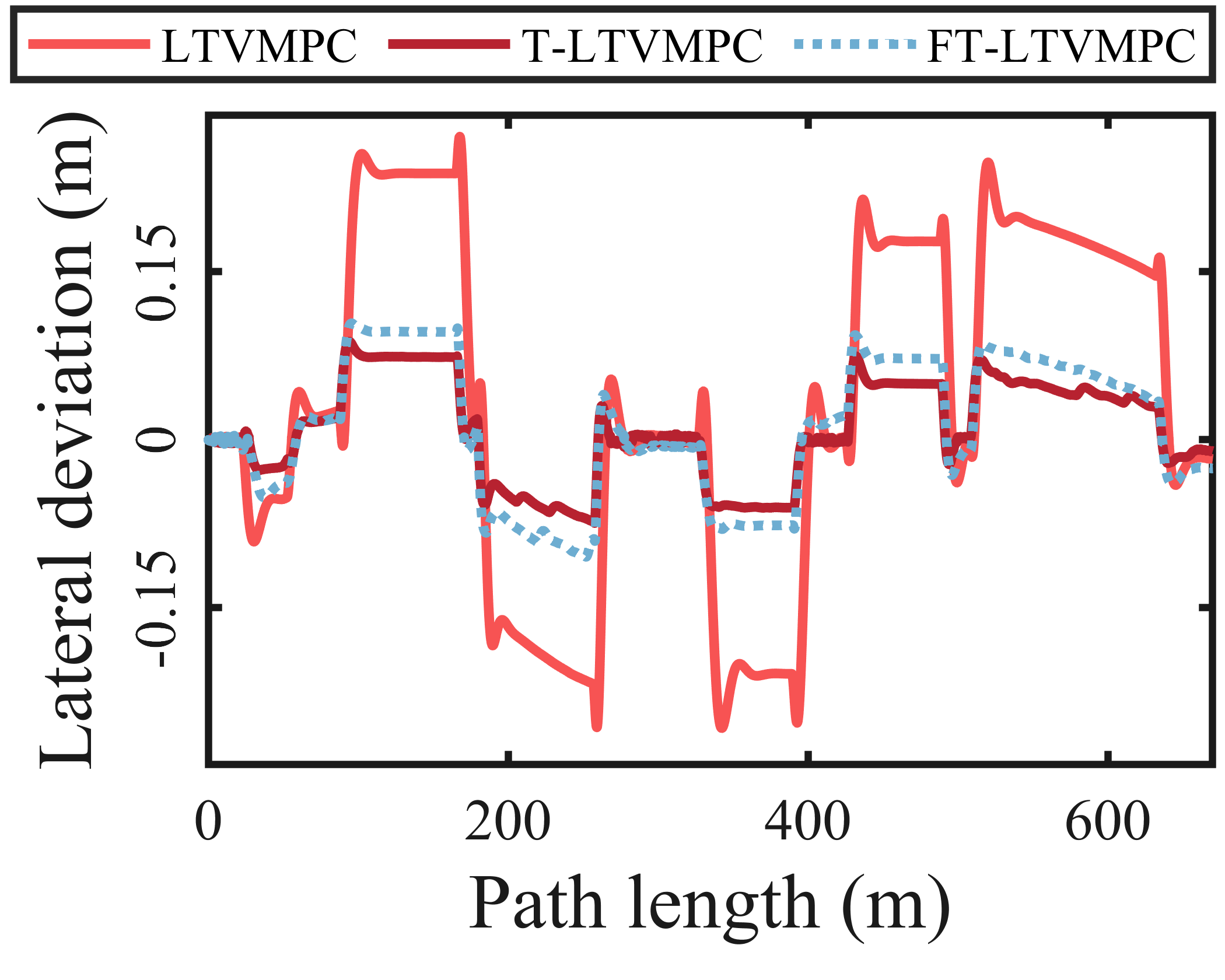} 
    }
    \hspace{2pt}
    \subcaptionbox{\label{fig:TrackingErrinHIL_b}}[0.3\textwidth]{
        \includegraphics[width=0.3\textwidth]{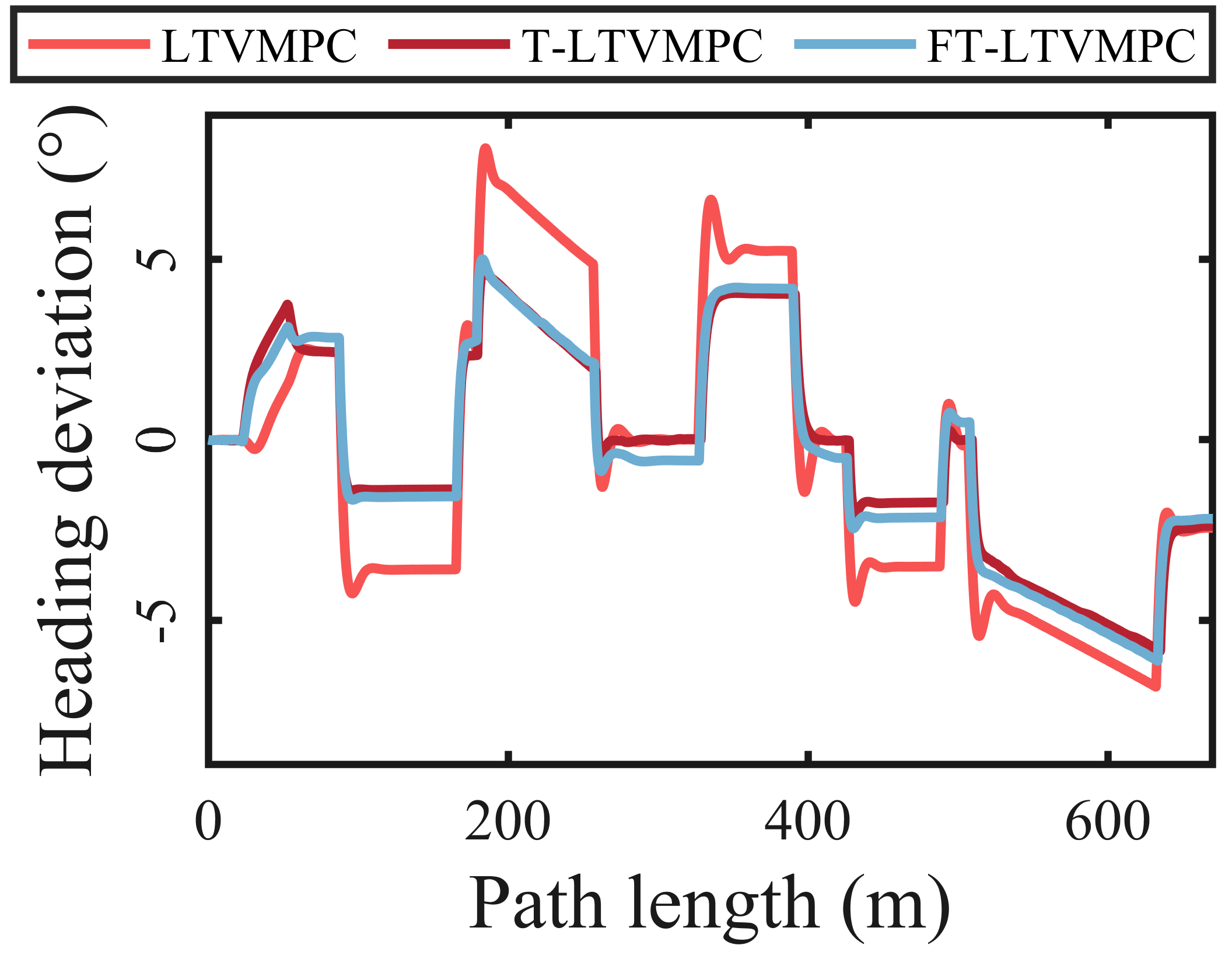}
    }

  \subcaptionbox{\label{fig:TrackingErrinHIL_c}}[0.3\textwidth]{
        \includegraphics[width=0.3\textwidth]{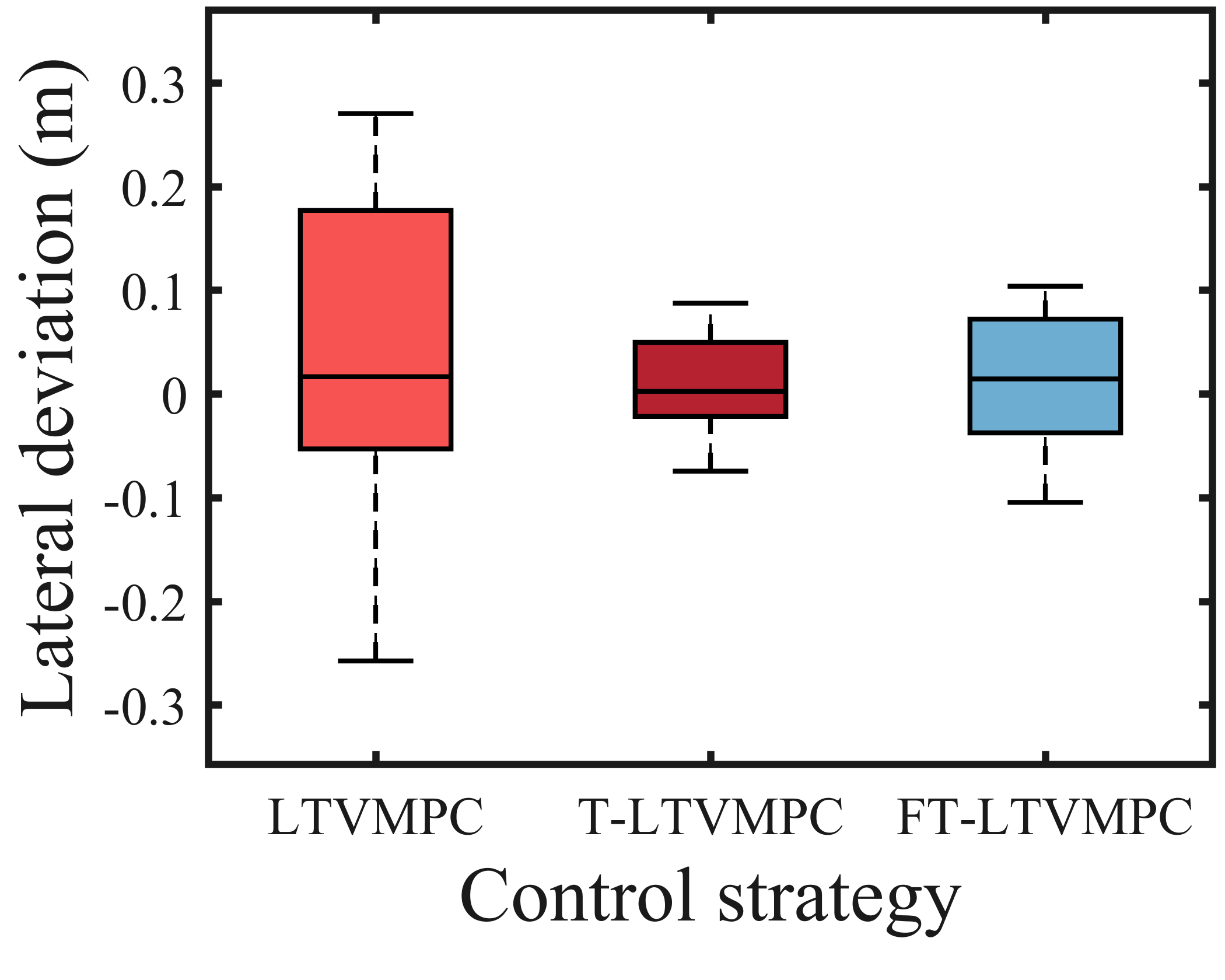} 
    }
    \hspace{2pt}
    \subcaptionbox{\label{fig:TrackingErrinHIL_d}}[0.3\textwidth]{
        \includegraphics[width=0.3\textwidth]{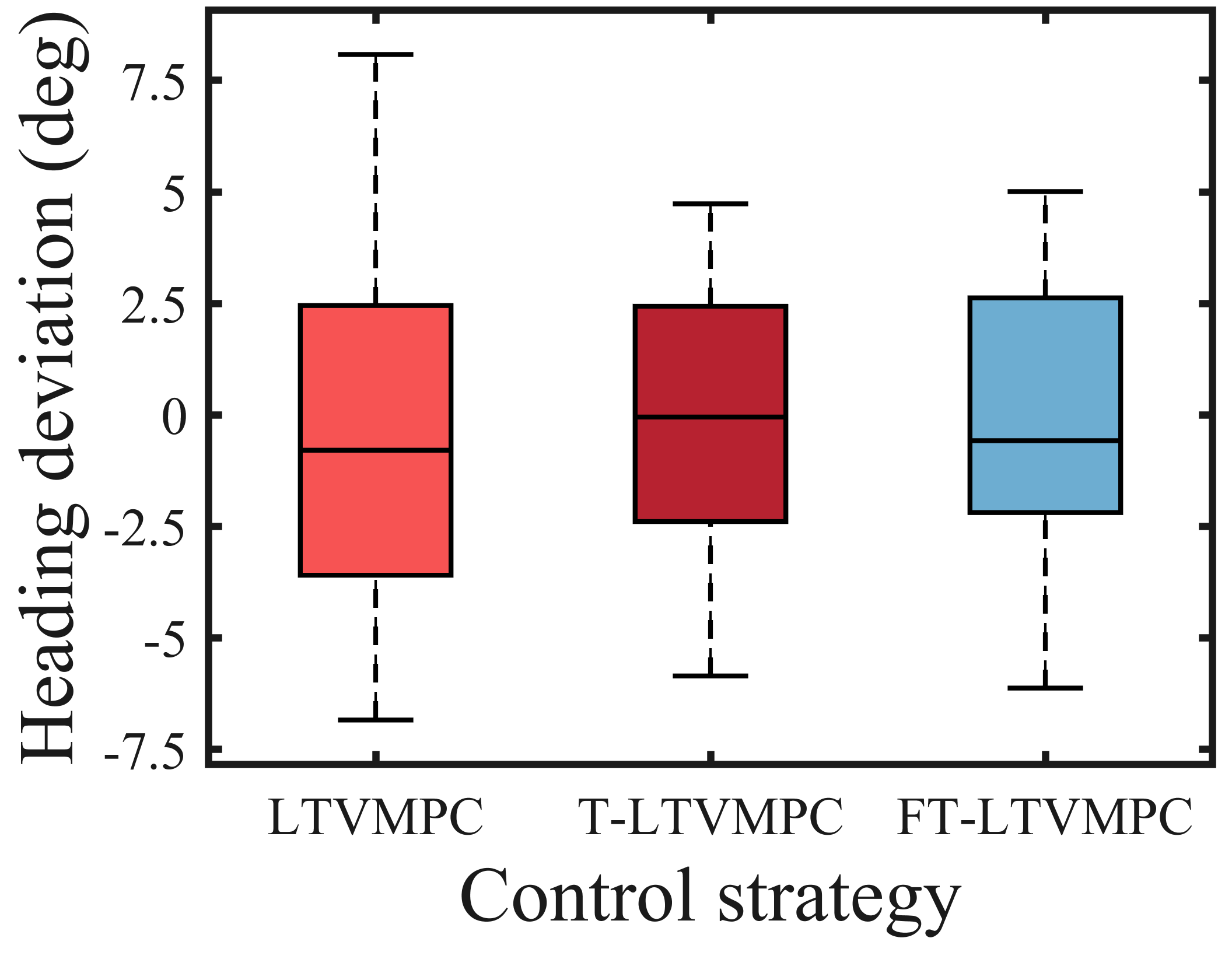}
    }

  \subcaptionbox{\label{fig:yawRateinHIL}}[0.6\textwidth]{
        \includegraphics[width=0.6\textwidth]{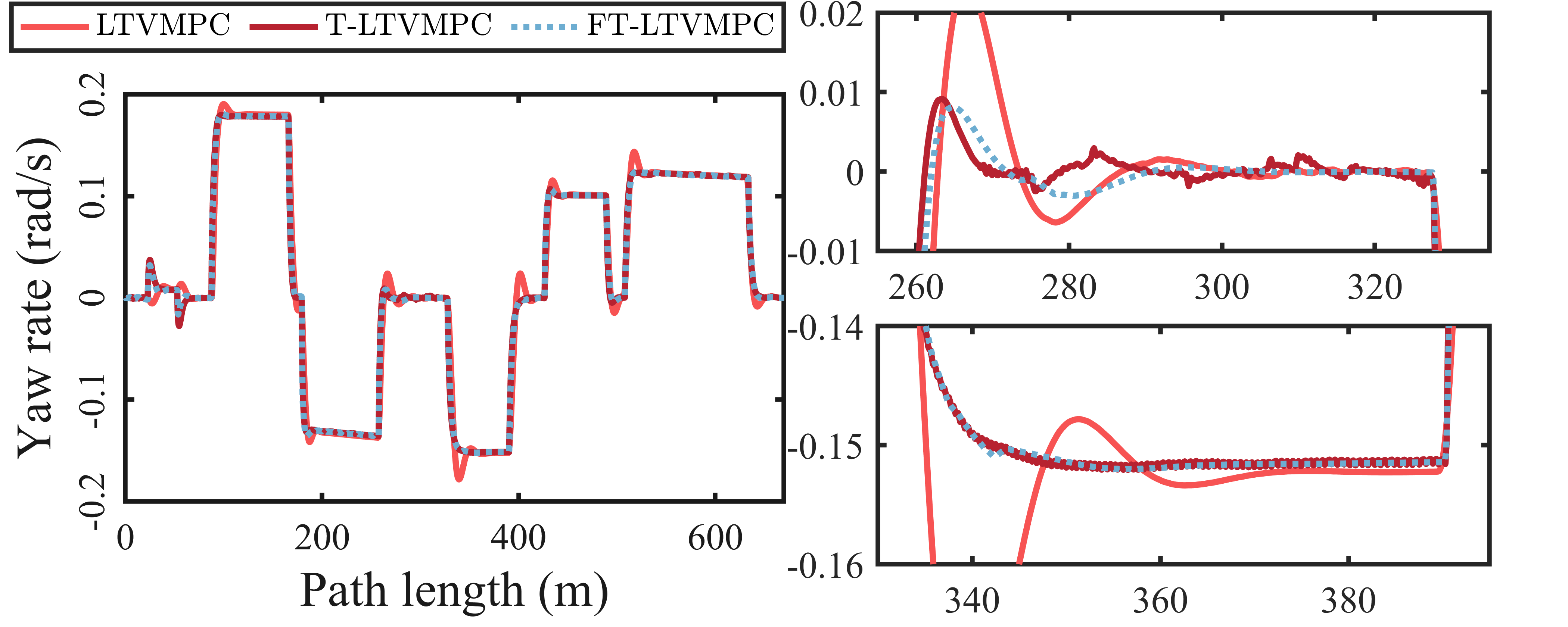} 
    }

  \subcaptionbox{\label{fig:whlAgleinHIL}}[0.6\textwidth]{
        \includegraphics[width=0.6\textwidth]{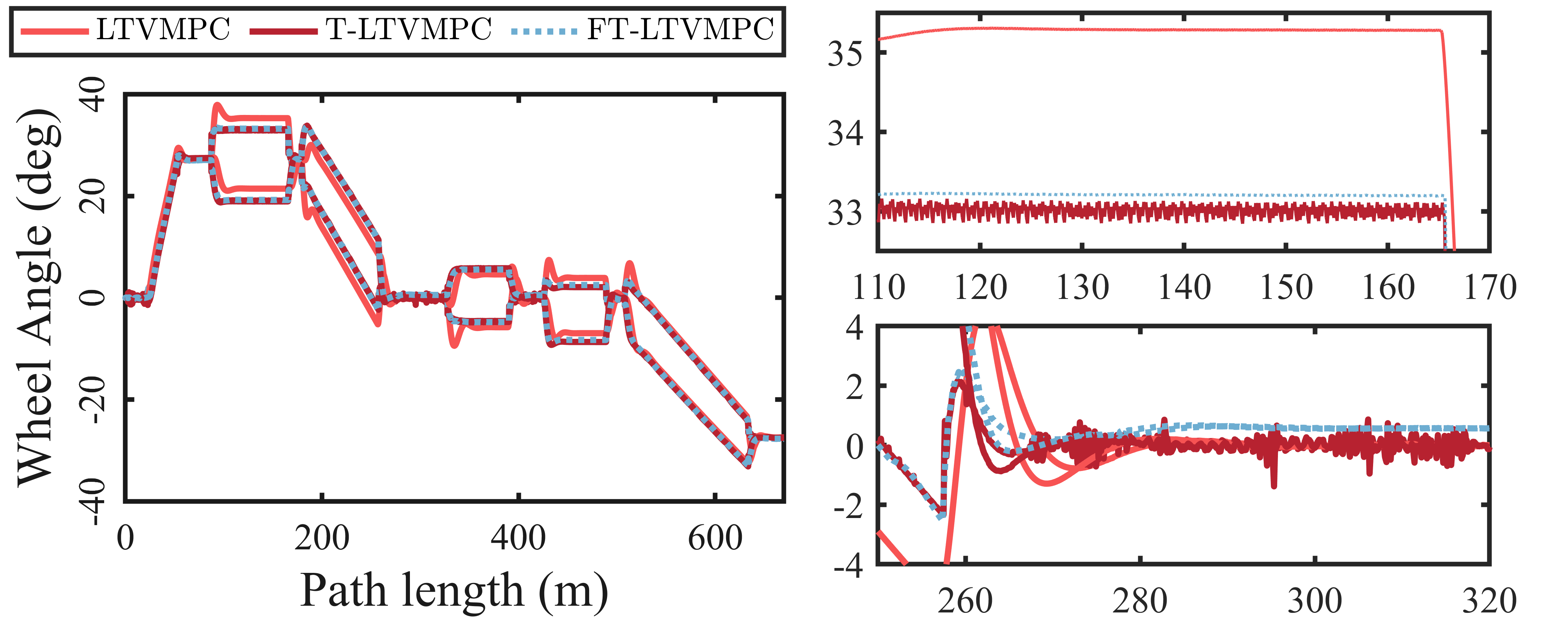} 
    }
  \caption{Results of case 3. (a)Reference path of the HIL experiment, (b) Lateral position tracking error (curve plot), (c) Heading angle tracking error (curve plot), (d) Lateral position tracking error (box plot), (e) Heading angle tracking error (box plot), (f)Yaw rate comparison, (g)Wheel angle comparison. }
  \label{fig:TrackingErrinHIL}
\end{figure}

A preliminary evaluation of five control approaches has already been conducted. LTV-MPC, T-LTV-MPC, and FT-LTV-MPC show excellent performance in simultaneously tracking the relative heading angle and lateral position, which are specific to the AWOISV, while achieving shorter solving times. To further compare control performance in real-time environments, HIL experiments are conducted in Case 3. To simulate real-world operating conditions for the AWOISV, a composite Path is designed as shown in the Fig.\ref{fig:hilRfrncPath}, which combines three motion modes, LoSM, LoSDM, and DSDM, with forward velocity varying between 0 and 8 m/s.


Fig.\ref{fig:TrackingErrinHIL} presents the path tracking performance of the LTVMPC, T-LTVMPC, and FT-LTVMPC strategies under varying forward velocities. As shown in Fig.\ref{fig:TrackingErrinHIL_a} and Fig.\ref{fig:TrackingErrinHIL_b}, all three strategies demonstrate acceptable tracking performance in this complex scenario, with T-LTVMPC and FT-LTVMPC significantly outperforming LTVMPC. Fig.\ref{fig:TrackingErrinHIL_c} and Fig.\ref{fig:TrackingErrinHIL_d} provide a quantitative assessment of tracking performance. Using the median tracking error as the evaluation metric, T-LTVMPC reduces the lateral position and relative heading angle errors by 73.9\% and 34.8\%, respectively, compared to LTVMPC. FT-LTVMPC achieves reductions of 61.3\% and 37.9\%. Additionally, Fig.\ref{fig:TrackingErrinHIL_a} and Fig.\ref{fig:TrackingErrinHIL_b} show that these improvements are particularly noticeable in high-curvature sections of the path. When the AWOISV undergoes high-curvature or large side-slip angle maneuvers, uncertainties arise not only from the linearization process but also from load transfer and vehicle body roll. Despite these challenges, both T-LTVMPC and FT-LTVMPC maintain strong tracking performance, demonstrating the robustness of these tube-based control strategies.



Fig.\ref{fig:yawRateinHIL} compares the yaw rate performance of the three control strategies. It can be observed that LTVMPC generates a harsher yaw rate during transitions between motion modes, such as the shift from DSDM to LoSM, which results in increased tracking error and less smooth movement of the AWOISV. In contrast, both T-LTVMPC and FT-LTVMPC deliver smoother yaw rates over the entire path-tracking task. However, the zoomed-in view shows T-LTVMPC has a noticeably higher yaw rate vibration frequency than FT-LTVMPC, reflecting a difference in the smoothness of the two strategies.

Fig.\ref{fig:whlAgleinHIL} uses the wheel steering angles of the first and fourth axles as examples to illustrate the variation in control inputs among the three strategies. It is clear that both LTV-MPC and FT-LTVMPC exhibit smoother control inputs compared to T-LTVMPC. The zoomed-in view highlights that this difference is particularly pronounced in scenarios with a constant curvature radius, such as during the circular and diagonal path conditions. This paper employs a sliding standard deviation method to quantify and access the vibration in control inputs among the three strategies. The sliding standard deviation \(\sigma_{mov}\) is calculated for the wheel steering angles of all eight wheels for each control strategy. It represents the local standard deviation based on a given window size \(W\), as follows:

\begin{figure}[b]
  \centering
    \subcaptionbox{}[0.3\textwidth]{
        \includegraphics[width=0.3\textwidth]{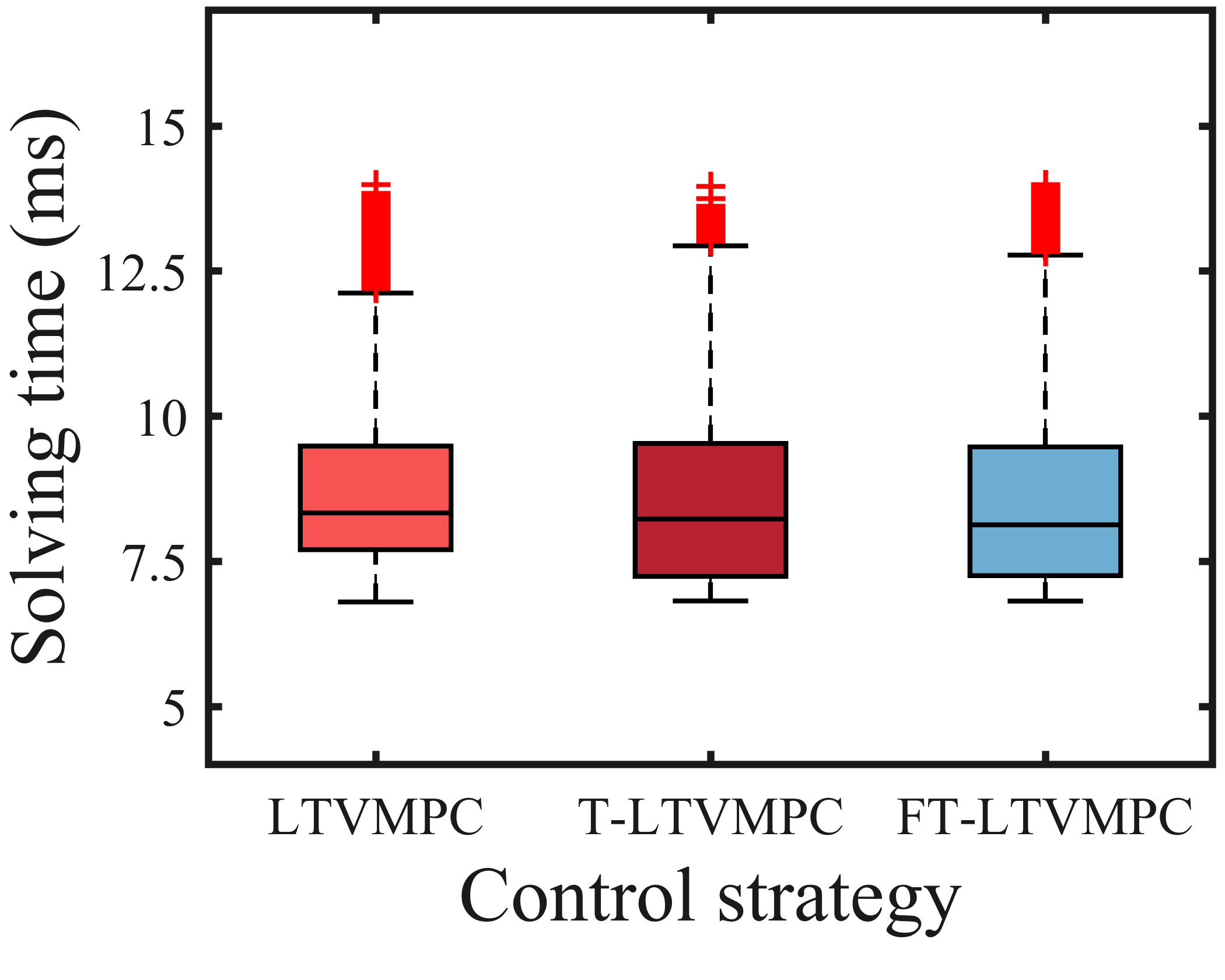} 
    }
    \hspace{2pt}
    \subcaptionbox{}[0.3\textwidth]{
        \includegraphics[width=0.3\textwidth]{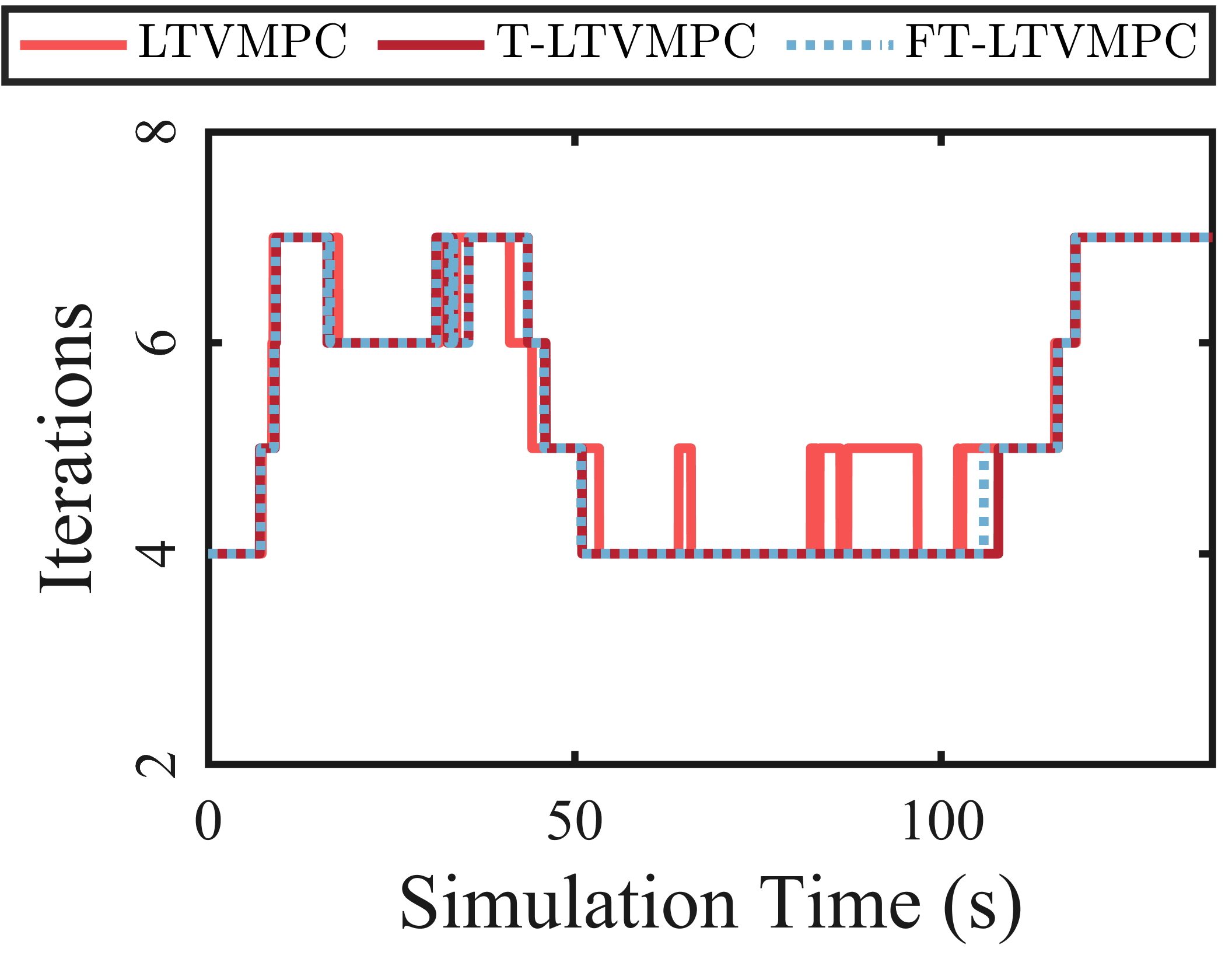}
    }
  \caption{Solving efficiency comparison. (a)Solving time, (b)Iterations.}
  \label{fig:slvEffCmprsn}
\end{figure}

\begin{equation}
\begin{aligned}
  &\sigma_{mov}(\delta_{i*}, W) \\
  &= \sqrt{\frac{1}{N - W + 1} \sum_{n=1}^{N - W + 1} 
  \left( \frac{1}{W} \sum_{j=0}^{W-1} (\delta_{i*, n+j} - {\bar \delta_{i*, n}})^2 \right)}  
\end{aligned}
\end{equation}
where \( N \) represents the total number of data points, and \( \bar \delta_{i*,n} \) denotes the mean wheel steering angle within the current window.

For each controller \( k \), with \( k = 1, 2, 3 \) corresponding to LTVMPC, T-LTVMPC, and FT-LTVMPC, the average sliding standard deviation of the control inputs \( \sigma_{\text{avg}}^k \) is calculated as follows:
\begin{equation}
  \sigma_{avg}^k = \frac{1}{8} \sum_{* = L}^{R} \sum_{i = 1}^{4} \sigma_{mov}(\delta_{i*}^k, W).
\end{equation}

The average standard deviations of LTVMPC, T-LTVMPC, and FT-LTVMPC are 0.0407, 0.0782, and 0.0341, respectively, which are consistent with the results shown in Fig.\ref{fig:whlAgleinHIL}. FT-LTVMPC significantly improves the smoothness of control inputs by 56.4\% compared to T-LTVMPC. This indicates that T-LTVMPC experiences significant fluctuations in control inputs due to the jitter of \(\bm e_k\) when compensating for model uncertainties. In contrast, FT-LTVMPC effectively addresses this issue through the filter for \(\bm e_k\), ensuring both the precision and robustness of the control strategy.

To verify the real-time computational efficiency of the proposed three strategies, the total solving time and the iteration counts for the MPC problem are presented in Fig.\ref{fig:slvEffCmprsn}. The solving times for all three strategies are approximately 8 ms, and the maximum solving times remain under 15 ms. Compared to the IPC's calculation period of 20 ms, the solving time is lower and meets the real-time requirements. Furthermore, since all three strategies are based on the AWOISV dynamic model introduced in Section \ref{sec:prdctvMdl}, the iteration counts for solving the MPC problem are nearly identical, with optimal control inputs obtained within 7 iterations.

In summary, LTVMPC, T-LTVMPC, and FT-LTVMPC all meet the real-time requirements in the HIL experiments. Among these, T-LTVMPC and FT-LTVMPC demonstrate higher path-tracking accuracy, while LTVMPC and T-LTVMPC show smoother control inputs. Overall, FT-LTVMPC proves fully capable of handling the simultaneous tracking of lateral position and heading angle for AWOISV, showing great potential for engineering applications.

\section{Conclusion}\label{sec:conclusion}
This paper focuses on the modeling and control of the AWOISV to achieve simultaneous tracking of lateral position and heading angle. The wheel steering angle is mapped to the ICR through the parameters \(\theta_R\) and \(\beta_R\). Based on the relative position of the ICR and WCR, six motion modes (FASM, RASM, NSM, YSM, PSM, and DSM) and their switching criteria are defined. A generalized \(v\)-\(\beta\)-\(r\) dynamic model, applicable to AWOISVs with any number of axles, is developed with forward velocity \(v\), sideslip angle \(\beta\), and yaw rate \(r\) as state variables, which explicitly represents the forward and rotational motions of the AWOISV. Model fidelity is validated by comparison with TruckSim.

The FT-LTVMPC strategy is proposed, using the \(v\)-\(\beta\)-\(r\) model in the Frenet coordinate system as the predictive model, with filtered feedback compensation. The approach is validated through both simulation and HIL platforms. Simulation results demonstrate the robustness of FT-LTVMPC across varying speeds and show significant advantages in simultaneously tracking lateral position and heading angle compared to traditional path-tracking MPC strategies. HIL results indicate that FT-LTVMPC improves position and heading tracking accuracy by 61.3\% and 37.9\%, respectively, over standard LTVMPC. Control smoothness also increases by 56.4\% compared to T-LTVMPC. Additionally, FT-LTVMPC achieves an 8 ms solution time, meeting real-time control requirements.

In future work, FT-LTVMPC will be integrated with driving force allocation control to enhance vehicle stability and adaptability. Additionally, FT-LTVMPC will be applied to a real AWOISV to evaluate its practical advantages and effectiveness under real-world conditions.


\bibliographystyle{apacite}
\bibliography{reference1103}

\end{document}